%% file: acl_latex.tex
\documentclass[11pt]{article}

\usepackage[final]{acl}

\usepackage{times}
\usepackage{latexsym}

\usepackage[T1]{fontenc}

\usepackage[utf8]{inputenc}

\usepackage{microtype}

\usepackage{inconsolata}

\usepackage{graphicx}

\usepackage{enumitem}
\usepackage{amsmath}
\usepackage{amssymb}
\usepackage{amsfonts}
\usepackage{mathtools}
\usepackage{booktabs}
\usepackage{multirow}
\usepackage{graphicx}
\usepackage{capt-of}
\usepackage{xcolor}
\usepackage{colortbl}
\usepackage{geometry}
\usepackage[most]{tcolorbox}
\usepackage[ruled,vlined,linesnumbered]{algorithm2e}
\definecolor{rowblue}{HTML}{EBF5FE}

\DontPrintSemicolon
\SetKwInput{KwIn}{Input}
\SetKwInput{KwInit}{Initialize}
\SetKwInput{KwOut}{Output}
\SetKwFor{For}{for}{do}{end for}

\title{Anti-Length Shift: Dynamic Outlier Truncation for \\Training Efficient Reasoning Models}

\author{%
  \textbf{Wei Wu}\textsuperscript{1,\footnotemark[4]},
  \textbf{Liyi Chen}\textsuperscript{2,\footnotemark[1]},
  \textbf{Congxi Xiao}\textsuperscript{1},
  \textbf{Tianfu Wang}\textsuperscript{3},
  \textbf{Qimeng Wang}\textsuperscript{2},\\[2mm]
  \textbf{Chengqiang Lu}\textsuperscript{2},
  \textbf{Yan Gao}\textsuperscript{2},
  \textbf{Yi Wu}\textsuperscript{2},
  \textbf{Yao Hu}\textsuperscript{2},
  \textbf{Hui Xiong}\textsuperscript{3,4,\footnotemark[1]}\\[2mm]
  \textsuperscript{1}University of Science and Technology of China,\;
  \textsuperscript{2}Xiaohongshu Inc.,\\
  \textsuperscript{3}The Hong Kong University of Science and Technology (Guangzhou),\\
  \textsuperscript{4}The Hong Kong University of Science and Technology\\[2mm]
  \fontsize{10pt}{12pt}\selectfont \texttt{urara@mail.ustc.edu.cn, liyichencly@gmail.com, xionghui@ust.hk}
}

\begin{document}
\maketitle

\renewcommand{\thefootnote}{\fnsymbol{footnote}}
\footnotetext[1]{Corresponding Authors.}
\footnotetext[4]{Work done during an internship at Xiaohongshu Inc.}
\footnotetext[2]{Models: \url{http://hf.co/collections/U-rara/dot}}
\footnotetext[3]{Codes: \url{https://github.com/U-rara/DOT}}

\input{sections/0_abstract}
\input{sections/1_introduction}

\input{sections/2_preliminary}
\input{sections/3_empirical_analysis}
\input{sections/4_methodology}
\input{sections/5_experiments}
\input{sections/6_related_work}
\input{sections/7_conclusion}

\bibliography{custom}

\appendix

\input{sections/8_appendix}

\end{document}

%% file: sections/0_abstract.tex
\begin{abstract}
Large reasoning models enhanced by reinforcement learning with verifiable rewards have achieved significant performance gains by extending their chain-of-thought. However, this paradigm incurs substantial deployment costs as models often exhibit excessive verbosity on simple queries. Existing efficient reasoning methods relying on explicit length penalties often introduce optimization conflicts and leave the generative mechanisms driving overthinking largely unexamined. In this paper, we identify a phenomenon termed length shift where models increasingly generate unnecessary reasoning on trivial inputs during training. To address this, we introduce Dynamic Outlier Truncation (DOT), a training-time intervention that selectively suppresses redundant tokens. This method targets only the extreme tail of response lengths within fully correct rollout groups while preserving long-horizon reasoning capabilities for complex problems. To complement this intervention and ensure stable convergence, we further incorporate auxiliary KL regularization and predictive dynamic sampling. Experimental results across multiple model scales demonstrate that our approach significantly pushes the efficiency-performance Pareto frontier outward. Notably, on the AIME-24, our method reduces inference token usage by 78\% while simultaneously increasing accuracy compared to the initial policy and surpassing state-of-the-art efficient reasoning methods.
\end{abstract}

%% file: sections/1_introduction.tex
\section{Introduction}
\input{figures/tex/fig1}
A widely used way to improve the reasoning capability of large language models (LLMs) is to make them “think longer”~\cite{tts}. Recent reasoning models~\cite{o1,r1,qwen} have systematized this idea by combining mid-training data refinement with post-training reinforcement learning with verifiable rewards (RLVR)~\cite{interplay}, yielding policies that allocate more computation at inference time via longer reasoning trajectories. This paradigm delivers substantial gains on challenging benchmarks, making long chain-of-thought (CoT)~\cite{cot} nearly a standard component of state-of-the-art models.

Nonetheless, this paradigm incurs substantial deployment costs. Long-CoT policies often expend excessive tokens on trivial queries, manifesting as repetitive restatements, backtracking, and post-hoc self-checks that rarely alter the final answer~\cite{overthinking}.
Given that modern reasoning models are predominantly post-trained via RLVR, recent efforts on efficient reasoning naturally targets the RL objective and encodes brevity by directly coupling response length with reward.
These methods commonly introduce explicit length-aware shaping through intra-group comparisons~\cite{o1-pruner,kimi-1.5,DAST,LCR}, explicit budget adherence~\cite{L1}, or thresholded rewards for correctness under a target length~\cite{ThinkPrune,AdaptThink,LASER,SIRI,DLER}, demonstrating that significant token reduction can be achieved with little or no accuracy loss.
Despite these empirical successes, such shaping induces optimization conflicts, as gradients for length reduction often diverge from accuracy maximization. This misalignment hinders convergence and suppresses exploration, forcing reliance on heuristic schedules or sensitive tuning to sustain a fragile trade-off. More importantly, existing methods primarily penalize overthinking as an outcome, leaving its generative mechanisms under-investigated. This motivates a mechanism-centric treatment that reduces response length without sacrificing the model's exploration capabilities.

In this paper, we first provide an empirical analysis of why reasoning models become verbose on easy inputs during mid and post-training. We term this phenomenon \textit{length shift}: prompts that are already solved correctly tend to elicit longer responses as training progresses. We find that length shift is accompanied by a higher propensity to emit reasoning words such as verification and hesitation markers; these behaviors are useful under uncertainty, but under a shared policy across difficulty regimes they are over-triggered on trivial queries and inflate length. This calls for an asymmetric intervention that selectively prunes habituated verbosity on trivial inputs, while strictly insulating hard queries from restrictive length constraints. 

To achieve this, we propose \textbf{Dynamic Outlier Truncation (DOT)}. DOT identifies redundancy solely within fully correct rollout groups, using group-wise length statistics to truncate trajectories exceeding a dynamic threshold $T = \mu + \alpha \cdot \sigma$. Since $T$ is determined a posteriori, DOT leaves hard queries unconstrained during generation; meanwhile, its reliance on dynamic distribution statistics minimizes susceptibility to reward hacking.
Further analysis shows that although DOT affects only a tiny fraction of responses (e.g., $\sim 0.5\%$) it can still induce global entropy reduction. We therefore introduce auxiliary KL regularizer~\cite{KL-Cov} to prevent premature entropy collapse, and propose a predictive dynamic sampling strategy to avoid late-stage training being dominated by all-correct queries. 
As shown in \textit{Fig.}~\ref{fig:fig1}, DOT moves the Pareto frontier outward by a large margin. 
For example, on \texttt{DeepSeek-R1-Distill-Qwen-1.5B}, DOT uses only 21.6\% of the average tokens on AIME-24 while improving pass@1 from 30.0\% to 43.1\%.
Moreover, compared with state-of-the-art efficient reasoning methods (e.g., SIRI), DOT uses only 33.3\% of the average tokens on AIME-24 with essentially the same accuracy. In summary, we make the following contributions:
\begin{itemize}[leftmargin=*,noitemsep,topsep=2pt]
    \item An empirical analysis of the length shift phenomenon, revealing that redundancy in reasoning models stems from the over-triggering of reasoning words on trivial queries during training.
    \item Dynamic Outlier Truncation (DOT), a training-time intervention based on group-wise statistics that selectively truncates redundant rollouts in all-correct groups, minimizing reward hacking while preserving long-horizon reasoning capacity.
    \item Results across 1.5B, 7B and 32B models demonstrate that our training recipe generalizes robustly across scales and consistently pushes the efficiency--performance Pareto frontier outward on challenging reasoning benchmarks.
\end{itemize}

%% file: figures/tex/fig1.tex
\begin{figure*}[t]
  \centering
  \includegraphics[width=\linewidth]{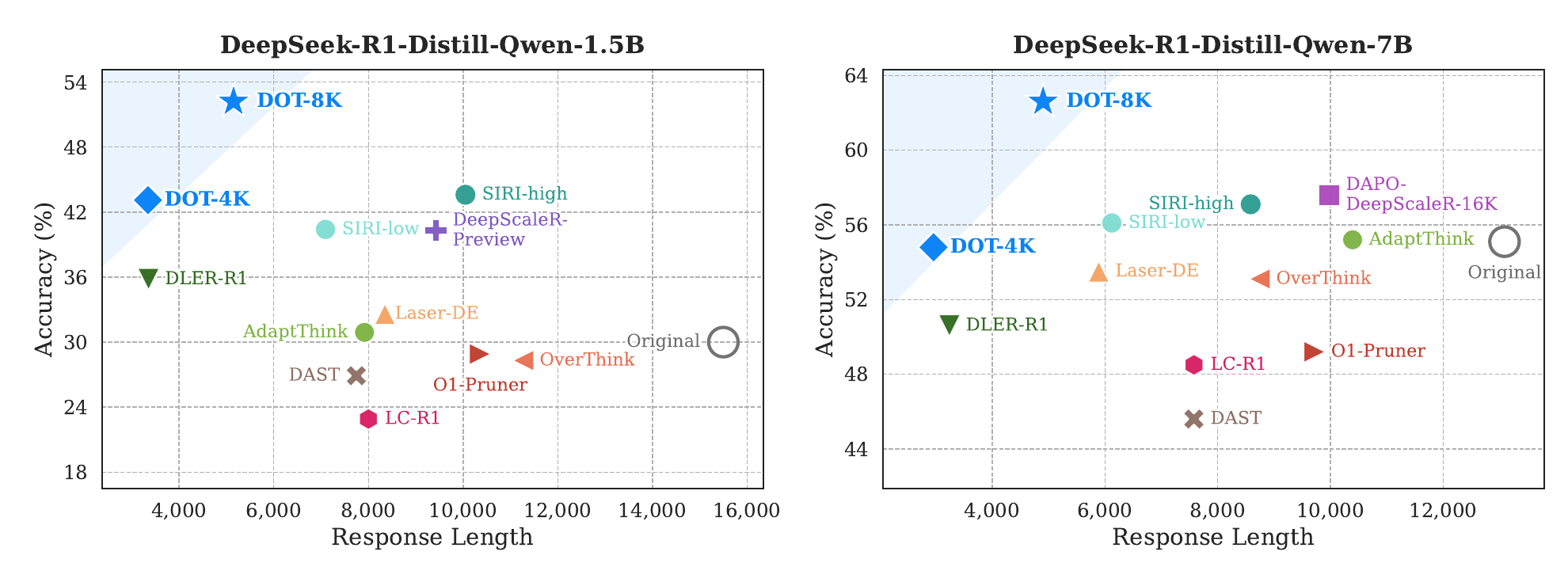}
  \vspace{-25pt}
  \caption{Performance-efficiency comparison on AIME-24 across two model scales.}
  \vspace{-10pt}
  \label{fig:fig1}
\end{figure*}

%% file: sections/2_preliminary.tex
\section{Preliminary}
\input{figures/tex/fig2_3}
In this work, we ground our analysis and discussion in the framework of Group Relative Policy Optimization (GRPO)~\cite{grpo}. GRPO estimates advantages in a group-relative manner and avoids fitting an explicit value function.
For each query--answer pair $(q,a)$, the behavior policy $\pi_{\theta_{\text{old}}}$ samples a rollout group of $G$ responses
$\{o_i\}_{i=1}^G \sim \pi_{\theta_{\text{old}}}(\cdot\mid q)$.
Let $R_i$ denote the reward assigned to $o_i$.
The advantage for the $i$-th response is obtained by normalizing rewards within the group:
\vspace{-5pt}
\begin{equation}
\small
\hat A_i
=
\frac{
R_i - \mathrm{mean}\!\left(\{R_j\}_{j=1}^{G}\right)
}{
\mathrm{std}\!\left(\{R_j\}_{j=1}^{G}\right)
}.
\label{eq:grpo-adv}
\end{equation}
GRPO updates the policy by maximizing the following objective:
\vspace{-5pt}
\begin{equation}
\small
\resizebox{\hsize}{!}{$
\begin{aligned}
\mathcal{J}_{\mathrm{GRPO}}(\theta)
&=\mathbb{E}_{(q,a)\sim\mathcal{D},\{o_i\}_{i=1}^G\sim \pi_{\theta_{\text{old}}}(\cdot\mid q)}
\Bigg[
\frac{1}{G}\sum_{i=1}^G \frac{1}{|o_i|}
\sum_{t=1}^{|o_i|} \\
&\min\!\Big(
r_{i,t}(\theta)\hat A_i,\
\mathrm{clip}\!\big(r_{i,t}(\theta),1-\epsilon,1+\epsilon\big)\hat A_i
\Big)
\Bigg],
\end{aligned}
$}
\label{eq:grpo-obj}
\end{equation}
where \(\epsilon\) is the clipping range of importance sampling ratio:
\vspace{-5pt}
\begin{equation}
\small
r_{i,t}(\theta)
=
\frac{\pi_\theta(o_{i,t}\mid q, o_{i,<t})}{\pi_{\theta_{\text{old}}}(o_{i,t}\mid q, o_{i,<t})}.
\label{eq:grpo-ratio}
\vspace{-5pt}
\end{equation}
In this paper, an explicit KL penalty to a reference policy is omitted following DAPO~\cite{dapo}, which encourages exploration during RL.



%% file: figures/tex/fig2_3.tex
\begin{figure*}[t]
  \begin{minipage}[t]{0.495\textwidth} 
    \centering
    \includegraphics[width=\linewidth]{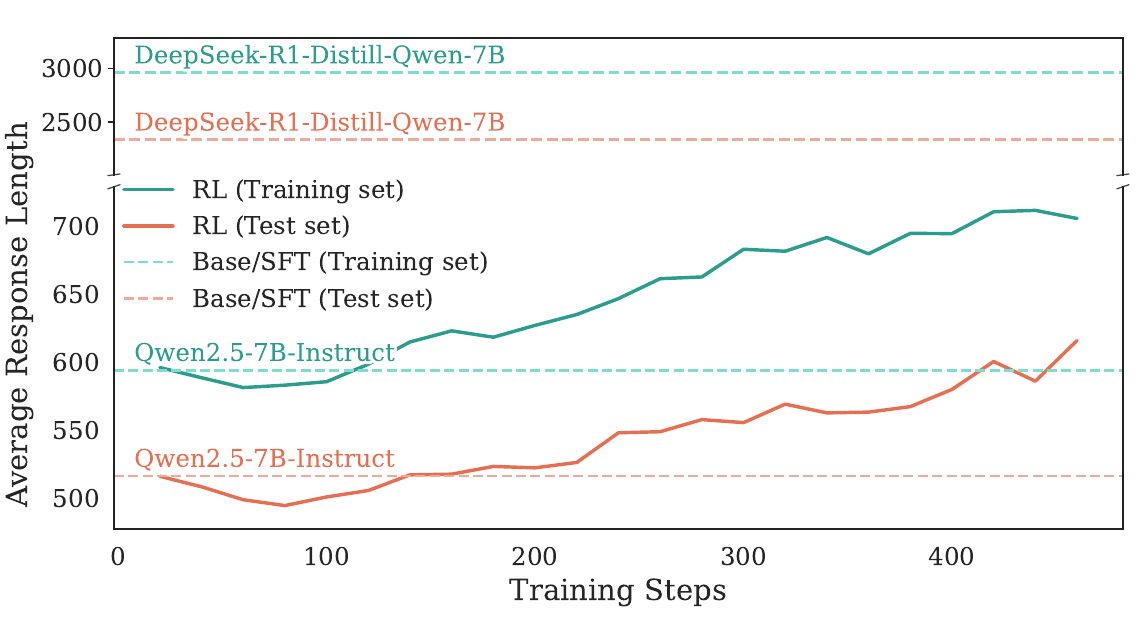}
    \captionsetup{width=0.985\linewidth}
    \vspace{-20pt}
    \captionof{figure}{Evolution of average response length on all-correct queries during RL and SFT training.}
    \label{fig:fig2}
  \end{minipage}
  \vspace{-4pt}
  \begin{minipage}[t]{0.495\textwidth}
    \centering
    \includegraphics[width=\linewidth]{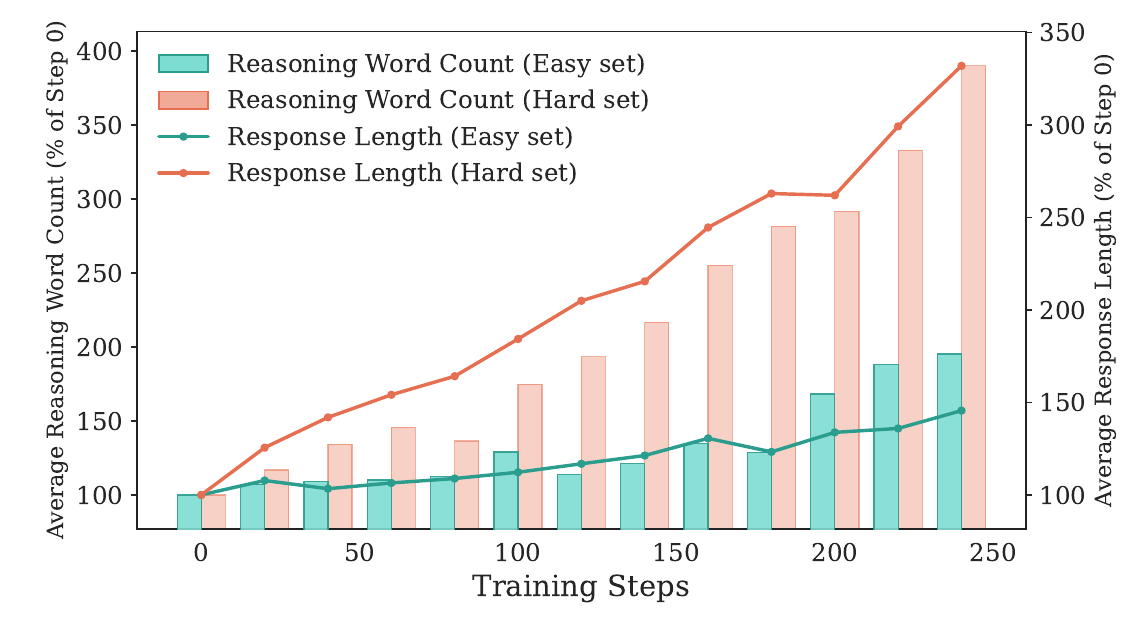}
    \captionsetup{width=0.925\linewidth}
    \vspace{-20pt}
    \captionof{figure}{Co-evolution of reasoning word count and response length on test problems of varying difficulty.}
    \label{fig:fig3}
  \end{minipage}
  \vspace{-4pt}
\end{figure*}

%% file: sections/3_empirical_analysis.tex
\section{Empirical Analysis}

\label{sec:empirical}
In this section, we attempt to answer a fundamental question through empirical analysis: \textit{why do reasoning models produce increasingly long responses on problems that can already be solved easily?} Our key observation is that length growth persists even on all-correct queries for which GRPO yields zero advantages, indicating that overthinking arises from a global shift in policy behavior rather than any need to improve correctness on those problems.

Concretely, under RL with GRPO, we construct an all-correct set consisting of queries that are already solved correctly at the start of training, where each rollout group satisfies \(\{R_i\}_{i=1}^{G}=\mathbf{1}\).
For such queries, \textit{Eq.}~(\ref{eq:grpo-adv}) implies that the group-wise reward has zero variance, yielding $\hat A_i = 0$ for all responses.
Consequently, these prompts contribute zero gradient signal under \textit{Eq.}~(\ref{eq:grpo-obj}) for improving correctness, and thus should not be systematically driven to become longer.
Counterintuitively, \textit{Fig.}~\ref{fig:fig2} shows the opposite: both SFT (as a proxy for mid-training) and RL (post-training), starting from the same base model, increase the average response length \(\mathbb{E}[|o|]\) on this all-correct set, and the trend consistently holds on both training and test splits, indicating a genuine distributional shift instead of overfitting.

To go beyond this phenomenological trend and uncover the mechanism behind length shift, we ask what internal behavioral change accompanies this drift and could plausibly mediate it. Specifically, we track the model's propensity to emit a set of reasoning words, lexical markers of hesitation, verification, and reflection (e.g., ``however'' and ``wait'') that often precede a new reasoning segment~\cite{ORZ,lp-reg}. 
\textit{Fig.}~\ref{fig:fig3} shows that during RL these markers increase over training steps on both easy (pass@1 = 1.0) and hard (pass@1 $\le$ 0.6) subsets, accompanied by a co-evolving increase in \(\mathbb{E}[|o|]\).
The hard subset exhibits a larger growth, consistent with the view that reasoning words supports exploration and helps resolve complex problems~\cite{dapo}.
However, the concurrent upward drift on the easy subset supports a different and more concerning implication: learning on hard queries increases the global prior of emitting reasoning words under a shared policy, and once such words are generated, they tend to initiate additional reasoning spans that compound into longer trajectories even when further deliberation is unnecessary.
Overall, this empirical analysis suggests that overthinking on trivial inputs can be viewed as cross-difficulty policy interference, where behaviors that are beneficial for handling uncertainty on hard problems are overly activated on easy ones. This perspective motivates interventions that suppress only the redundant long tail on already-solved queries, while preserving exploration capacity where it is genuinely needed.

%% file: sections/4_methodology.tex
\input{figures/tex/fig4}

\vspace{-10pt}
\section{Methodology}
\vspace{-2pt}

\label{sec:methodology}
In this section, we propose a simple RL training recipe to counter length shift while preserving the model’s exploration ability. Motivated by the analysis in \textit{Sec.}~\ref{sec:empirical}, we depart from prior approaches that explicitly incorporate response length into the reward. Instead, we introduce \textbf{Dynamic Outlier Truncation (DOT)}, a training-time intervention that trims only the extreme length tail after rollouts are sampled, and only on prompts that the policy already solves reliably (as illustrated in \textit{Fig.}~\ref{fig:fig4}). Furthermore, by inspecting the resulting training dynamics, we incorporate two targeted auxiliary techniques that further improve the stability and scalability of RL training.

\subsection{Dynamic Outlier Truncation (DOT)}
DOT operates on rollout groups in GRPO-style RL. Given a query \(q\) with a rollout group \(\{o_i\}_{i=1}^G\) and rewards \(\{R_i\}_{i=1}^G\), we apply DOT only when \(\{R_i\}_{i=1}^{G}=\mathbf{1}\), i.e., all sampled responses are correct under the task reward. In GRPO, such groups exhibit no residual learning pressure for correctness (their group-relative advantages vanish due to zero reward variance), so an overly long trajectory should be understood as redundancy rather than uncertainty-driven exploration. 

To identify redundant verbosity without introducing a length objective into the reward, DOT uses a-posteriori, group-wise statistics. Let \(L_i = |o_i|\) denote the response length in tokens. We compute the group mean and standard deviation,
\vspace{-6pt}
\begin{equation}
\small
\mu_L = \mathrm{mean}(\{L_i\}_{i=1}^G), \quad \sigma_L = \mathrm{std}(\{L_i\}_{i=1}^G),
\vspace{-6pt}
\end{equation}
and define an outlier cutoff
\vspace{-6pt}
\begin{equation}
\small
T(q) = \left\lfloor \mu_L + \alpha \cdot \sigma_L \right\rfloor.
\label{eq:tq}
\vspace{-6pt}
\end{equation}
Using a ``three-sigma''-style threshold makes DOT target only statistical outliers, leaving typical rollouts unchanged.

To avoid unstable behavior when \(\sigma_L\) becomes small late in training, we apply truncation only when the potential reduction is non-trivial, \(L_i - T(q) \ge m\).
This margin \(m\) prevents frequent minor edits from injecting gradient noise.

Formally, DOT can be viewed as a post processing on sampled rollouts:
\vspace{-6pt}
\begin{equation}
\small
\hat{o}_i \;=\;
\begin{cases}
o_{i,1:T(q)}, & \text{if } \{R_j\}_{j=1}^{G}= \mathbf{1}\ \text{and}\ L_i - T(q) \ge m\\
o_i, & \text{else}
\end{cases}
\label{eq:dot}
\vspace{-6pt}
\end{equation}
After truncation, we recompute the reward on the modified rollouts, ensuring that any dependence on later tokens is reflected in the task reward, while leaving the standard GRPO update unchanged.

\input{figures/tex/fig5_6_7}
\subsection{Stabilizing and Scaling RL Training}
Although DOT affects only a small fraction of rollouts within the all correct groups, it introduces subtle yet critical shifts in the training dynamics. 

\vspace{-4pt}
\paragraph{Structural Degeneration of the Reasoning Subspace.} Even rare interventions in the tail are associated with a rapid decline in policy entropy (\textit{Fig.}~\ref{fig:fig4}). This trend persists even when employing the asymmetric clipping strategy (Clip-Higher) from DAPO, suggesting that the observed decline is not merely an instance of indiscriminate entropy collapse. Instead, it reflects a structural degeneration of the reasoning subspace. Once redundant long tails are removed on already-solved prompts, the policy update aggressively optimizes for efficiency by rapidly driving the reasoning words that originally facilitated exploratory branching into a near-deterministic regime (see \textit{Fig.}~\ref{fig:fig7}). To counteract this behavior, we incorporate KL-Cov~\cite{KL-Cov} as a regularizer. This method identifies the specific subset of tokens exhibiting high covariance between their log-probabilities and advantage estimates, which serves as a statistical signature for aggressive policy shifts. By imposing a targeted KL penalty \(\mathbb{D}_{KL}(\pi_\theta \| \pi_{old})\) exclusively on these high-covariance tokens, KL-Cov effectively curbs the drastic updates that lead to premature determinism, thereby maintaining exploration stability without constraining the entire policy distribution.

\vspace{-4pt}
\paragraph{Sampling Inefficiency in Evolving Distributions.} When training stronger base models or scaling RL to longer schedules, more prompts gradually become all correct. Consequently, although DOT affects a small absolute number of responses, these truncated trajectories increasingly dominate the RL training dynamics (\textit{Fig.}~\ref{fig:fig5}). While previous work like DAPO addresses this via dynamic sampling, such approaches incur significant synchronization overheads and latency. In our context, the challenge is further compounded because DOT dynamically reactivates a subset of zero-gradient groups by re-computing rewards on truncated outputs, making the effective group ratio difficult to tune with fixed oversampling. To address this, we propose Predictive Dynamic Sampling. Instead of relying on costly iterative generation or wasteful fixed oversampling, we estimate the required oversampling factor based on the historical ratio of effective groups. This allows us to perform efficient single-round sampling that adapts automatically as the model improves and stabilizes the effective batch size. Formally, we present Predictive Dynamic Sampling with DOT in Algorithm~\ref{alg:pds} (Appendix~\ref{app:algo}).

%% file: figures/tex/fig4.tex
\begin{figure*}[t]
  \centering
  \includegraphics[width=\linewidth]{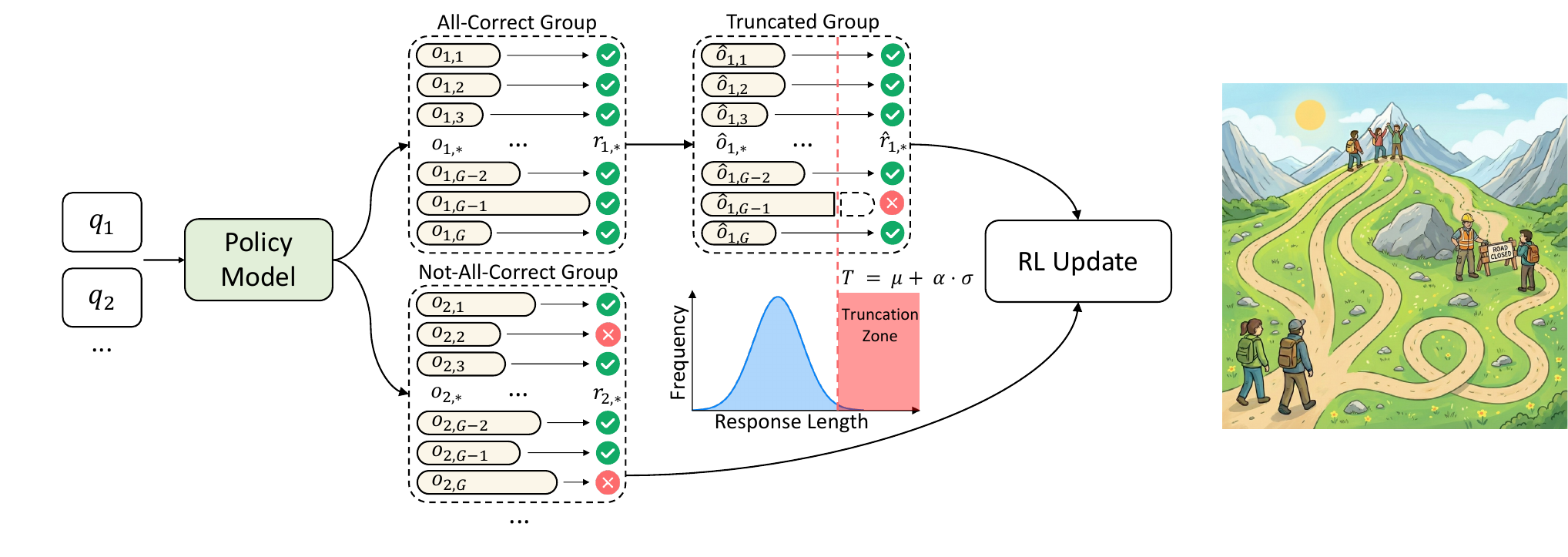}
  \vspace{-20pt}
  \caption{Illustration of Dynamic Outlier Truncation (DOT). By truncating statistical outliers in all-correct groups during GRPO-style RL, DOT creates a negative signal for redundancy without hindering exploration.}
  \vspace{-10pt}
  \label{fig:fig4}
\end{figure*}

%% file: figures/tex/fig5_6_7.tex
\begin{figure*}[t]
  \centering

  \begin{minipage}[t]{0.495\textwidth}
    \centering
    \includegraphics[width=\linewidth]{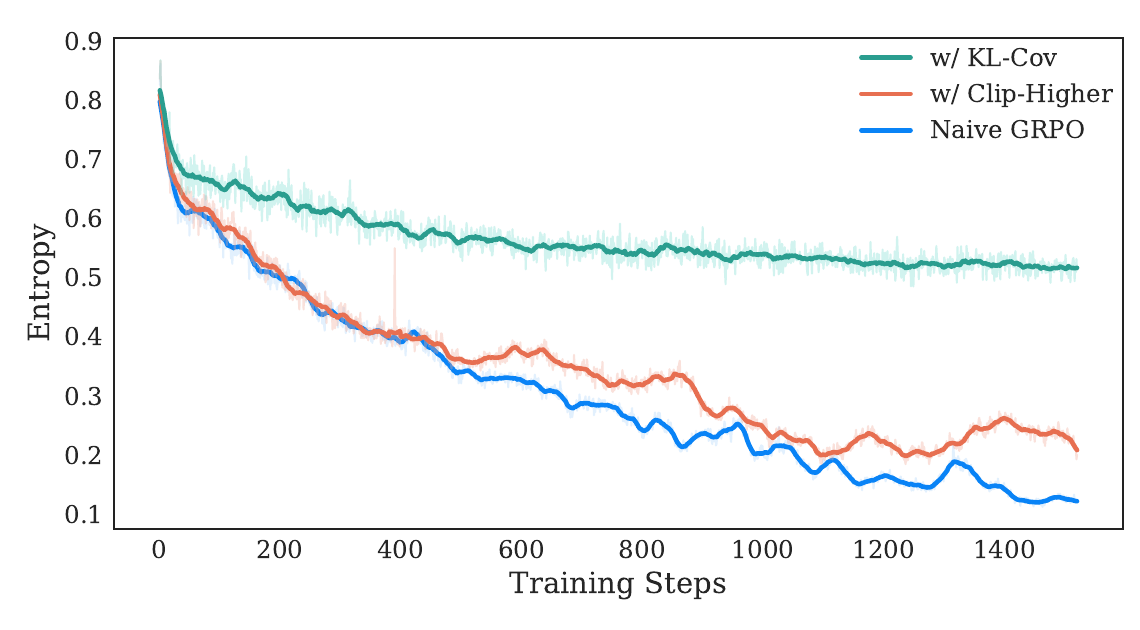}
    \captionsetup{width=0.95\linewidth}
    \vspace{-20pt}
    \captionof{figure}{Evolution of policy entropy during training.}
    \label{fig:fig5}
  \end{minipage}
  \begin{minipage}[t]{0.495\textwidth}
    \centering
    \includegraphics[width=\linewidth]{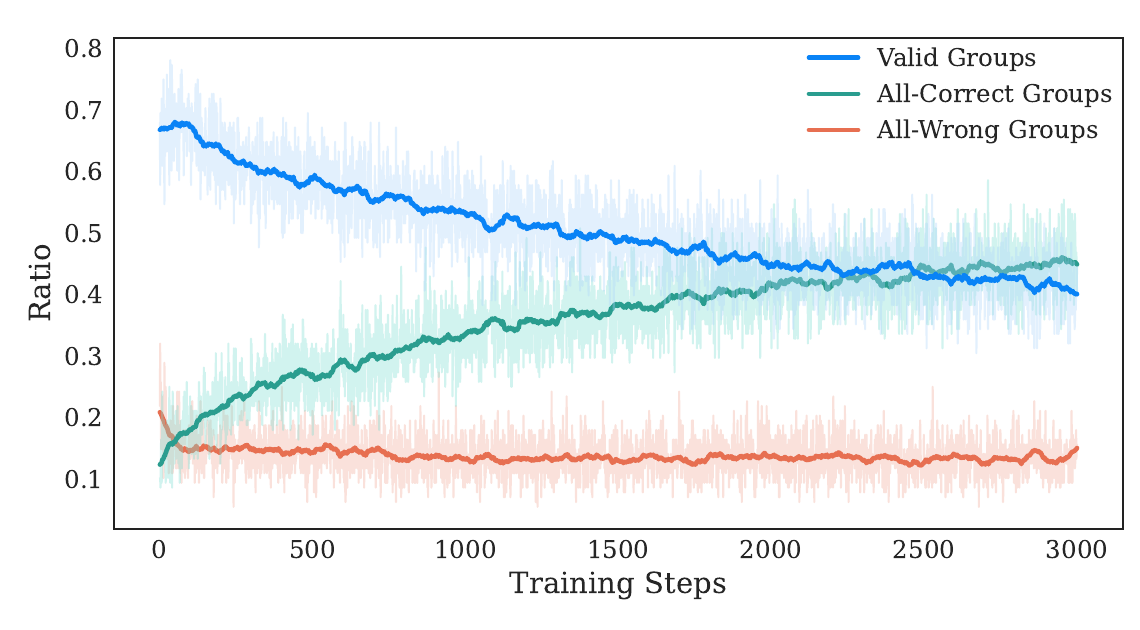}
    \captionsetup{width=0.9\linewidth}
    \vspace{-20pt}
    \captionof{figure}{Evolution of group ratios during training.}
    \label{fig:fig6}
  \end{minipage}

  \vspace{10pt}
  \begin{minipage}[t]{\textwidth}
    \centering
    \includegraphics[width=\linewidth]{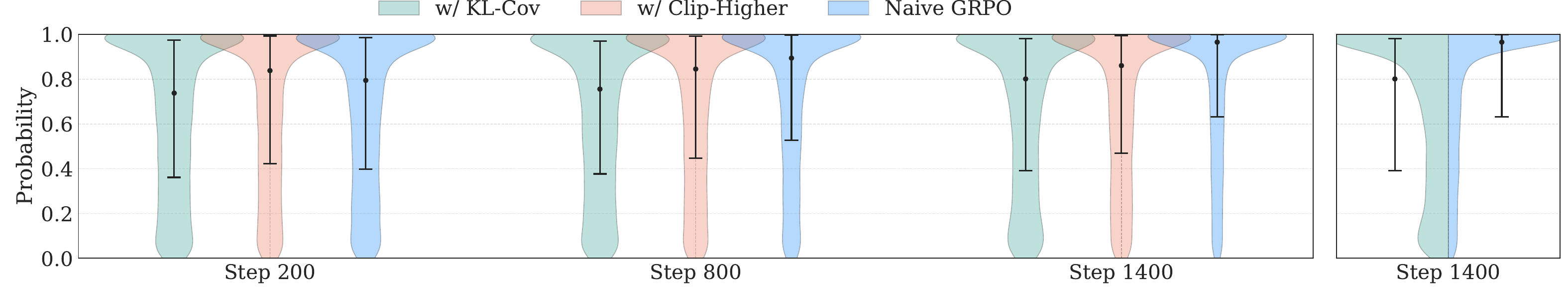}
    \captionsetup{width=\linewidth}
    \vspace{-20pt}
    \captionof{figure}{Distribution of observed sampling probabilities of reasoning words on AIME-24/25 during training.}
    \label{fig:fig7}
  \end{minipage}
  \vspace{-5pt}

\end{figure*}

%% file: sections/5_experiments.tex
\section{Experiments}
In this section, we introduce the experimental setup and evaluate both the performance and generation efficiency of our method on challenging reasoning benchmarks, with additional analyses and supplementary results provided in the appendix.

\input{tables/table1}
\input{tables/table2}

\subsection{Experimental Setup}

\paragraph{Datasets.}
We train our models on the dataset of DeepScaleR-Preview~\cite{deepscaler} and evaluate on four challenging math benchmarks:
AIME-24\footnote{\href{https://huggingface.co/datasets/Maxwell-Jia/AIME_2024}{HuggingFace dataset: Maxwell-Jia/AIME\_2024}},
AIME-25\footnote{\href{https://huggingface.co/datasets/yentinglin/aime_2025}{HuggingFace dataset: yentinglin/aime\_2025}},
AMC (AMC-22 and AMC-23)\footnote{\href{https://huggingface.co/datasets/AI-MO/aimo-validation-amc}{HuggingFace dataset: AI-MO/aimo-validation-amc}},
and MATH-500~\cite{math500}.
Following the official recommendation, we use the prompt template:
\texttt{``Please reason step by step, and put your final answer within \(\backslash\)boxed\{\}.''}
During evaluation, we decode with temperature \(t=0.6\), \(top\_p=0.95\) and \(top\_k=20\), and set the maximum generation budget to 32,768 tokens to avoid premature truncation.
To reduce variance, we sample 32 outputs per problem and report average pass@1 accuracy.
We adopt the Qwen-Math~\cite{qwen-math} evaluation tool to extract boxed final answers and compute accuracy, and additionally report the average generated length.

\paragraph{Baselines.}
We conduct experiments with the \texttt{DeepSeek-R1-Distill-Qwen} family as the base model, covering three scales (1.5B, 7B and 32B).
We compare DOT against recent methods that target reasoning performance and efficiency, including DeepScaleR-Preview, OverThink, DAST, O1-Pruner, LCR1, Laser, AdaptThink, DLER-R1, and SIRI.
For baselines with released checkpoints, we re-evaluate them under our unified protocol (same prompt template, decoding configuration, token budget, and the Qwen-Math evaluation tool) to ensure a fair comparison.
Detailed descriptions of these methods are provided in Appendix~\ref{app:baseline}.

\paragraph{Implementation Details.}
We instantiate DOT with a standard GRPO-based RL pipeline, and do not employ auxiliary tricks such as token-level loss or asymmetric clipping~\cite{dapo}.
Rewards are computed using the rule-based math verifier from ~\citet{deepscaler}.
Our implementation is built on verl~\cite{verl}, using FSDP~\cite{FSDP} for distributed training and SGLang~\cite{SGLang} to serve rollouts efficiently.
For policy optimization, following~\citet{SIRI}, we sample 32 rollouts per prompt, with a batch size of 128 and a mini-batch size of 32, i.e., the off-policy staleness is 4.
Unless otherwise stated, we cap the maximum training-time generation length at 8K tokens (DOT-4K is trained with a stricter 4K-token cap), setting the outlier threshold \(\alpha = 3\) in \textit{Eq.}~(\ref{eq:tq}) and the reduction margin \(m = 32\) in \textit{Eq.}~(\ref{eq:dot}).
Models are trained with a sampling temperature of 1.0 and a learning rate of \(1\times10^{-6}\).
For the KL-Cov regularization, we adopt the default parameter setting from~\citet{KL-Cov}. A complete list of training configurations is provided in Appendix~\ref{app:detailed_conf}.

\subsection{Main Results}

\paragraph{Significant Extension of the Pareto Frontier.} As presented in Table~\ref{tab:table1}, DOT consistently pushes the efficiency–performance Pareto frontier outward across varying model scales and benchmarks. In contrast to prior length-penalty methods (e.g., O1-Pruner, Laser) that often trade accuracy for brevity, or threshold-based methods (e.g., SIRI) that necessitate larger token budgets to sustain performance, DOT achieves substantial gains in both dimensions simultaneously. For instance, at the 1.5B scale, DOT-8K improves AIME-24 accuracy from 30.0\% (Original) to 52.2\%, while reducing the average response length by over 66\%. Notably, concurrent work DLER~\cite{DLER} shares our philosophy of avoiding explicit length-aware reward shaping, thus also achieving extreme length compression. However, by employing a dynamic truncation strategy that eliminates redundancy without stifling the exploration essential for complex reasoning, our DOT-4K achieves equivalent extreme brevity while further lifting accuracy by 7.3\%. Moreover, compared with the state-of-the-art method SIRI-high, DOT-8K achieves a 8.6\% absolute accuracy gain on AIME-24 while consuming only half the inference budget. This trend holds robustly on the 7B model, where DOT-8K establishes a new state-of-the-art with 62.6\% accuracy, surpassing the original model by 7.5\% while using only \(\sim\)37\% of the response length. Synthesizing results across all four benchmarks, DOT successfully realizes adaptive reasoning, i.e., allocating extended computation to challenging problems while aggressively pruning redundancy on simpler ones.

\paragraph{Scalability to Larger Models.}
The benefits of DOT generalize effectively to the 32B scale. DOT-8K consistently achieves superior accuracy across all benchmarks while reducing token consumption by \(\sim\)60\%. This substantial efficiency gain suggests that DOT enables larger models to exhibit a distinct scaling trend in ``per-token intelligence''.

\paragraph{Generalization to Out-of-Distribution Domains.}
DOT demonstrates robust transferability to out-of-distribution code generation tasks. As shown in Table~\ref{tab:table2}, our method consistently outperforms all baselines across both datasets. This indicates that DOT has internalized a universal pattern of concise reasoning, capable of adaptively allocating computational budget commensurate with the intrinsic difficulty of diverse problems.

\input{tables/table3}
\input{figures/tex/fig8}
\input{figures/tex/fig9}

\subsection{Ablation Study}
To investigate the contribution of each component in our training recipe, we conduct an ablation study as summarized in Table~\ref{tab:table3}. The most critical finding lies in the effectiveness of Dynamic Outlier Truncation. Removing this mechanism results in significantly longer responses while yielding almost identical accuracy, demonstrating that our method successfully mitigates overthinking while preserving reasoning capabilities. In contrast, omitting Group-Conditional Truncation causes a marked decline in performance, suggesting that applying a uniform threshold is detrimental to solving complex problems that necessitate extended CoTs. Regarding optimization objectives, incorporating the token-level loss into the GRPO leads to increased verbosity. This occurs because token-level aggregation inherently assigns larger gradient magnitude to longer responses, thereby exacerbating the model's tendency towards length inflation. Moreover, the removal of the KL-Cov or Predictive Dynamic Sampling negatively impacts both accuracy and efficiency, verifying their necessity in stabilizing the training process. Furthermore, we examine the hyperparameter of truncation threshold \(\alpha\) in \textit{Fig}~\ref{fig:fig8}. While moderate settings ensure stability, overly aggressive thresholds (e.g., \(\alpha=1.0, 2.0\)) trigger training collapse. This confirms that a three-sigma deviation serves as a statistically robust boundary for identifying genuine outliers, whereas tighter bounds risk truncating valid reasoning steps essential for convergence.

\subsection{Training Dynamics}
\textit{Fig.}~\ref{fig:fig9} shows the training evolution of DOT-8K, illustrating simultaneous gains in both reasoning accuracy and token efficiency.

\paragraph{Decoupling Length from Performance.}
Our method inverts the detrimental length shift. While benchmark accuracy steadily improves (\textit{Fig.}~\ref{fig:fig9}a, c), response length monotonically decreases (\textit{Fig.}~\ref{fig:fig9}b, d). Rising reward (\textit{Fig.}~\ref{fig:fig9}f) accompanied by reduced length verifies that the policy eliminates redundancy while preserving reasoning capacity.

\paragraph{Systemic Correction via Minimal Intervention.}
Substantial efficiency gains are achieved through minimal interventions. \textit{Fig.}~\ref{fig:fig9}h shows that DOT affects \(<0.5\%\) of generated responses, yet this suffices to drive a global length reduction (\textit{Fig.}~\ref{fig:fig9}e). By targeting only extreme outlier tails in all-correct groups, DOT corrects the shared policy prior responsible for redundancy rather than merely suppressing local symptoms.

\paragraph{Optimization Stability and Efficiency.}
Training remains robust against collapse, with policy entropy exhibiting a controlled decline (\textit{Fig.}~\ref{fig:fig9}g) to sustain exploration. Furthermore, \textit{Fig.}~\ref{fig:fig9}i demonstrates the precision of our predictive dynamic sampling. The effective batch size adheres strictly to the target with negligible variance. By adapting the oversampling rate to the improving pass rate, our predictive dynamic sampling eliminates the computational waste of fixed rate strategies of standard dynamic sampling.

%% file: tables/table1.tex
\begin{table*}[t]
\centering
\renewcommand{\arraystretch}{0.95}
\resizebox{0.88\textwidth}{!}{%
\begin{tabular}{lcccccccc}
\toprule
 & \multicolumn{2}{c}{AIME-24} & \multicolumn{2}{c}{AIME-25} & \multicolumn{2}{c}{AMC} & \multicolumn{2}{c}{MATH-500} \\
\cmidrule(lr){2-3} \cmidrule(lr){4-5} \cmidrule(lr){6-7} \cmidrule(lr){8-9}
\multirow{-2}{*}{Method} & Acc & Length & Acc & Length & Acc & Length & Acc & Length \\
\midrule
\multicolumn{9}{l}{\textit{DeepSeek-R1-Distill-Qwen-1.5B}} \\
\midrule
\textit{Original} & 30.0 & 15498 & 23.5 & 15604 & 64.1 & 10316 & 84.0 & 5483 \\
DeepScaleR-Preview~\cite{deepscaler} & 40.3 & 9430 & 30.2 & 9778 & 73.8 & 5538 & 88.9 & 3102 \\
OverThink$^*$~\cite{overthinking} & 28.3 & 11269 & — & — & — & — & 81.2 & 4131 \\
DAST$^*$~\cite{DAST} & 26.9 & 7745 & — & — & — & — & 83.0 & 2428 \\
O1-Pruner$^*$~\cite{o1-pruner} & 28.9 & 10361 & — & — & — & — & 82.2 & 3212 \\
LC-R1~\cite{LCR} & 22.9 & 8000 & 21.0 & 7961 & 60.7 & 4568 & 81.8 & 2362 \\
Laser-DE-L4096~\cite{LASER} & 32.6 & 8349 & 23.6 & 7839 & 67.5 & 4994 & 84.8 & 2763 \\
AdaptThink~\cite{AdaptThink} & 30.9 & 7917 & 23.3 & 8166 & 63.0 & 3710 & 82.5 & 1964 \\
DLER-R1~\cite{DLER} & 35.8 & \underline{3354} & 25.6 & \underline{3101} & 73.5 & \underline{2544} & 87.1 & 1777 \\
SIRI-low$^*$~\cite{SIRI} & 40.4 & 7093 & 29.6 & 6509 & 74.6 & 4700 & 87.7 & 2881 \\
SIRI-high$^*$~\cite{SIRI} & \underline{43.6} & 10049 & \underline{32.2} & 9739 & 75.9 & 7396 & 88.4 & 4633 \\
\rowcolor{rowblue} DOT-4K (Ours) & 43.1 & \textbf{3342} & 29.2 & \textbf{2979} & \underline{77.5} & \textbf{2281} & \underline{89.2} & \textbf{1249} \\
\rowcolor{rowblue} DOT-8K (Ours) & \textbf{52.2} & 5151 & \textbf{34.2} & 5143 & \textbf{80.6} & 3140 & \textbf{89.9} & \underline{1423} \\
\midrule
\multicolumn{9}{l}{\textit{DeepSeek-R1-Distill-Qwen-7B}} \\
\midrule
\textit{Original} & 55.1 & 13088 & 39.9 & 14240 & 82.5 & 7668 & 92.2 & 4026 \\
DAPO-DeepScaleR & \underline{57.6} & 9983 & 40.8 & 10705 & 84.5 & 6508 & 92.5 & 3658 \\
OverThink$^*$~\cite{overthinking} & 53.1 & 8744 & — & — & — & — & 89.4 & 2435 \\
DAST$^*$~\cite{DAST} & 45.6 & 7578 & — & — & — & — & 89.6 & 2162 \\
O1-Pruner$^*$~\cite{o1-pruner} & 49.2 & 9719 & — & — & — & — & 86.6 & 2534 \\
LC-R1~\cite{LCR} & 48.5 & 7580 & 35.6 & 7984 & 79.2 & 3765 & 90.1 & 1536 \\
Laser-DE-L4096~\cite{LASER} & 53.5 & 5890 & 37.4 & 6324 & 83.0 & 3381 & 92.6 & 1883 \\
AdaptThink~\cite{AdaptThink} & 55.2 & 10393 & 38.3 & 11723 & 81.5 & 5177 & 91.0 & 2008 \\
DLER-R1~\cite{DLER} & 50.6 & \underline{3241} & 33.6 & \underline{3357} & 83.5 & \underline{2262} & 92.4 & 1438 \\
SIRI-low$^*$~\cite{SIRI} & 56.1 & 6122 & 41.5 & 6386 & 85.8 & 4015 & 93.5 & 2452 \\
SIRI-high$^*$~\cite{SIRI} & 57.1 & 8585 & \underline{45.4} & 9106 & \underline{86.7} & 5773 & \underline{93.7} & 3378 \\
\rowcolor{rowblue} DOT-4K (Ours) & 54.8 & \textbf{2958} & 41.1 & \textbf{2835} & 86.1 & \textbf{1836} & 93.4 & \textbf{1008} \\
\rowcolor{rowblue} DOT-8K (Ours) & \textbf{62.6} & 4903 & \textbf{48.5} & 5464 & \textbf{87.6} & 2779 & \textbf{94.3} & \underline{1293} \\
\midrule
\multicolumn{9}{l}{\textit{DeepSeek-R1-Distill-Qwen-32B}} \\
\midrule
\textit{Original} & \underline{72.4} & 10299 & \underline{56.0} & 12385 & \underline{88.9} & 6578 & 94.3 & 3557 \\
Laser-DE-L8192$^*$~\cite{LASER} & 70.8 & 6785 & — & — & — & — & 93.2 & 2314 \\
\rowcolor{rowblue} DOT-4K (Ours) & 65.3 & \textbf{2622} & 52.5 & \textbf{2782} & 87.4 & \textbf{1472} & \underline{94.5} & \textbf{861} \\
\rowcolor{rowblue} DOT-8K (Ours) & \textbf{73.2} & \underline{4151} & \textbf{59.6} & \underline{5301} & \textbf{90.6} & \underline{2786} & \textbf{95.0} & \underline{1369} \\
\bottomrule
\end{tabular}%
}
\vspace{-2pt}
\caption{Performance comparison on AIME-24, AIME-25, AMC, and MATH-500 benchmarks. We report pass@1 accuracy (\%) and average response length (tokens). (For methods marked with an asterisk, we cite results from~\citet{SIRI} and~\citet{LASER} as public checkpoints are unavailable.)}
\vspace{-6pt}
\label{tab:table1}
\end{table*}

%% file: tables/table2.tex
\begin{table}[t]
\centering
\renewcommand{\arraystretch}{0.925}
\resizebox{\columnwidth}{!}{%
\begin{tabular}{lcccc}
\toprule
 & \multicolumn{2}{c}{HumanEval} & \multicolumn{2}{c}{LiveCodeBench} \\
\cmidrule(lr){2-3} \cmidrule(lr){4-5}
\multirow{-2}{*}{Method} & Acc & Length & Acc & Length \\
\midrule
\multicolumn{5}{l}{\textit{DeepSeek-R1-Distill-Qwen-1.5B}} \\
\midrule
Original & 64.7 & 4377 & 16.4 & 13706 \\
DeepScaleR-Preview & 69.6 & 4657 & 21.0 & 10076 \\
LC-R1 & 59.8 & 2814 & 15.1 & 11128 \\
Laser-DE-L4096 & 64.5 & 2372 & 17.5 & 6223 \\
AdaptThink & 64.4 & 3859 & 17.7 & 11117 \\
DLER-R1 & 68.2 & \underline{2350} & 20.8 & \textbf{4132} \\
\rowcolor{rowblue} DOT-4K (Ours) & \underline{70.5} & \textbf{2306} & \underline{21.7} & \underline{4481} \\
\rowcolor{rowblue} DOT-8K (Ours) & \textbf{70.7} & 2860 & \textbf{22.6} & 6903 \\
\midrule
\multicolumn{5}{l}{\textit{DeepSeek-R1-Distill-Qwen-7B}} \\
\midrule
Original & 81.9 & 3265 & 31.8 & 9718 \\
LC-R1 & 81.2 & 2173 & 31.4 & 6634 \\
Laser-DE-L4096 & 82.9 & 2118 & \underline{33.0} & 6051 \\
AdaptThink & 81.6 & 2862 & 32.2 & 8767 \\
DLER-R1 & 82.9 & 2118 & \underline{33.0} & 6050 \\
\rowcolor{rowblue} DOT-4K (Ours) & \underline{85.0} & \textbf{1474} & \underline{33.0} & \textbf{3988} \\
\rowcolor{rowblue} DOT-8K (Ours) & \textbf{85.1} & \underline{2019} & \textbf{34.8} & \underline{5979} \\
\bottomrule
\end{tabular}%
}
\vspace{-5pt}
\caption{Performance comparison on code generation benchmarks (HumanEval~\cite{human-eval} and LiveCodeBench(v6)~\cite{LCB}).}
\vspace{-15pt}
\label{tab:table2}
\end{table}

%% file: tables/table3.tex
\begin{table}[t]
\centering
\renewcommand{\arraystretch}{0.95}
\resizebox{\columnwidth}{!}{%
\begin{tabular}{lcccc}
\toprule
 & \multicolumn{2}{c}{AIME-24} & \multicolumn{2}{c}{AMC} \\
\cmidrule(lr){2-3} \cmidrule(lr){4-5}
\multirow{-2}{*}{Variant} & Acc & Length & Acc & Length \\
\midrule
DOT-8K & 52.2 & 5151 & 80.6 & 3140 \\
\midrule
w/o Dynamic Outlier Truncation & 51.5 & 6879 & 81.3 & 4851 \\
w/o Group-Conditional Truncation & 47.8 & 5071 & 79.7 & 3014 \\
w/o Minimum Truncation Margin & 50.0 & 5696 & 77.3 & 2902 \\
w/o KL-Cov & 47.9 & 6057 & 79.3 & 3856 \\
w/o Predictive Dynamic Sampling & 48.9 & 5637 & 79.1 & 3576 \\
w/ Token-Level Loss & 49.5 & 6334 & 79.4 & 3964 \\
\bottomrule
\end{tabular}%
}
\vspace{-5pt}
\caption{Performance comparison of different variants on \texttt{DeepSeek-R1-Distill-Qwen-1.5B}.}
\vspace{-15pt}
\label{tab:table3}
\end{table}

%% file: figures/tex/fig8.tex
\begin{figure*}[t]
  \centering
  \includegraphics[width=\linewidth]{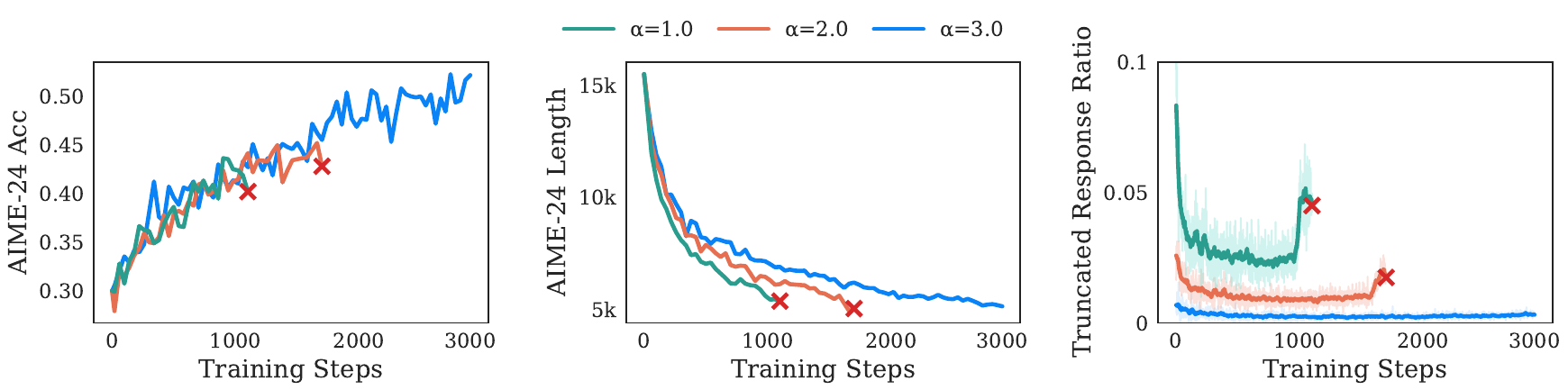}
  \vspace{-20pt}
  \caption{Impact of the hyperparameter of truncation threshold \(\alpha\). We report the curves for pass@1 accuracy, average response length, and the ratio of truncated responses. The `X' markers indicate training collapse.}
  \vspace{-5pt}
  \label{fig:fig8}
\end{figure*}

%% file: figures/tex/fig9.tex
\begin{figure*}[t]
  \centering
  \includegraphics[width=\linewidth]{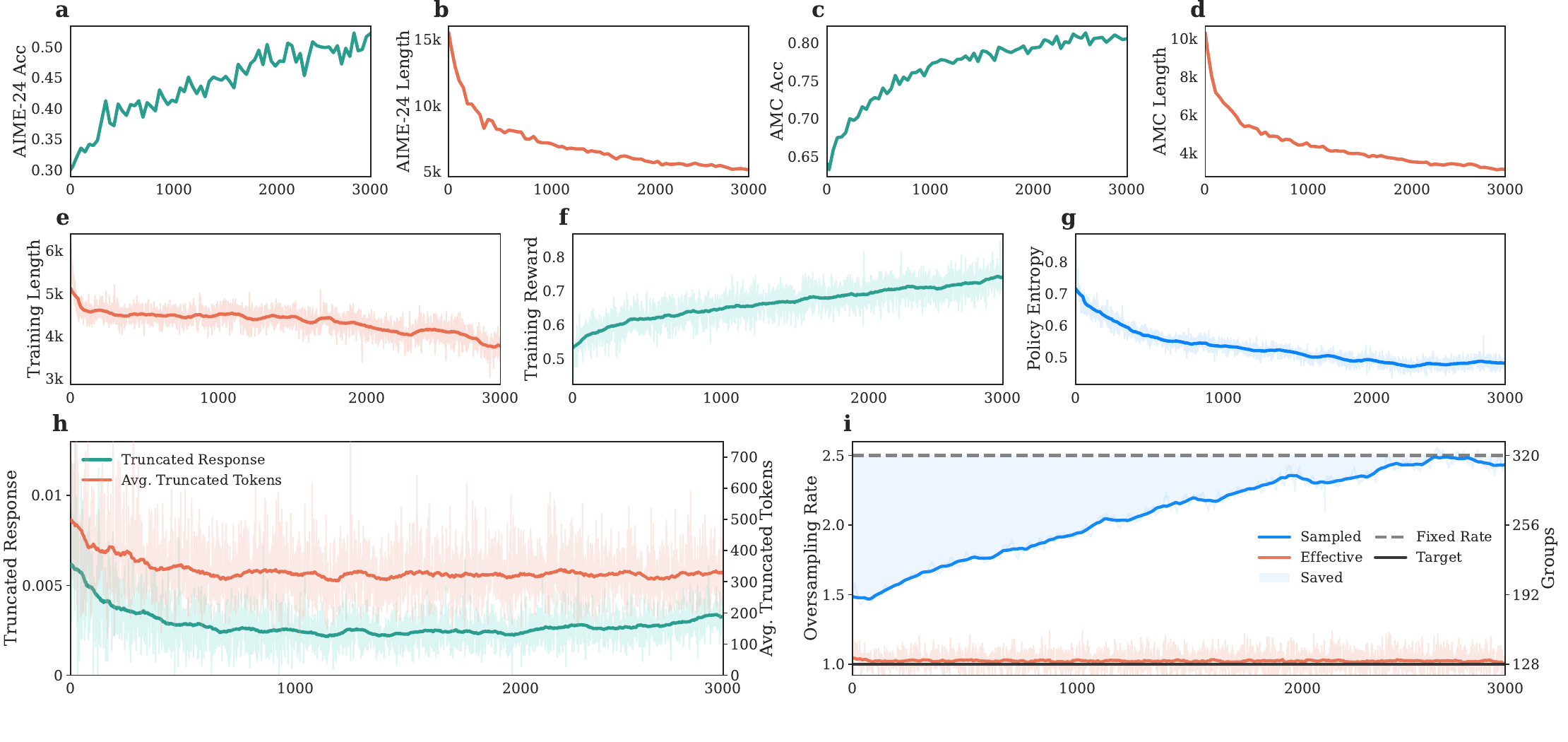}
  \vspace{-30pt}
  \caption{Metric curves monitoring the DOT-8K training process on \texttt{DeepSeek-R1-Distill-Qwen-1.5B.}}
  \vspace{-10pt}
  \label{fig:fig9}
\end{figure*}

%% file: sections/6_related_work.tex
\section{Related Work}
Recent reasoning models have demonstrated exceptional problem-solving abilities through extended CoT~\cite{tokenselect,MMEA,LEMMA,SEPIT}. However, they frequently exhibit the ``overthinking'' phenomenon~\cite{overthinking}, incurring high computational costs on simple tasks. Current research addresses this through three primary paradigms: reasoning compression, adaptive routing, and length-aware reinforcement learning.
\paragraph{Reasoning Compression and Model Merging.}
To reduce inference latency, compression methods transfer reasoning capabilities to compact formats. Beyond standard distillation, approaches like TokenSkip~\cite{TokenSkip} and Chain of Draft~\cite{CoD} prune semantic redundancies. Further pushing compression, CODI~\cite{CODI} maps explicit reasoning steps into continuous latent representations. Alternatively, model merging techniques, such as Kimi k1.5 \cite{kimi-k1.5} and CoT-Valve~\cite{CoT-Valve}, interpolate weights between reasoning and non-reasoning models to balance verbosity without extensive retraining.

\paragraph{Adaptive Routing and Mode Switching.}
Dynamic routing frameworks allocate compute based on query complexity. Systems like ThinkSwitcher \cite{ThinkSwitcher} utilize lightweight classifiers to toggle between ``Fast'' and ``Slow'' thinking modes. Similarly, Thinkless~\cite{Thinkless} and LHRMs~\cite{HRMS} train models to autonomously determine the necessity of reasoning via special control tokens.

\paragraph{RL with Length-Aware Reward Shaping.}
RL has emerged as a pivotal paradigm for refining reasoning efficiency. Early approaches like L1~\cite{L1} and O1-Pruner~\cite{o1-pruner} impose explicit length penalties to harmonize token budgets. To address structural inefficiencies, LC-R1~\cite{LCR} employs dual rewards to prune invalid verification steps, while DLER~\cite{DLER} optimizes truncation penalties to prevent training collapse. Recognizing that optimal length varies by complexity, DAST~\cite{DAST} and Laser~\cite{LASER} introduce difficulty-aware shaping, dynamically adjusting penalties to tolerate longer chains for complex queries. Other works explore autonomous mode switching~\cite{AdaptThink} or interleaved training schedules~\cite{SIRI} to optimize the trade-off.

In this work, we focus on RL-based efficient reasoning methods. We further analyze the mechanisms that trigger overthinking and propose a targeted training recipe that significantly pushes the Pareto frontier of reasoning efficiency.




%% file: sections/7_conclusion.tex
\section{Conclusion}
\vspace{-4pt}
In this paper, we identified length shift as the primary driver of redundancy in reasoning models, revealing how uncertainty-driven exploration on complex problems inadvertently causes verbosity on trivial inputs. To address this, we introduced Dynamic Outlier Truncation (DOT), a training-time intervention that selectively eliminates statistical outliers in all-correct groups. By complementing this mechanism with entropy stabilization and predictive dynamic sampling, DOT achieves a significant reduction in response length while preserving, and often enhancing accuracy. Our results across multiple scales demonstrate that DOT effectively pushes the efficiency--performance Pareto frontier outward, offering a simple, robust and scalable paradigm for training efficient reasoning models.

\section*{Limitations}
Despite significantly extending the efficiency-performance Pareto frontier, our study acknowledges certain limitations. 
As a training-time intervention rooted in RL, the effectiveness of DOT is inherently bounded by the quality of the training data and the initial policy. 
Applying DOT to further refine state-of-the-art reasoning models (e.g., \texttt{DeepSeek-V3.2}~\cite{deepseek-v3.2}, \texttt{Qwen3-235B-A22B-Thinking}~\cite{qwen}) poses challenges, as these models have typically undergone extensive post-training on datasets vastly exceeding the scale of open-source counterparts. 
Consequently, they exhibit extremely low policy entropy on open-source datasets, making it nearly impossible to extract further gains using current public resources. 
Nevertheless, given DOT's simplicity and plug-and-play nature, it holds substantial promise for application in the post-training phase of next-generation foundation models, particularly when leveraged with high-quality in-house data or integrated with test-time reinforcement learning strategies~\cite{ttrl}.

Furthermore, our current investigation does not yet encompass agentic tasks. 
The core principle of length shift likely extends to these scenarios, manifesting as redundant tool invocations or cyclic planning steps. 
Although we leave this exploration for future work, extending the DOT mechanism to prune redundancy in agent trajectories, thereby optimizing action spaces rather than merely token spaces, represents a critical necessity for building efficient and scalable agents.

\section*{Acknowledgments}
This work was supported in part by the National Natural Science Foundation of China (Grant No. 92370204), in part by the National Key R\&D Program of China (Grant No. 2023YFF0725001), in part by the Guangdong Basic and Applied Basic Research Foundation (Grant No. 2023B1515120057), and in part by the Key-Area Special Project of Guangdong Provincial Ordinary Universities (2024ZDZX1007).

%% file: sections/8_appendix.tex
\input{tables/algo1}
\input{tables/tablea}

\newpage

\section{Formal Description of Algorithms}
\label{app:algo}
In \textit{Sec.}~\ref{sec:methodology}, we propose Predictive Dynamic Sampling, which estimates the required oversampling factor based on historical pass rates to stabilize the effective batch size. Formally, the detailed algorithm is presented in Algorithm~\ref{alg:pds}.
\section{Case Study and Detailed Analysis}
\label{app:case}
\input{figures/tex/figa}
\input{figures/tex/figb}

In this section, we provide both qualitative examples and quantitative statistical analyses to further illustrate the impact of our approach.

\textit{Fig.}~\ref{fig:case_aime24} through~\ref{fig:case_livecodebench} present a series of side-by-side case studies across multiple benchmarks, comparing the reasoning trajectories of the original long-CoT policy with our DOT-optimized model. These cases consistently demonstrate that while the original model often falls into redundant verification loops and overthinking, our model arrives at the same correct answer through a significantly more concise and purposeful path without sacrificing logical rigor.

As a supplement to these qualitative case studies, \textit{Fig.}~\ref{fig:figb1} through~\ref{fig:figb3} provide a macro-statistical perspective on the response length distributions and differences across various model scales and datasets. Our analysis shows that the response length ratio (Original / Ours) remains consistently high across different difficulty levels, confirming that the efficiency gains of DOT are not limited to simple queries but remain effective across the entire complexity spectrum. Furthermore, the scatter plots reveal a strong correlation between the original policy's response length/variance and the resulting token savings. This indicates that DOT is particularly effective at optimizing policies that are inherently redundant, guiding the model toward a more efficient reasoning distribution during training while maintaining stable and high-quality performance at inference time.

\section{Orthogonality with Advanced \\\quad\;\; Policy Optimization Algorithms}
\label{app:orthogonality}
To investigate the orthogonality of DOT with state-of-the-art policy optimization algorithms, we integrated it with representative algorithmic enhancements~\cite{gmpo,gspo} involving entropy control and importance-sampling estimation. The results in Table~\ref{tab:tablea} demonstrate that DOT, functioning as an intervention on generated rollouts, synergizes effectively with these algorithm-level improvements. Notably, this combination simultaneously amplifies performance and token efficiency, highlighting the substantial potential of DOT to synergize with advanced optimization techniques to drive the next frontier of efficient and scalable reasoning.

\input{tables/tabled}
\input{tables/tablee}
\section{Additional Generalization Results}
\label{app:additional_generalization}

Beyond the code generation results in Table~\ref{tab:table2}, we further evaluate DOT on general reasoning benchmarks~\cite{MMLU,GPQA} and open-ended evaluation suites~\cite{Arena-Hard,AlpacaEval} using \texttt{DeepSeek-R1-Distill-Qwen-1.5B}. As shown in Tables~\ref{tab:tabled} and~\ref{tab:tablee}, although DOT is trained exclusively on verifiable math data, it generalizes effectively beyond the training domain, substantially reducing response length while preserving or slightly improving performance. This indicates that DOT does not merely exploit benchmark-specific answer formats, but instead encourages a more concise and efficient reasoning style that transfers robustly to broader tasks.

\input{tables/tablef}
\section{Results on a More Recent Base Model}
\label{app:qwen3_results}

To verify that DOT is not tied to the DeepSeek-R1-Distill family, we further evaluate it on the more recent base model \texttt{Qwen3-4B}~\cite{qwen}. As shown in Table~\ref{tab:tablef}, DOT exhibits the same qualitative trend, consistently reducing response length while maintaining or improving reasoning performance across multiple math benchmarks. This suggests that DOT captures a general training phenomenon rather than an idiosyncrasy of a specific base model family.

\input{tables/tableg}
\section{Agentic Evaluation}
\label{app:agentic_eval}

A natural concern is whether suppressing redundant long tails may harm the exploration capacity required in agentic settings. To examine this, we further evaluate DOT on BFCL v3~\cite{BFCL} using \texttt{Qwen3-4B}~\cite{qwen}. As shown in Table~\ref{tab:tableg}, DOT preserves the overall task score and the average number of steps while substantially reducing response length. This suggests that DOT removes habituated verbosity without disrupting the effective decision-making structure required for multi-step agentic reasoning.

\input{tables/tableh}
\section{Comparison with Training-Free Methods}
\label{app:training_free}

Although DOT is a training-based method, it is still informative to compare it with recent training-free efficient reasoning methods under a unified evaluation protocol. As shown in Table~\ref{tab:tableh}, DOT achieves a more favorable efficiency--performance trade-off than these inference-time baselines, reaching substantially stronger benchmark performance at comparable or much shorter response lengths.

\input{figures/tex/figc}

\section{Additional Training Dynamics} 
\label{app:training_dynamics}
\textit{Fig.}~\ref{fig:figc1} through~\ref{fig:figc5} provide supplementary visualizations of the training evolution across various model scales (1.5B, 7B, 32B) and hyperparameter settings. These results consistently demonstrate that our training recipe remains robust and stable under diverse conditions. Across all settings, the policy successfully decouples reasoning performance from response length, achieving a steady increase in accuracy while simultaneously eliminating redundant verbosity.

\section{Detailed Descriptions on Baselines}
\label{app:baseline}
To comprehensively evaluate the effectiveness of our proposed method, we compare it against a diverse set of state-of-the-art baselines:

\begin{itemize}[leftmargin=*]
    \item \textbf{DeepScaleR-Preview}~\cite{deepscaler}: A strong unconstrained baseline that enhances reasoning capabilities through iterative context lengthening (scaling from 8K to 24K tokens) during reinforcement learning.

    \item \textbf{OverThink}~\cite{overthinking}: A data-centric approach that addresses the ``overthinking'' phenomenon. It employs efficiency metrics to identify and filter out non-essential steps (such as redundant post-answer verification) to construct concise supervised fine-tuning data.

    \item \textbf{DAST}~\cite{DAST}: A difficulty-adaptive framework that introduces a Token Length Budget (TLB) metric. It dynamically adjusts reward shaping during RL to penalize verbosity on simple queries while tolerating longer reasoning chains for complex problems.

    \item \textbf{O1-Pruner}~\cite{o1-pruner}: An off-policy optimization method that introduces a length-harmonizing reward. It aligns the student model's length distribution with a concise reference model while penalizing accuracy degradation to reduce inference overhead.

    \item \textbf{LC-R1}~\cite{LCR}: A structural compression framework utilizing a dual-reward mechanism. It combines a global length penalty with a specific ``compress reward'' to target and prune invalid thinking loops or redundant self-verifications.

    \item \textbf{Laser-DE}~\cite{LASER}: A dynamic length-based reward shaping method. The DE variant (Dynamic \& Exploration) specifically encourages exploration by relaxing penalties for incorrect responses while strictly constraining correct ones to be concise.

    \item \textbf{AdaptThink}~\cite{AdaptThink}: A mode-switching framework that trains the model to explicitly select between ``Thinking'' (long CoT) and ``NoThinking'' (direct answer) modes based on the estimated complexity of the input query.

    \item \textbf{DLER-R1}~\cite{DLER}: An optimization-focused approach that refines simple truncation penalties. It mitigates reward collapse issues in RL through batch-wise reward normalization and dynamic sampling strategies.

    \item \textbf{SIRI}~\cite{SIRI}: A training scheduling method that alternates between compression and expansion phases. We compare against both \textbf{SIRI-low} and \textbf{SIRI-high} checkpoints to evaluate performance at different points along its efficiency-accuracy Pareto frontier.
\end{itemize}

\input{tables/tableb}
\section{Detailed Training Configurations}
\label{app:detailed_conf}
We conduct all experiments on NVIDIA H800 and H20 GPUs. For 1.5B models, we use either 16$\times$H800 or 32$\times$H20; for 7B models, we use 32$\times$H800 or 64$\times$H20; and for 32B models, we use 64$\times$H800 or 128$\times$H20. The typical wall-clock training time reported here is measured on H800: approximately 7 days for 1.5B models, and around 10 days for both 7B and 32B models; runs on H20 are generally slower. The detailed training configurations are summarized in Table~\ref{tab:tableb}.

\input{tables/tablei}
\section{Training Efficiency and Latency}
\label{app:latency}

Beyond inference-time compression, DOT also improves the physical efficiency of RL training. By suppressing excessively long tails, DOT shortens rollout time and reduces idle waiting caused by overlong generations, thereby improving overall wall-clock throughput. As shown in Table~\ref{tab:tablei}, DOT substantially reduces both rollout latency and average time per step while maintaining stable training dynamics.

\input{tables/tablec}
\section{List of Reasoning Words}
\label{app:reasoning_words}
In \textit{Sec.}~\ref{sec:empirical} and \textit{Sec.}~\ref{sec:methodology}, we analyzed the behavioral propensity of the reasoning model to emit specific lexical markers to investigate the length shift phenomenon. These markers typically signal the onset of reasoning steps, self-correction, or verification. Table~\ref{tab:tablec} provides the comprehensive list of these reasoning words.

\section{Use of AI Assistants}

In this paper, AI assistants are used for literature retrieval and grammar checking.

%% file: tables/algo1.tex
\begin{algorithm*}[t]
\caption{Predictive Dynamic Sampling (with DOT)}
\label{alg:pds}

\KwIn{training prompts \( \mathcal{D} \); target prompt batch size \( B \); rollout group size \( G \); history window \( W \)}
\KwInit{history buffer \( \mathcal{H} \) storing recent effective group ratio}

\For{\( \mathrm{step}=1,\ldots,M \)}{
    \( \bar{p}\leftarrow \mathrm{mean}(\mathcal{H}) \), \( s_p\leftarrow \mathrm{std}(\mathcal{H}) \) \tcp*[r]{effective group ratio stats}
    \( \gamma \leftarrow \max\!\left(1.0,\ \frac{1}{\bar{p}}(1+s_p)\right) \) \tcp*[r]{predicted oversampling rate}
    Sample a prompt batch \( \mathcal{B}\subset\mathcal{D} \) with \( |\mathcal{B}|=\lceil \gamma B\rceil \) \;
    Sample \( G \) rollouts for each \( q\in\mathcal{B} \) and compute rewards \;
    Apply DOT and recompute rewards \;
    \( \mathcal{B}_{\mathrm{eff}} \leftarrow \{q\in\mathcal{B}\mid \mathrm{std}(\{R_i(q)\}_{i=1}^{G})>0\} \) \tcp*[r]{keep non-zero std groups}
    \( \mathcal{U}\ \leftarrow \) Assemble exactly \( B \) groups from \( \mathcal{B}_{\mathrm{eff}} \) by prompt-level dropping and masked padding \;
    Execute an RL update on \( \mathcal{U} \) \;
    \( p \leftarrow |\mathcal{B}_{\mathrm{eff}}|/|\mathcal{B}| \) \tcp*[r]{effective group ratio}
    \( \mathcal{H} \leftarrow \mathrm{Tail}_{W}\!\left(\mathcal{H} \oplus [p]\right) \) \tcp*[r]{sliding-window update}
}

\KwOut{trained policy model}
\end{algorithm*}

%% file: tables/tablea.tex
\begin{table*}[t]
    \centering
    \renewcommand{\arraystretch}{0.95}
    \resizebox{0.88\textwidth}{!}{%
    \begin{tabular}{lcccccccc}
        \toprule
         & \multicolumn{2}{c}{AIME-24} & \multicolumn{2}{c}{AIME-25} & \multicolumn{2}{c}{AMC} & \multicolumn{2}{c}{MATH-500} \\
        \cmidrule(lr){2-3} \cmidrule(lr){4-5} \cmidrule(lr){6-7} \cmidrule(lr){8-9}
        \multirow{-2}{*}{Method} & Acc & Length & Acc & Length & Acc & Length & Acc & Length \\
        \midrule
        DOT-8K & 52.2 & 5151 & 34.2 & 5143 & 80.6 & 3140 & 89.4 & 1423 \\
        w/o KL-Cov~\cite{KL-Cov} & 47.9 & 6057 & 33.3 & 6029 & 79.3 & 3856 & 90.5 & 1932 \\
        \hspace{1em} w/ GMPO~\cite{gmpo} & 48.1 & 5745 & 34.3 & 5814 & 80.5 & 3550 & 90.7 & 1762 \\
        \hspace{1em} w/ GSPO~\cite{gspo} & 52.7 & 5007 & 37.1 & 5006 & 81.2 & 3163 & 90.6 & 1565 \\
        \bottomrule
    \end{tabular}%
    }
    \caption{Performance comparison on \texttt{DeepSeek-R1-Distill-Qwen-1.5B}. We examine the integration of DOT with advanced policy optimization strategies, where KL-Cov is explicitly disabled.}
    \label{tab:tablea}
\end{table*}

%% file: figures/tex/figa.tex
\makeatletter
\providecommand{\captionof}[2]{\begingroup\def\@captype{#1}\caption{#2}\endgroup}
\makeatother

\definecolor{caseQHeader}{HTML}{7A7A7A}
\definecolor{caseQFrame}{HTML}{7A7A7A}
\definecolor{caseQBack}{HTML}{F2F2F2}
\definecolor{caseOursHeader}{HTML}{AED6FC}
\definecolor{caseOursFrame}{HTML}{0983F6}
\definecolor{caseBaseHeader}{HTML}{7EDDD2}
\definecolor{caseBaseFrame}{HTML}{2A9D8F}
\definecolor{caseOmit}{HTML}{E76F51}

\providecommand{\CaseOmit}[1]{\par\noindent\textcolor{caseOmit}{\textit{... (omitted: #1)}}\par}
\providecommand{\CaseAnswerBox}[1]{\begin{center}\fbox{\strut #1}\end{center}}
\providecommand{\CaseGTBox}[1]{\begin{center}\fbox{\strut #1}\end{center}}

\tcbset{
  casestudy/base/.style={
    enhanced,
    boxrule=0.5mm,
    arc=2mm,
    left=1.5mm,right=1.5mm,top=1.5mm,bottom=1.5mm,
    colback=white,
    coltitle=black,
    fonttitle=\bfseries\small,
    fontupper=\scriptsize,
    before skip=0pt,
    after skip=0pt,
    before upper={\setlength{\abovedisplayskip}{3pt}\setlength{\belowdisplayskip}{3pt}\setlength{\parskip}{0.3em}},
  },
  casestudy/question/.style={
    casestudy/base,
    colback=caseQBack,
    colframe=caseQFrame,
    colbacktitle=caseQHeader,
    coltitle=white,
  },
  casestudy/ours/.style={
    casestudy/base,
    colframe=caseOursFrame,
    colbacktitle=caseOursHeader,
    coltitle=black,
  },
  casestudy/baseline/.style={
    casestudy/base,
    colframe=caseBaseFrame,
    colbacktitle=caseBaseHeader,
    coltitle=black,
  },
}

\begin{figure*}[t]
    \centering
    \begin{minipage}[t]{0.48\textwidth}
        \vspace{0pt}
        
        \begin{tcolorbox}[casestudy/question, title={\textbf{Question and Ground Truth}}]
            \textbf{Question:} There exist real numbers \(x\) and \(y\), both greater than 1, such that \(\log_x\left(y^x\right)=\log_y\left(x^{4y}\right)=10\). Find \(xy\).
            
            \vspace{0.5em}
            \textbf{Ground Truth:} \CaseGTBox{25}
        \end{tcolorbox}
        
        \vspace{0.3cm}
        
        \begin{tcolorbox}[casestudy/ours, title={\textbf{Response from DOT}\hfill \textbf{959 Tokens}}]
            Okay, so we have this problem where we need to find real numbers \(x\) and \(y\), both greater than 1, such that:
            \[ \log_x(y^x) = 10 \quad \text{and} \quad \log_y(x^{4y}) = 10 \]
            We need to find \(xy\).
            
            Hmm. Let me rewrite these logarithms using logarithm properties.
            First equation: \(\log_x(y^x) = 10\). Using the power rule, log base \(x\) of \((y^x)\) is \(x \cdot \log_x(y)\). So:
            \[ x \cdot \log_x(y) = 10 \]
            Similarly, \(\log_x(y)\) can be written as \(\ln y / \ln x\), so:
            \[ x \cdot (\ln y / \ln x) = 10 \implies (x / \ln x) \cdot \ln y = 10 \]
            Let me note that as equation (1).
            
            Now the second equation: \(\log_y(x^{4y}) = 10\). Again, using power rule: \(4y \cdot \log_y(x) = 10\).
            Similarly, \(\log_y(x)\) is \(\ln x / \ln y\), so:
            \[ 4y \cdot (\ln x / \ln y) = 10 \implies (4y / \ln y) \cdot \ln x = 10 \]
            Let me note that as equation (2).
            
            So now we have two equations:
            \[ 1)\; (x / \ln x) \cdot \ln y = 10 \quad 2)\; (4y / \ln y) \cdot \ln x = 10 \]
            From equation (1): \(\ln y = (10 \cdot \ln x) / x\). Let me call this equation (1a).
            From equation (2): \((4y / \ln y) \cdot \ln x = 10\).
            
            Let me rearrange equation (2):
            \[ 4y \cdot (\ln x / \ln y) = 10 \]
            But notice that \((\ln x / \ln y)\) is equal to \(\log_y x\), which is reciprocal of \(\log_x y\).
            Wait, maybe substitute \(\ln y\) from equation (1a).
            From equation (1a): \(\ln y = (10 \ln x)/x\).
            Therefore, \(\ln y = (10 / x) \cdot \ln x\).
            
            So, let's compute \(\ln x / \ln y\):
            \[ \frac{\ln x}{\ln y} = \frac{\ln x}{[(10 / x) \cdot \ln x]} = \frac{\ln x}{(10 \ln x)/x} = \ln x \cdot \frac{x}{10 \ln x} = \frac{x}{10} \]
            So, \(\ln x / \ln y = x / 10\).
            
            Therefore, equation (2): \(4y \cdot (\ln x / \ln y) = 10\).
            Substitute \(\ln x / \ln y = x / 10\):
            \[ 4y \cdot (x / 10) = 10 \]
            So: \((4xy)/10 = 10 \implies 4xy = 100 \implies xy = 25\).
            Therefore, the answer is 25.
            
            \vspace{0.5em}
            \CaseAnswerBox{\textbf{25}}
        \end{tcolorbox}
    \end{minipage}
    \hfill
    \begin{minipage}[t]{0.48\textwidth}
        \vspace{0pt}
        
        \begin{tcolorbox}[casestudy/baseline, title={\textbf{Response from R1-Distill-Qwen-1.5B}\hfill \textbf{22514 Tokens}}]
            Alright, so I have this problem here... I remember that logarithms can be rewritten using the power rule.
            Starting with the first equation: \(\log_x(y^x) = 10 \implies x \cdot \log_x(y) = 10\).
            Similarly, the second equation: \(\log_y(x^{4y}) = 10 \implies 4y \cdot \log_y(x) = 10\).
            
            Let me denote \(\log_x(y) = a\). Then, \(\log_y(x) = 1/a\).
            So, substituting back into the equations:
            \[ 1)\; x \cdot a = 10 \implies x = 10/a \]
            \[ 2)\; 4y \cdot (1/a) = 10 \implies 4y = 10a \implies y = 5a/2 \]
            
            Now, I need another equation to relate x and y. I know that \(a = \log_x(y)\). So, \(x^a = y\).
            Substituting x and y:
            \[ (10/a)^a = 5a/2 \]
            Let me take natural logarithm on both sides:
            \[ \begin{aligned} a \cdot \ln(10/a) &= \ln(5a/2) \\ \implies a \ln(10) - a \ln(a) &= \ln(5) + \ln(a) - \ln(2) \end{aligned} \]
            
            \CaseOmit{attempting to solve the transcendental equation numerically using Newton-Raphson}
            \CaseOmit{trial and error with integer values $t=1, 2, 3$ leading to contradictions}
            
            Wait, let me think differently.
            From the first equation, \(\log_x(y^x)=10 \implies x \cdot (\ln y / \ln x)=10\).
            From the second equation, \(\log_y(x^{4y})=10 \implies 4y \cdot (\ln x / \ln y)=10\).
            
            Let me denote \(u = \ln x, v = \ln y\).
            1. \(x \cdot (v/u) = 10 \implies e^u \cdot (v/u) = 10\)
            2. \(4y \cdot (u/v) = 10 \implies 4e^v \cdot (u/v) = 10\)
            
            From (1): \(e^u = 10u/v\). From (2): \(e^v = 10v/(4u) = 5v/(2u)\).
            Let's compute \(e^u / e^v\):
            \[ \frac{e^u}{e^v} = \frac{10u/v}{5v/2u} = \frac{10u}{v} \cdot \frac{2u}{5v} = \frac{20u^2}{5v^2} = \frac{4u^2}{v^2} \]
            So \(e^{u-v} = 4(u/v)^2\).
            Let \(k = u/v\). Then \(u = kv\).
            \(e^{kv - v} = 4k^2 \implies e^{v(k-1)} = 4k^2\).
            
            \CaseOmit{substitution $u=kv$ and solving the resulting exponential equation for k}
            
            Wait, let's look at the equations again.
            \(x \cdot \log_x y = 10\)
            \(4y \cdot \log_y x = 10\)
            Multiply them:
            \[ (x \log_x y) \cdot (4y \log_y x) = 10 \cdot 10 \]
            \[ 4xy (\log_x y \cdot \log_y x) = 100 \]
            
            We know that \(\log_x y \cdot \log_y x = 1\).
            So:
            \[ 4xy \cdot 1 = 100 \implies 4xy = 100 \implies xy = 25 \]
            
            Let me verify this. If \(xy=25\), then \(y = 25/x\).
            Substitute into first equation:
            \[ x \log_x(25/x) = 10 \implies x (\log_x 25 - 1) = 10 \]
            \[ \log_x (25^x) = x + 10 \implies 25^x = x^{x+10} \]
            
            This seems solvable. Let's try \(x=5\).
            \(25^5 = (5^2)^5 = 5^{10}\).
            RHS: \(5^{5+10} = 5^{15}\).
            \(5^{10} \neq 5^{15}\). So \(x=5\) is not the solution.
            
            \CaseOmit{extensive re-verification with $x=10, 100, 32$ and checking constraints}
            
            Let's re-solve \(x \log_x y = 10\) and \(4y \log_y x = 10\).
            We found \(xy = 25\).
            Let's check if there exist such x, y.
            \(x (\ln y / \ln x) = 10 \implies \ln y = (10 \ln x)/x\).
            We also have \(y = 25/x\).
            Take ln: \(\ln y = \ln 25 - \ln x\).
            So \(\frac{10 \ln x}{x} = \ln 25 - \ln x\).
            \((10/x + 1) \ln x = \ln 25 \implies \frac{10+x}{x} \ln x = \ln 25\).
            
            Let \(f(x) = \frac{10+x}{x} \ln x - \ln 25\).
            \(f(1) < 0\).
            \(f(5) = 3 \ln 5 - 2 \ln 5 = \ln 5 > 0\).
            So there is a solution between 1 and 5.
            
            So yes, real numbers x and y exist.
            The value required is \(xy\).
            Calculation: \(4xy = 100 \implies xy = 25\).
            
            \vspace{0.5em}
            \CaseAnswerBox{\textbf{25}}
        \end{tcolorbox}
    \end{minipage}

    \vspace{2mm}
    \captionof{figure}{Case study from AIME-24.}
    \label{fig:case_aime24}
\end{figure*}

\makeatletter
\providecommand{\captionof}[2]{\begingroup\def\@captype{#1}\caption{#2}\endgroup}
\makeatother

\definecolor{caseQHeader}{HTML}{7A7A7A}
\definecolor{caseQFrame}{HTML}{7A7A7A}
\definecolor{caseQBack}{HTML}{F2F2F2}
\definecolor{caseOursHeader}{HTML}{AED6FC}
\definecolor{caseOursFrame}{HTML}{0983F6}
\definecolor{caseBaseHeader}{HTML}{7EDDD2}
\definecolor{caseBaseFrame}{HTML}{2A9D8F}
\definecolor{caseOmit}{HTML}{E76F51}

\providecommand{\CaseOmit}[1]{\par\noindent\textcolor{caseOmit}{\textit{... (omitted: #1)}}\par}
\providecommand{\CaseAnswerBox}[1]{\begin{center}\fbox{\strut #1}\end{center}}
\providecommand{\CaseGTBox}[1]{\begin{center}\fbox{\strut #1}\end{center}}

\tcbset{
  casestudy/base/.style={
    enhanced,
    boxrule=0.5mm,
    arc=2mm,
    left=1.5mm,right=1.5mm,top=1.5mm,bottom=1.5mm,
    colback=white,
    coltitle=black,
    fonttitle=\bfseries\small,
    fontupper=\scriptsize,
    before skip=0pt,
    after skip=0pt,
    before upper={\setlength{\abovedisplayskip}{3pt}\setlength{\belowdisplayskip}{3pt}\setlength{\parskip}{0.3em}},
  },
  casestudy/question/.style={
    casestudy/base,
    colback=caseQBack,
    colframe=caseQFrame,
    colbacktitle=caseQHeader,
    coltitle=white,
  },
  casestudy/ours/.style={
    casestudy/base,
    colframe=caseOursFrame,
    colbacktitle=caseOursHeader,
    coltitle=black,
  },
  casestudy/baseline/.style={
    casestudy/base,
    colframe=caseBaseFrame,
    colbacktitle=caseBaseHeader,
    coltitle=black,
    before upper={\setlength{\abovedisplayskip}{4pt}\setlength{\belowdisplayskip}{4pt}\setlength{\parskip}{0.5em}},
  },
}

\begin{figure*}[t]
    \centering
    \begin{minipage}[t]{0.48\textwidth}
        \vspace{0pt}
        
        \begin{tcolorbox}[casestudy/question, title={\textbf{Question and Ground Truth}}]
            \textbf{Question:} Sixteen chairs are arranged in a row. Eight people each select a chair in which to sit so that no person sits next to two other people. Let \(N\) be the number of subsets of the 16 chairs that could be selected. Find the remainder when \(N\) is divided by \(1000\).
            
            \vspace{0.5em}
            \textbf{Ground Truth:} \CaseGTBox{907}
        \end{tcolorbox}
        
        \vspace{0.3cm}
        
        \begin{tcolorbox}[casestudy/ours, title={\textbf{Response from DOT}\hfill \textbf{1695 Tokens}}]
            Okay, so I have this problem with 16 chairs arranged in a row. Eight people need to sit such that no one is next to two others. Hmm, so that means no three people are sitting consecutively?
            Yes, if three people sit in chairs 1, 2, 3, then the person in chair 2 is next to two others. So the condition is equivalent to selecting 8 chairs such that no three are consecutive.
            
            This is a combinatorial problem where we need to count the number of binary strings of length 16 with exactly 8 ones, such that there are no three consecutive ones.
            We can model this using blocks. Let's denote the number of blocks as \(m\). Each block is either size 1 or 2 (since size 3 is forbidden).
            Let \(t\) be the number of blocks of size 1 and \(s\) be the number of blocks of size 2.
            Total ones: \(t \cdot 1 + s \cdot 2 = 8\).
            Total blocks: \(t + s\).
            
            Between these blocks, we need at least one zero separating them.
            Total zeros: \(16 - 8 = 8\).
            Zeros used as separators: \((t + s) - 1\).
            Remaining zeros: \(8 - (t + s - 1) = 9 - (t + s)\).
            We distribute these remaining zeros into \(t + s + 1\) positions (before, between, after blocks).
            Using stars and bars, the number of ways is:
            \[ \binom{(9 - (t+s)) + (t+s+1) - 1}{(t+s+1) - 1} = \binom{9}{t+s} \]
            
            Now we iterate over possible values of \(s\) (blocks of size 2):
            Since \(2s \le 8\), \(s\) can range from 0 to 4.
            For each \(s\), \(t = 8 - 2s\).
            Number of ways to arrange the blocks themselves: \(\binom{t+s}{s} = \binom{8-s}{s}\).
            Total ways for a fixed \(s\): \(\binom{8-s}{s} \cdot \binom{9}{8-s}\).
            
            Let's calculate the sum for \(s = 0\) to \(4\):
            \(s=0: \binom{8}{0}\binom{9}{8} = 1 \cdot 9 = 9\).
            \(s=1: \binom{7}{1}\binom{9}{7} = 7 \cdot 36 = 252\).
            \(s=2: \binom{6}{2}\binom{9}{6} = 15 \cdot 84 = 1260\).
            \(s=3: \binom{5}{3}\binom{9}{5} = 10 \cdot 126 = 1260\).
            \(s=4: \binom{4}{4}\binom{9}{4} = 1 \cdot 126 = 126\).
            
            Total sum: \(9 + 252 + 1260 + 1260 + 126 = 2907\).
            The remainder modulo 1000 is \(907\).
            
            \vspace{0.5em}
            \CaseAnswerBox{\textbf{907}}
        \end{tcolorbox}
    \end{minipage}
    \hfill
    \begin{minipage}[t]{0.48\textwidth}
        \vspace{0pt}
        
        \begin{tcolorbox}[casestudy/baseline, title={\textbf{Response from R1-Distill-Qwen-32B}\hfill \textbf{17061 Tokens}}]
            Alright, so I have this problem where there are 16 chairs arranged in a row... The condition is that no person is sitting next to two other people.
            That means no three people can be sitting consecutively. So, the problem reduces to finding the number of ways to choose 8 chairs out of 16 such that no three are consecutive.
            
            Let's define a recursive function \(f(n, k)\) as the number of ways to choose \(k\) chairs from \(n\) with no three consecutive. The recurrence relation is:
            \[ f(n, k) = f(n-1, k) + f(n-2, k-1) + f(n-3, k-2) \]
            This seems correct. I need to compute \(f(16, 8)\).
            
            \CaseOmit{lengthy manual computation of DP table for n=1 to 10, checking base cases and encountering index errors}
            
            Wait, let me think about another approach. Perhaps using generating functions? The generating function for binary strings with no "111" pattern is related to \(\frac{1}{1-x-x^2}\). But here we have a fixed number of ones (8) and a fixed length (16). Maybe the generating function \(G(n) = 1 + x + x^2\) corresponds to coefficients?
            No, that's for compositions. The number of such subsets is the coefficient of \(x^8\) in some polynomial related to length 16.
            
            \CaseOmit{exploring the generating function $G(n) = G(n-1) + xG(n-2) + x^2G(n-3)$ and attempting to calculate coefficients}
            
            Let's try the Combinations with Blocks approach again. This seems more robust.
            Let’s denote the number of blocks as \(m\). Each block is either size 1 or 2. Let \(t\) be the number of blocks of size 2. Then \(m-t\) is blocks of size 1. Total people: \(2t + (m-t) = t + m = 8 \implies m = 8 - t\).
            Wait, \(t\) is number of size 2 blocks. So \(t\) can range from 0 to 4. For each \(t\), the total number of blocks is \(8-t\).
            
            Number of ways to arrange the blocks (order of 1s and 2s):
            \[ \binom{\text{total blocks}}{\text{blocks of size 2}} = \binom{8-t}{t} \]
            
            Now we need to place these \(8-t\) blocks into the 16 chairs with at least one empty chair between them. Let \(x_0, x_1, \dots, x_{8-t}\) be the number of empty chairs in the gaps. Total empty chairs is 8.
            Constraint: \(x_i \ge 1\) for the internal gaps \(1 \le i \le 8-t-1\). Let \(y_i = x_i - 1\) for these internal gaps.
            The equation becomes a stars and bars problem. Total sum of empty chairs available to distribute freely:
            \[ 8 - (\text{number of internal gaps}) = 8 - (8-t-1) = t+1 \]
            The number of bins is \((8-t) + 1 = 9-t\). So we are distributing \(t+1\) items into \(9-t\) bins. Number of solutions is \(\binom{(t+1) + (9-t) - 1}{(9-t) - 1} = \binom{9}{8-t}\).
            
            So total ways for a fixed \(t\) is \(\binom{8-t}{t} \binom{9}{8-t}\). Let's sum over \(t=0\) to 4.
            
            \CaseOmit{calculation of each term: t=0 -> 9, t=1 -> 252, t=2 -> 1260, t=3 -> 1260, t=4 -> 126}
            
            Summing them up: \(2907\).
            Total \(N = 2907\). Find remainder when divided by 1000: \(2907 \pmod{1000} = 907\).
            
            Let me double check the DP calculation for a small case, say n=4, k=2.
            Using blocks: t=0 (blocks of 1,1): \(\binom{2}{0}\binom{3}{2} = 3\).
            t=1 (block of 2): \(\binom{1}{1}\binom{3}{1} = 3\).
            Total = 6. Matches manual count: {1,3}, {1,4}, {2,4}, {1,2}, {2,3}, {3,4}. Wait, {1,2} valid? Yes.
            
            So the formula holds. Final answer is 907.
            
            \vspace{0.5em}
            \CaseAnswerBox{\textbf{907}}
        \end{tcolorbox}
    \end{minipage}

    \vspace{2mm}
    \captionof{figure}{Case study from AIME-25.}
    \label{fig:case_aime25}
\end{figure*}


\makeatletter
\providecommand{\captionof}[2]{\begingroup\def\@captype{#1}\caption{#2}\endgroup}
\makeatother

\definecolor{caseQHeader}{HTML}{7A7A7A}
\definecolor{caseQFrame}{HTML}{7A7A7A}
\definecolor{caseQBack}{HTML}{F2F2F2}
\definecolor{caseOursHeader}{HTML}{AED6FC}
\definecolor{caseOursFrame}{HTML}{0983F6}
\definecolor{caseBaseHeader}{HTML}{7EDDD2}
\definecolor{caseBaseFrame}{HTML}{2A9D8F}
\definecolor{caseOmit}{HTML}{E76F51}

\providecommand{\CaseOmit}[1]{\par\noindent\textcolor{caseOmit}{\textit{... (omitted: #1)}}\par}
\providecommand{\CaseAnswerBox}[1]{\begin{center}\fbox{\strut #1}\end{center}}
\providecommand{\CaseGTBox}[1]{\begin{center}\fbox{\strut #1}\end{center}}

\tcbset{
  casestudy/base/.style={
    enhanced,
    boxrule=0.5mm,
    arc=2mm,
    left=1.5mm,right=1.5mm,top=1.5mm,bottom=1.5mm,
    colback=white,
    coltitle=black,
    fonttitle=\bfseries\small,
    fontupper=\scriptsize,
    before skip=0pt,
    after skip=0pt,
    before upper={\setlength{\abovedisplayskip}{3pt}\setlength{\belowdisplayskip}{3pt}\setlength{\parskip}{0.3em}},
  },
  casestudy/question/.style={
    casestudy/base,
    colback=caseQBack,
    colframe=caseQFrame,
    colbacktitle=caseQHeader,
    coltitle=white,
  },
  casestudy/ours/.style={
    casestudy/base,
    colframe=caseOursFrame,
    colbacktitle=caseOursHeader,
    coltitle=black,
  },
  casestudy/baseline/.style={
    casestudy/base,
    colframe=caseBaseFrame,
    colbacktitle=caseBaseHeader,
    coltitle=black,
    before upper={\setlength{\abovedisplayskip}{4pt}\setlength{\belowdisplayskip}{4pt}\setlength{\parskip}{0.6em}},
  },
}

\begin{figure*}[t]
    \centering
    \begin{minipage}[t]{0.48\textwidth}
        \vspace{0pt}
        
        \begin{tcolorbox}[casestudy/question, title={\textbf{Question and Ground Truth}}]
            \textbf{Question:} The infinite product
            \[\sqrt[3]{10} \cdot \sqrt[3]{\sqrt[3]{10}} \cdot \sqrt[3]{\sqrt[3]{\sqrt[3]{10}}} \cdots\]
            evaluates to a real number \(\sqrt{m}\). What is the value of \(m\)?
            
            \vspace{0.5em}
            \textbf{Ground Truth:} \CaseGTBox{10.0}
        \end{tcolorbox}
        
        \vspace{0.3cm}
        
        \begin{tcolorbox}[casestudy/ours, title={\textbf{Response from DOT}\hfill \textbf{464 Tokens}}]
            Okay, so I need to evaluate this infinite product:
            \[ \sqrt[3]{10} \cdot \sqrt[3]{\sqrt[3]{10}} \cdot \sqrt[3]{\sqrt[3]{\sqrt[3]{10}}} \cdots \]
            And it's supposed to equal \(\sqrt{m}\). I have to find \(m\).
            
            Let me try to understand the structure of this product.
            First, each term is a cube root of the previous term. Let's write the terms out:
            First term: \(\sqrt[3]{10} = 10^{1/3}\)
            Second term: \(\sqrt[3]{\sqrt[3]{10}} = (10^{1/3})^{1/3} = 10^{1/9}\)
            Third term: \(\sqrt[3]{\sqrt[3]{\sqrt[3]{10}}} = (10^{1/9})^{1/3} = 10^{1/27}\)
            
            So in general, the \(n\)-th term is \(10^{1/3^n}\).
            Therefore, the infinite product is:
            \[ \prod_{n=1}^{\infty} 10^{1/3^n} = 10^{\sum_{n=1}^{\infty} (1/3)^n} \]
            Because when you multiply exponents with the same base, you add the exponents.
            
            So sum of exponents is \(\sum_{n=1}^{\infty} (1/3)^n\).
            That's a geometric series with first term \(a = 1/3\) and common ratio \(r = 1/3\).
            Sum \(S = \frac{a}{1 - r} = \frac{1/3}{1 - 1/3} = \frac{1/3}{2/3} = \frac{1}{2}\).
            
            Therefore, the product is \(10^{1/2} = \sqrt{10}\).
            But wait, the problem says it evaluates to \(\sqrt{m}\). So \(\sqrt{m} = \sqrt{10}\), hence \(m=10\).
            
            Wait, let me double-check.
            Yes, the exponents add up to \(1/2\), so \(10^{1/2}\) is \(\sqrt{10}\), so \(m\) is 10.
            
            \vspace{0.5em}
            \CaseAnswerBox{\textbf{10}}
        \end{tcolorbox}
    \end{minipage}
    \hfill
    \begin{minipage}[t]{0.48\textwidth}
        \vspace{0pt}
        
        \begin{tcolorbox}[casestudy/baseline, title={\textbf{Response from R1-Distill-Qwen-1.5B}\hfill \textbf{23454 Tokens}}]
            Okay, so I have this infinite product to evaluate... It's written as:
            \[ \sqrt[3]{10} \cdot \sqrt[3]{\sqrt[3]{10}} \cdot \sqrt[3]{\sqrt[3]{\sqrt[3]{10}}} \cdots \]
            And it's supposed to evaluate to a real number which is the square root of some number \(m\).
            
            First, let's try to understand the structure. Each term is a cube root of the previous term.
            So, if I write out the first few terms, it's:
            \(10^{1/3} \cdot 10^{1/9} \cdot 10^{1/27} \cdot 10^{1/81} \cdot \dots\)
            The exponents are \(1/3, 1/9, 1/27, 1/81\), etc.
            The sum of the exponents is the sum of the infinite geometric series with \(a = 1/3\) and \(r = 1/3\).
            \(S = \frac{1/3}{1 - 1/3} = \frac{1/3}{2/3} = \frac{1}{2}\).
            Therefore, the entire infinite product is \(10^{1/2} = \sqrt{10}\).
            So \(\sqrt{m} = \sqrt{10} \implies m = 10\).
            
            Wait, that seems too simple. Let me double-check.
            Let me compute the partial products numerically to see.
            \(P_1 = 10^{1/3} \approx 2.154\).
            \(P_2 = 10^{1/3} \cdot 10^{1/9} = 10^{4/9} \approx 2.783\).
            Wait, \(4/9 \approx 0.444\) which is less than \(0.5\).
            
            \CaseOmit{detailed numerical computation of P3, P4, P5 finding P5 approx 2.887}
            
            Wait, so \(P_5 \approx 2.887\), which is less than \(\sqrt{10} \approx 3.162\).
            My partial products are approaching 2.8 something, not 3.16. Why?
            Let me compute more accurately.
            
            \CaseOmit{re-calculating P1 to P6 with more precision, still finding discrepancy}
            
            So the partial products are approaching approximately 2.815. This is less than \(\sqrt{10}\).
            Maybe my exponent series is wrong?
            The exponents are \(1/3, 1/9, 1/27\)... Sum is \(1/2\).
            So theoretically it is \(10^{1/2}\).
            But numerically, it looks like it's converging to something else.
            
            \CaseOmit{questioning the geometric series sum formula and re-deriving it}
            \CaseOmit{calculating P7, P8... P12, finding values slowly creeping up to 2.963}
            
            \(P_{12} \approx 2.9632\). Still less than 3.162.
            This is confusing. The math says \(\sqrt{10}\), the numbers say 2.96.
            Maybe I need to compute even more terms.
            
            \CaseOmit{computing up to P24, P30, P41, finding values like 2.9632263}
            
            So up to \(P_{41}\), the product is approximately 2.963226.
            Wait, \(10^{1/2} \approx 3.162\).
            Why is there such a gap? Is it possible the limit is not \(\sqrt{10}\)?
            
            Let's rethink. \(P = \prod_{n=1}^{\infty} 10^{1/3^n} = 10^{\sum 1/3^n}\).
            Sum is \(1/2\). So \(P = 10^{1/2}\).
            Why do my numbers disagree?
            Maybe my calculator precision is the issue? Or maybe convergence is very slow?
            Actually, let's look at the partial sum of exponents.
            \(S_n = \sum_{k=1}^n \frac{1}{3^k} = \frac{1}{2}(1 - \frac{1}{3^n})\).
            So \(P_n = 10^{0.5(1 - 3^{-n})}\).
            As \(n \to \infty\), \(3^{-n} \to 0\), so \(P_n \to 10^{0.5}\).
            
            Let's check \(n=2\): \(P_2 = 10^{0.5(1 - 1/9)} = 10^{0.5(8/9)} = 10^{4/9}\).
            \(4/9 \approx 0.444\). \(10^{0.444} \approx 2.78\). Correct.
            \(n=10\): \(10^{0.5(1 - 1/59049)} \approx 10^{0.5} \approx 3.162\).
            Wait, earlier I calculated \(P_{10} \approx 2.96\). Why?
            Ah, I must have been accumulating rounding errors or multiplying wrong terms in my manual simulation.
            
            The theoretical derivation is solid. The sum of the geometric series is \(1/2\).
            The product must be \(10^{1/2}\).
            The numerical confusion is likely due to calculation errors or slow convergence interpretation.
            The problem states it evaluates to \(\sqrt{m}\).
            So \(\sqrt{m} = \sqrt{10} \implies m = 10\).
            
            \vspace{0.5em}
            \CaseAnswerBox{\textbf{10}}
        \end{tcolorbox}
    \end{minipage}

    \vspace{2mm}
    \captionof{figure}{Case study from AMC.}
    \label{fig:case_amc_20}
\end{figure*}


\makeatletter
\providecommand{\captionof}[2]{\begingroup\def\@captype{#1}\caption{#2}\endgroup}
\makeatother

\definecolor{caseQHeader}{HTML}{7A7A7A}
\definecolor{caseQFrame}{HTML}{7A7A7A}
\definecolor{caseQBack}{HTML}{F2F2F2}
\definecolor{caseOursHeader}{HTML}{AED6FC}
\definecolor{caseOursFrame}{HTML}{0983F6}
\definecolor{caseBaseHeader}{HTML}{7EDDD2}
\definecolor{caseBaseFrame}{HTML}{2A9D8F}
\definecolor{caseOmit}{HTML}{E76F51}

\providecommand{\CaseOmit}[1]{\par\noindent\textcolor{caseOmit}{\textit{... (omitted: #1)}}\par}
\providecommand{\CaseAnswerBox}[1]{\begin{center}\fbox{\strut #1}\end{center}}
\providecommand{\CaseGTBox}[1]{\begin{center}\fbox{\strut #1}\end{center}}

\tcbset{
  casestudy/base/.style={
    enhanced,
    boxrule=0.5mm,
    arc=2mm,
    left=1.5mm,right=1.5mm,top=1.5mm,bottom=1.5mm,
    colback=white,
    coltitle=black,
    fonttitle=\bfseries\small,
    fontupper=\scriptsize,
    before skip=0pt,
    after skip=0pt,
    before upper={\setlength{\abovedisplayskip}{4pt}\setlength{\belowdisplayskip}{4pt}\setlength{\parskip}{0.4em}},
  },
  casestudy/question/.style={
    casestudy/base,
    colback=caseQBack,
    colframe=caseQFrame,
    colbacktitle=caseQHeader,
    coltitle=white,
  },
  casestudy/ours/.style={
    casestudy/base,
    colframe=caseOursFrame,
    colbacktitle=caseOursHeader,
    coltitle=black,
  },
  casestudy/baseline/.style={
    casestudy/base,
    colframe=caseBaseFrame,
    colbacktitle=caseBaseHeader,
    coltitle=black,
    before upper={\setlength{\abovedisplayskip}{6pt}\setlength{\belowdisplayskip}{6pt}\setlength{\parskip}{0.7em}},
  },
}

\begin{figure*}[t]
    \centering
    \begin{minipage}[t]{0.48\textwidth}
        \vspace{0pt}
        
        \begin{tcolorbox}[casestudy/question, title={\textbf{Question and Ground Truth}}]
            \textbf{Question:} Compute \(997^{-1}\) modulo \(1000\). Express your answer as an integer from \(0\) to \(999\).
            
            \vspace{0.5em}
            \textbf{Ground Truth:} \CaseGTBox{333}
        \end{tcolorbox}
        
        \vspace{0.3cm}
        
        \begin{tcolorbox}[casestudy/ours, title={\textbf{Response from DOT}\hfill \textbf{547 Tokens}}]
            Okay, so I need to find the inverse of 997 modulo 1000. That means I need to find an integer \(x\) such that \(997 \cdot x \equiv 1 \pmod{1000}\). Hmm, so \(997x - 1000y = 1\) for some integer \(y\). This is a linear Diophantine equation.
            
            Maybe I can use the Extended Euclidean Algorithm. Let's see.
            First, let's compute \(\gcd(997, 1000)\).
            1000 divided by 997 is 1 with a remainder of 3.
            So \(1000 = 997 \cdot 1 + 3\).
            Then, 997 divided by 3 is 332 with a remainder of 1.
            \(997 = 3 \cdot 332 + 1\).
            Then, 3 divided by 1 is 3 with remainder 0.
            So gcd is 1, which means inverse exists.
            
            Now, to backtrack.
            \(1 = 997 - 3 \cdot 332\).
            But \(3 = 1000 - 997 \cdot 1\).
            Substitute:
            \[
            \begin{aligned}
            1 &= 997 - (1000 - 997 \cdot 1) \cdot 332 \\
              &= 997 - 1000 \cdot 332 + 997 \cdot 332 \\
              &= 997 \cdot (1 + 332) - 1000 \cdot 332 \\
              &= 997 \cdot 333 - 1000 \cdot 332.
            \end{aligned}
            \]
            So, \(1 = 997 \cdot 333 - 1000 \cdot 332\).
            Therefore, \(997 \cdot 333 \equiv 1 \pmod{1000}\).
            So inverse is 333.
            
            \vspace{0.5em}
            \CaseAnswerBox{\textbf{333}}
        \end{tcolorbox}
    \end{minipage}
    \hfill
    \begin{minipage}[t]{0.48\textwidth}
        \vspace{0pt}
        
        \begin{tcolorbox}[casestudy/baseline, title={\textbf{Response from R1-Distill-Qwen-32B}\hfill \textbf{18984 Tokens}}]
            Alright, so I need to find the modular inverse of 997 modulo 1000. Hmm, that means I need to find an integer \(x\) such that \(997x \equiv 1 \pmod{1000}\).
            
            First, check if 997 and 1000 are coprime. 997 is prime, and 1000 is divisible by 2 and 5. So yes, \(\gcd(997, 1000) = 1\). The inverse exists.
            
            \textbf{Method 1: Extended Euclidean Algorithm.}
            \[
            \begin{aligned}
            1000 &= 997 \cdot 1 + 3 \\
            997 &= 3 \cdot 332 + 1 \\
            3 &= 1 \cdot 3 + 0
            \end{aligned}
            \]
            Working backwards:
            \[
            \begin{aligned}
            1 &= 997 - 3 \cdot 332 \\
              &= 997 - (1000 - 997 \cdot 1) \cdot 332 \\
              &= 997 \cdot 333 - 1000 \cdot 332
            \end{aligned}
            \]
            So \(x = 333\). Let's verify: \(997 \cdot 333 = 332001 \equiv 1 \pmod{1000}\). Correct.
            
            \textbf{Method 2: Solving Congruence directly.}
            \(997 \equiv -3 \pmod{1000}\).
            So \(-3x \equiv 1 \pmod{1000}\).
            \(3x \equiv -1 \equiv 999 \pmod{1000}\).
            Divide by 3: \(x = 333\). Same result.
            
            \textbf{Method 3: Euler's Theorem.}
            \(\phi(1000) = 1000(1-1/2)(1-1/5) = 400\).
            So \(997^{400} \equiv 1 \pmod{1000}\).
            Inverse is \(997^{399} \pmod{1000}\).
            This seems tedious to compute manually.
            
            Let's try to compute \(3^{399} \pmod{1000}\) using exponentiation by squaring.
            \(399 = 256 + 128 + 8 + 4 + 2 + 1\).
            Let's calculate powers of 3:
            \[
            \begin{aligned}
            3^1 &= 3 \\
            3^2 &= 9 \\
            3^4 &= 81 \\
            3^8 &= 6561 \equiv 561 \\
            3^{16} &\equiv 561^2 \equiv 314721 \equiv 721 \\
            3^{32} &\equiv 721^2 \equiv 519841 \equiv 841 \\
            3^{64} &\equiv 841^2 \equiv 707281 \equiv 281 \\
            3^{128} &\equiv 281^2 \equiv 78961 \equiv 961 \\
            3^{256} &\equiv 961^2 \equiv 923521 \equiv 521
            \end{aligned}
            \]
            Now multiply terms for 399:
            \(521 \cdot 961 \equiv 500681 \equiv 681\)
            \(681 \cdot 561 \equiv 382041 \equiv 41\)
            \(41 \cdot 81 = 3321 \equiv 321\)
            \(321 \cdot 9 = 2889 \equiv 889\)
            \(889 \cdot 3 = 2667 \equiv 667\)
            So \(3^{399} \equiv 667\).
            Since \(997 \equiv -3\), \(997^{399} \equiv (-3)^{399} \equiv -667 \equiv 333\). Matches.
            
            \CaseOmit{attempting method 4: trial multiplication 997*1, 997*2... inefficient}
            
            \textbf{Method 5: Inverse of 3.}
            We need inverse of -3.
            Inverse of 3 is 667 (since \(3 \cdot 667 = 2001 \equiv 1\)).
            So inverse of -3 is \(-667 \equiv 333\).
            
            All methods confirm the result is 333.
            
            \vspace{0.5em}
            \CaseAnswerBox{\textbf{333}}
        \end{tcolorbox}
    \end{minipage}

    \vspace{2mm}
    \captionof{figure}{Case study from MATH-500.}
    \label{fig:case_math}
\end{figure*}


\makeatletter
\providecommand{\captionof}[2]{\begingroup\def\@captype{#1}\caption{#2}\endgroup}
\makeatother

\definecolor{caseQHeader}{HTML}{7A7A7A}
\definecolor{caseQFrame}{HTML}{7A7A7A}
\definecolor{caseQBack}{HTML}{F2F2F2}
\definecolor{caseOursHeader}{HTML}{AED6FC}
\definecolor{caseOursFrame}{HTML}{0983F6}
\definecolor{caseBaseHeader}{HTML}{7EDDD2}
\definecolor{caseBaseFrame}{HTML}{2A9D8F}
\definecolor{caseOmit}{HTML}{E76F51}

\providecommand{\CaseOmit}[1]{\par\noindent\textcolor{caseOmit}{\textit{... (omitted: #1)}}\par}
\providecommand{\CaseAnswerBox}[1]{\begin{center}\fbox{\strut #1}\end{center}}
\providecommand{\CaseGTBox}[1]{\begin{center}\fbox{\strut #1}\end{center}}

\tcbset{
  casestudy/base/.style={
    enhanced,
    boxrule=0.5mm,
    arc=2mm,
    left=1.5mm,right=1.5mm,top=1.5mm,bottom=1.5mm,
    colback=white,
    coltitle=black,
    fonttitle=\bfseries\small,
    fontupper=\scriptsize,
    before skip=0pt,
    after skip=0pt,
    before upper={\setlength{\abovedisplayskip}{3pt}\setlength{\belowdisplayskip}{3pt}\setlength{\parskip}{0.3em}},
  },
  casestudy/question/.style={
    casestudy/base,
    colback=caseQBack,
    colframe=caseQFrame,
    colbacktitle=caseQHeader,
    coltitle=white,
  },
  casestudy/ours/.style={
    casestudy/base,
    colframe=caseOursFrame,
    colbacktitle=caseOursHeader,
    coltitle=black,
  },
  casestudy/baseline/.style={
    casestudy/base,
    colframe=caseBaseFrame,
    colbacktitle=caseBaseHeader,
    coltitle=black,
    before upper={\setlength{\abovedisplayskip}{5pt}\setlength{\belowdisplayskip}{5pt}\setlength{\parskip}{0.6em}},
  },
}

\begin{figure*}[t]
    \centering
    \begin{minipage}[t]{0.48\textwidth}
        \vspace{0pt}
        
        \begin{tcolorbox}[casestudy/question, title={Question and Examples}]
            \textbf{Prompt:} Given a list of numbers, return the sum of squares of the numbers in the list that are odd. Ignore numbers that are negative or not integers.
            
            \vspace{0.5em}
            \textbf{Examples:}
            \begin{itemize}
                \item `[1, 3, 2, 0]` \(\to\) \(1^2 + 3^2 = 10\)
                \item `[-1, -2, 0]` \(\to\) 0
                \item `[9, -2]` \(\to\) 81
                \item `[0]` \(\to\) 0
            \end{itemize}
        \end{tcolorbox}
        
        \vspace{0.3cm}
        
        \begin{tcolorbox}[casestudy/ours, title={\textbf{Response from DOT}\hfill \textbf{546 Tokens}}]
            Okay, I need to write a Python function called \texttt{double\_the\_difference}. Hmm, wait, looking at the examples, the function's name might be a bit misleading because the docstring says it's the sum of squares of odd numbers, ignoring negatives and non-integers.
            
            So the steps are: for each number in the list, check if it's an integer and not negative. Then, check if it's odd. If all conditions met, square it and add to sum.
            
            First, iterate through each element in lst.
            For each element, check if it's an integer. Wait, but in Python, how to determine if it's an integer. Because for example, 3.0 is technically a float but represents integer. But the problem says to ignore numbers that are not integers. So perhaps, the element must be an instance of int.
            
            So the condition is: type must be int, and the number must be >=0, and odd.
            So for each num in lst:
            if \texttt{type(num)} is int and \texttt{num >= 0} and \texttt{num \% 2 != 0}:
            then square it and add to sum.
            
            \textbf{Solution Code:}
            \begin{verbatim}
def double_the_difference(lst):
    total = 0
    for num in lst:
        if isinstance(num, int) and num >= 0 and num % 2 == 1:
            total += num ** 2
    return total
            \end{verbatim}
            
            This handles the examples:
            [1,3,2,0] -> 1 is odd (1), 3 is odd (9), 2 even, 0 even -> 10. Correct.
            [-1, -2, 0] -> negative/even -> 0. Correct.
            
            The logic holds.
        \end{tcolorbox}
    \end{minipage}
    \hfill
    \begin{minipage}[t]{0.48\textwidth}
        \vspace{0pt}
        
        \begin{tcolorbox}[casestudy/baseline, title={\textbf{Response from R1-Distill-Qwen-32B}\hfill \textbf{5486 Tokens}}]
            Okay, I need to write a Python function called \texttt{double\_the\_difference}.
            Let me break it down. For each number in the list, I have to check two things: is it a positive integer, and is it odd.
            
            \CaseOmit{checking simple cases like [1,3,2,0] and confirming odd/even logic}
            
            The problem says to ignore numbers that are not integers.
            Question: Is a float like \texttt{3.0} considered an integer?
            The examples only include pure integers.
            If \texttt{3.0} is passed, \texttt{isinstance(3.0, int)} is False.
            If I use \texttt{num.is\_integer()}, it is True.
            Which one does the problem want?
            
            \CaseOmit{debating the ambiguity of "not integers" in Python context}
            
            Let's assume strict type checking. Only \texttt{int} is allowed.
            
            Now, what about Booleans?
            In Python, \texttt{True} and \texttt{False} are instances of \texttt{int}.
            \texttt{isinstance(True, int)} returns \texttt{True}.
            If the list contains \texttt{True}, it acts like \texttt{1}.
            \texttt{1} is positive and odd. So \texttt{True} would be squared and added (1).
            
            \CaseOmit{testing code snippet with isinstance against boolean inputs}
            
            Is this desired behavior? The input is a "list of numbers". Booleans are technically numbers in Python, but semantically usually distinct.
            If I want to exclude Booleans, I must check the type explicitly.
            
            Let's compare:
            \begin{itemize}
                \item \texttt{isinstance(x, int)}: Accepts \texttt{True}, \texttt{False}.
                \item \texttt{type(x) is int}: Rejects \texttt{True}, \texttt{False}.
            \end{itemize}
            
            \CaseOmit{considering if the problem implies strict number types only}
            
            I think strict type checking \texttt{type(num) is int} is the safest bet to avoid edge cases like Booleans or subclasses of int that might behave weirdly.
            
            So the plan is:
            1. Iterate over \texttt{lst}.
            2. Check \texttt{type(num) is int}.
            3. Check \texttt{num > 0}.
            4. Check \texttt{num \% 2 != 0}.
            
            \CaseOmit{verifying the negative number logic and zero handling}
            
            Let's double check example 3: \texttt{[9, -2]}.
            \texttt{9}: int, positive, odd. Add 81.
            \texttt{-2}: int, not positive. Skip.
            Result 81. Matches.
            
            \CaseOmit{verifying empty list behavior}
            
            Final check on the implementation details.
            
            \vspace{0.5em}
            \textbf{Solution Code:}
            \begin{verbatim}
def double_the_difference(lst):
    total = 0
    for num in lst:
        # Strict type check to exclude bools
        if type(num) is int:
            if num > 0 and num % 2 != 0:
                total += num ** 2
    return total
            \end{verbatim}
        \end{tcolorbox}
    \end{minipage}

    \vspace{2mm}
    \captionof{figure}{Case study from HumanEval.}
    \label{fig:case_humaneval}
\end{figure*}


\makeatletter
\providecommand{\captionof}[2]{\begingroup\def\@captype{#1}\caption{#2}\endgroup}
\makeatother

\definecolor{caseQHeader}{HTML}{7A7A7A}
\definecolor{caseQFrame}{HTML}{7A7A7A}
\definecolor{caseQBack}{HTML}{F2F2F2}
\definecolor{caseOursHeader}{HTML}{AED6FC}
\definecolor{caseOursFrame}{HTML}{0983F6}
\definecolor{caseBaseHeader}{HTML}{7EDDD2}
\definecolor{caseBaseFrame}{HTML}{2A9D8F}
\definecolor{caseOmit}{HTML}{E76F51}

\providecommand{\CaseOmit}[1]{\par\noindent\textcolor{caseOmit}{\textit{... (omitted: #1)}}\par}
\providecommand{\CaseAnswerBox}[1]{\begin{center}\fbox{\strut #1}\end{center}}
\providecommand{\CaseGTBox}[1]{\begin{center}\fbox{\strut #1}\end{center}}

\tcbset{
  casestudy/base/.style={
    enhanced,
    boxrule=0.5mm,
    arc=2mm,
    left=1.5mm,right=1.5mm,top=1.5mm,bottom=1.5mm,
    colback=white,
    coltitle=black,
    fonttitle=\bfseries\small,
    fontupper=\scriptsize,
    before skip=0pt,
    after skip=0pt,
    before upper={\setlength{\abovedisplayskip}{3pt}\setlength{\belowdisplayskip}{3pt}\setlength{\parskip}{0.2em}},
  },
  casestudy/question/.style={
    casestudy/base,
    colback=caseQBack,
    colframe=caseQFrame,
    colbacktitle=caseQHeader,
    coltitle=white,
  },
  casestudy/ours/.style={
    casestudy/base,
    colframe=caseOursFrame,
    colbacktitle=caseOursHeader,
    coltitle=black,
  },
  casestudy/baseline/.style={
    casestudy/base,
    colframe=caseBaseFrame,
    colbacktitle=caseBaseHeader,
    coltitle=black,
    before upper={\setlength{\abovedisplayskip}{4pt}\setlength{\belowdisplayskip}{4pt}\setlength{\parskip}{0.5em}},
  },
}

\begin{figure*}[t]
    \centering
    \begin{minipage}[t]{0.48\textwidth}
        \vspace{0pt}
        
        \begin{tcolorbox}[casestudy/question, title={Question and Examples}]
            \textbf{Problem:} There are \(N\) beans, each with deliciousness \(A_i\) and color \(C_i\). You choose one color, then eat one bean of that color. The deliciousness you get is the minimum among all beans of that color. Find the color that maximizes this minimum value.
            
            \vspace{0.5em}
            \textbf{Input:} \(N\) pairs of \((A_i, C_i)\).
            \textbf{Output:} The maximum possible value of the minimum deliciousness.
        \end{tcolorbox}
        
        \vspace{0.3cm}
        
        \begin{tcolorbox}[casestudy/ours, title={\textbf{Response from DOT}\hfill \textbf{999 Tokens}}]
            Okay, so I have this problem to solve. Let's see what it's about. Hmm.
            
            The problem is about choosing a color to eat a bean. Each bean has a deliciousness \(A_i\) and color \(C_i\). When you choose a color, you eat one bean of that color. But since multiple beans can have the same color, we have to pick the optimal color that maximizes the minimum deliciousness.
            
            So the goal is to select a color such that when you take all beans of that color, the minimum \(A_i\) among them is as large as possible. Then, the answer is the maximum of these minima.
            
            So the approach is:
            1. Group the beans by their color.
            2. For each color group, find the minimum \(A_i\) in that group.
            3. Find the maximum value among these minima.
            
            That's straightforward.
            
            \textbf{Code Implementation:}
            \begin{verbatim}
n = int(input())
from collections import defaultdict
groups = defaultdict(list)
for _ in range(n):
    a, c = map(int, input().split())
    groups[c].append(a)

min_per_color = [min(group) for group in groups.values()]
max_min = max(min_per_color)
print(max_min)
            \end{verbatim}
            
            Testing with Sample 1:
            Color 1: 100, 40 -> min 40.
            Color 5: 20, 30 -> min 20.
            Max of (40, 20) is 40. Correct.
            
            This approach efficiently groups the beans and computes the required values in linear time.
        \end{tcolorbox}
    \end{minipage}
    \hfill
    \begin{minipage}[t]{0.48\textwidth}
        \vspace{0pt}
        
        \begin{tcolorbox}[casestudy/baseline, title={\textbf{Response from R1-Distill-Qwen-7B}\hfill \textbf{12018 Tokens}}]
            Okay, so I've got this problem to solve. Let's see what it's about.
            The problem is about selecting a color of beans to maximize the minimum deliciousness.
            For each color, we look at all the beans of that color. The minimum deliciousness among those beans is what contributes to the overall maximum.
            
            \CaseOmit{initial confusion about binary search and threshold values}
            
            Maybe the optimal is somewhere in between. Perhaps we can think of this as trying to find a threshold value \(x\) such that there exists at least one color group with all beans having \(A_i \ge x\).
            
            Wait, maybe another approach. Let's collect all the colors, and for each color, compute the minimum \(A_i\) among its beans. Then, the answer is the maximum of these minima.
            Let's check Sample 2 with this logic.
            The minima are 68, 5, 2, 4, 4, 5, 2, 1, 3. The maximum is 68.
            But the sample output is 35. So that approach is incorrect.
            
            \CaseOmit{lengthy confusion about why Sample 2 output is 35, suspecting the max-of-min strategy is wrong}
            
            Wait, I might be misunderstanding the problem.
            Let me re-read Sample 2 carefully.
            Colors and values:
            68 (3), 17 (2), 99 (2), 92 (4), 82 (4), 10 (3), 100 (2), 78 (1), 3 (1), 35 (4).
            
            Let's group them properly:
            Color 3: 68, 10 -> Min is 10.
            Color 2: 17, 99, 100 -> Min is 17.
            Color 4: 92, 82, 35 -> Min is 35.
            Color 1: 78, 3 -> Min is 3.
            
            Minima list: 10, 17, 35, 3.
            Maximum of these is 35.
            
            Ah! My manual calculation earlier was wrong. I thought the minima were much different.
            Wait, previously I listed minima as "68, 5, 2..." where did those come from? I must have misread the input format or mixed up lines.
            
            \CaseOmit{re-verifying the logic multiple times, still doubting the simple solution because of previous calculation errors}
            
            So, let's confirm the logic again.
            The problem asks to maximize the minimum possible deliciousness.
            1. Group by color.
            2. Find min deliciousness for each group.
            3. Find max of those minimums.
            
            In Sample 2:
            - Color 3 group: {68, 10}. Min = 10.
            - Color 2 group: {17, 99, 100}. Min = 17.
            - Color 4 group: {92, 82, 35}. Min = 35.
            - Color 1 group: {78, 3}. Min = 3.
            
            Max(10, 17, 35, 3) = 35.
            This matches the sample output perfectly.
            
            So the initial "simple" approach was correct all along, I just failed to parse the sample input correctly in my head.
            The confusion came from misinterpreting the input lines or maybe grouping incorrectly in the thought process.
            
            \CaseOmit{final code construction using dictionary}
            
            Solution seems robust.
            \vspace{0.5em}
            \textbf{Solution Code:}
            \begin{verbatim}
import sys
from collections import defaultdict

def main():
    n = int(sys.stdin.readline())
    colors = defaultdict(list)
    for _ in range(n):
        a, c = map(int, sys.stdin.readline().split())
        colors[c].append(a)
    
    max_min = 0
    for c in colors:
        min_a = min(colors[c])
        if min_a > max_min:
            max_min = min_a
    print(max_min)

if __name__ == "__main__":
    main()
            \end{verbatim}
        \end{tcolorbox}
    \end{minipage}

    \vspace{2mm}
    \captionof{figure}{Case study from LiveCodeBench.}
    \label{fig:case_livecodebench}
\end{figure*}

%% file: figures/tex/figb.tex
\begin{figure*}[t]
    \centering
    \includegraphics[width=\textwidth]{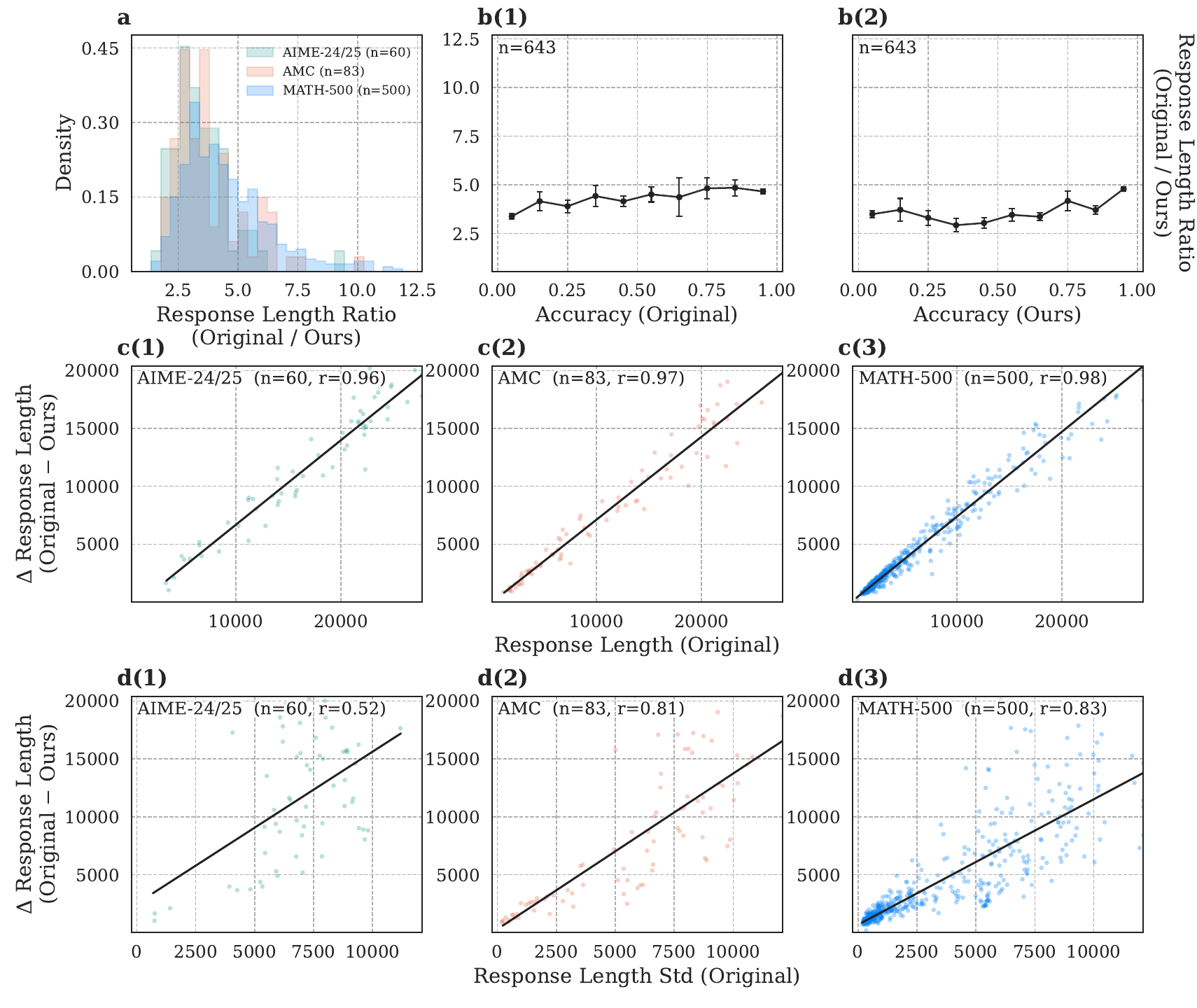}
    \caption{Quantitative analysis of reasoning efficiency for DOT-8K on \texttt{DeepSeek-R1-Distill-Qwen-1.5B}. (a) Density distribution of response length ratios between the original and DOT-optimized models. (b) Response length ratio across problems of varying difficulty, represented by original and our accuracy. (c) Token savings relative to the original response length. (d) Token savings relative to the standard deviation of the original response length.}
    \label{fig:figb1}
\end{figure*}
\begin{figure*}[t]
    \centering
    \includegraphics[width=\textwidth]{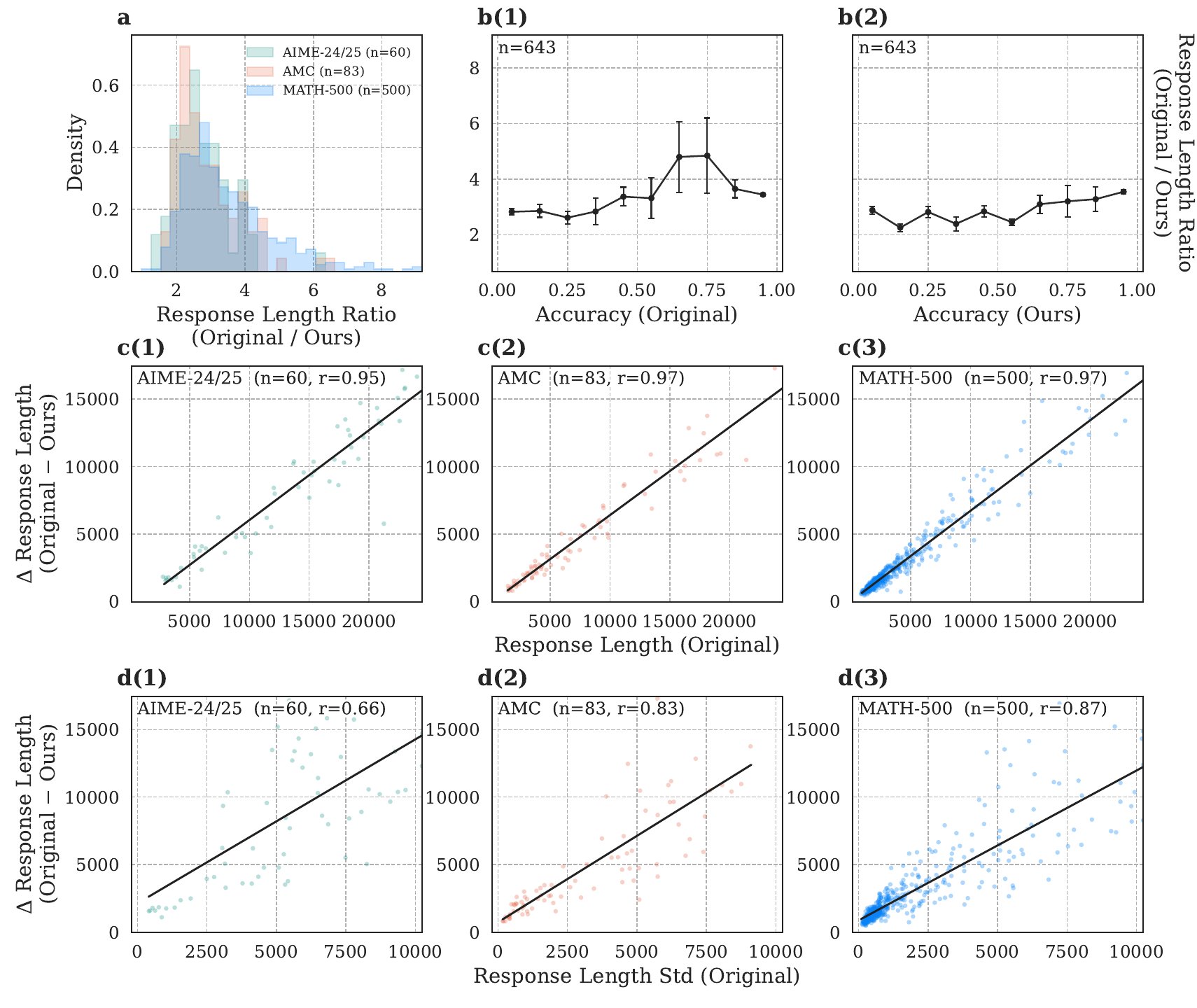}
    \caption{Quantitative analysis of reasoning efficiency for DOT-8K on \texttt{DeepSeek-R1-Distill-Qwen-7B}. (a) Density distribution of response length ratios between the original and DOT-optimized models. (b) Response length ratio across problems of varying difficulty, represented by original and our accuracy. (c) Token savings relative to the original response length. (d) Token savings relative to the standard deviation of the original response length.}
    \label{fig:figb2}
\end{figure*}
\begin{figure*}[t]
    \centering
    \includegraphics[width=\textwidth]{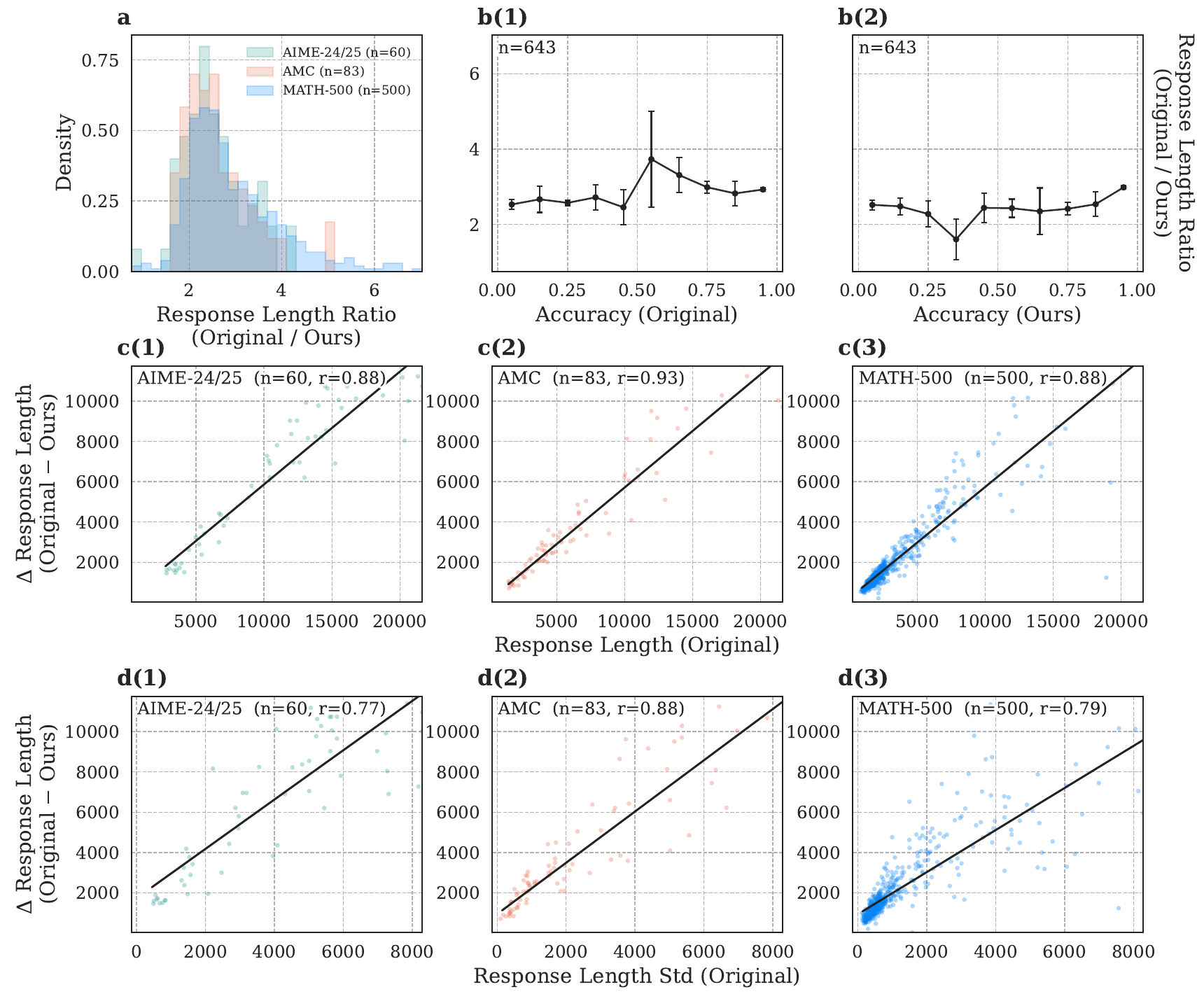}
    \caption{Quantitative analysis of reasoning efficiency for DOT-8K on \texttt{DeepSeek-R1-Distill-Qwen-32B}. (a) Density distribution of response length ratios between the original and DOT-optimized models. (b) Response length ratio across problems of varying difficulty, represented by original and our accuracy. (c) Token savings relative to the original response length. (d) Token savings relative to the standard deviation of the original response length.}
    \label{fig:figb3}
\end{figure*}

%% file: tables/tabled.tex
\begin{table}[t]
\centering
\renewcommand{\arraystretch}{0.925}
\resizebox{0.94\columnwidth}{!}{%
\begin{tabular}{lcccc}
\toprule
 & \multicolumn{2}{c}{GPQA} & \multicolumn{2}{c}{MMLU} \\
\cmidrule(lr){2-3} \cmidrule(lr){4-5}
\multirow{-2}{*}{Method} & Acc & Length & Acc & Length \\
\midrule
Original & 35.39 & 9826 & 49.34 & 2814 \\
\rowcolor{rowblue} DOT-4K (Ours) & 35.39 & \textbf{3036} & \underline{49.62} & \textbf{1300} \\
\rowcolor{rowblue} DOT-8K (Ours) & \textbf{36.58} & \underline{4791} & \textbf{49.68} & \underline{1709} \\
\bottomrule
\end{tabular}%
}
\caption{Performance comparison on general reasoning benchmarks (GPQA and MMLU).}
\label{tab:tabled}
\end{table}

%% file: tables/tablee.tex
\begin{table*}[t]
\centering
\renewcommand{\arraystretch}{0.925}
\resizebox{0.65\textwidth}{!}{%
\begin{tabular}{lcccc}
\toprule
 & \multicolumn{2}{c}{AlpacaEval 2.0} & \multicolumn{2}{c}{ArenaHard} \\
\cmidrule(lr){2-3} \cmidrule(lr){4-5}
\multirow{-2}{*}{Method} & LC-Winrate & Length & Score(GPT-4-1106) & Length \\
\midrule
Original & 1.74 & 9493 & 1.28 & 16828 \\
\rowcolor{rowblue} DOT-4K (Ours) & \underline{1.99} & \textbf{1887} & \textbf{1.46} & \textbf{4635} \\
\rowcolor{rowblue} DOT-8K (Ours) & \textbf{2.98} & \underline{2812} & \underline{1.32} & \underline{6454} \\
\bottomrule
\end{tabular}%
}
\caption{Performance comparison onpen-ended evaluations (AlpacaEval 2.0 and ArenaHard).}
\label{tab:tablee}
\end{table*}

%% file: tables/tablef.tex
\begin{table*}[t]
\centering
\renewcommand{\arraystretch}{0.95}
\resizebox{0.85\textwidth}{!}{%
\begin{tabular}{lcccccccc}
\toprule
 & \multicolumn{2}{c}{AIME-24} & \multicolumn{2}{c}{AIME-25} & \multicolumn{2}{c}{AMC} & \multicolumn{2}{c}{MATH-500} \\
\cmidrule(lr){2-3} \cmidrule(lr){4-5} \cmidrule(lr){6-7} \cmidrule(lr){8-9}
\multirow{-2}{*}{Method} & Acc & Length & Acc & Length & Acc & Length & Acc & Length \\
\midrule
Original & \underline{73.41} & 14407 & \underline{65.83} & 17307 & \underline{88.59} & 9973 & 94.35 & 5197 \\
\rowcolor{rowblue} DOT-16K (Ours) & 73.19 & \textbf{8983} & 64.33 & \textbf{10629} & 88.44 & \textbf{5050} & \underline{94.98} & \textbf{2072} \\
\rowcolor{rowblue} DOT-32K (Ours) & \textbf{74.06} & \underline{11487} & \textbf{67.70} & \underline{13099} & \textbf{89.34} & \underline{6022} & \textbf{95.03} & \underline{2203} \\
\bottomrule
\end{tabular}%
}
\caption{Performance comparison on AIME-24, AIME-25, AMC, and MATH-500 benchmarks with \texttt{Qwen3-4B}.}
\label{tab:tablef}
\end{table*}

%% file: tables/tableg.tex
\begin{table}[t]
\centering
\renewcommand{\arraystretch}{0.95}
\resizebox{0.88\columnwidth}{!}{%
\begin{tabular}{lccc}
\toprule
Method & Score & Length & Steps \\
\midrule
Original & 64.86 & 1415 & 3.03 \\
\rowcolor{rowblue} DOT-16K (Ours) & 63.89 & \textbf{1034} & \underline{3.07} \\
\rowcolor{rowblue} DOT-32K (Ours) & \textbf{64.98} & \underline{1118} & \textbf{3.03} \\
\bottomrule
\end{tabular}%
}
\vspace{-5pt}
\caption{Agentic evaluation on BFCL v3.}
\vspace{-15pt}
\label{tab:tableg}
\end{table}

%% file: tables/tableh.tex
\begin{table*}[t]
\centering
\renewcommand{\arraystretch}{0.95}
\resizebox{0.65\textwidth}{!}{%
\begin{tabular}{lcccc}
\toprule
 & \multicolumn{2}{c}{AIME-24} & \multicolumn{2}{c}{MATH-500} \\
\cmidrule(lr){2-3} \cmidrule(lr){4-5}
\multirow{-2}{*}{Method} & Acc & Length & Acc & Length \\
\midrule
TCC~\cite{TCC} & 48.4 & 10603 & 89.2 & 3864 \\
CoD~\cite{CoD} & 40.0 & 10519 & 83.2 & 1987 \\
NoThinking~\cite{NoThinking} & 26.7 & 4427 & 80.6 & 834 \\
Dynasor-CoT~\cite{Dynasor-CoT} & 46.7 & 12695 & 89.0 & 2971 \\
SEAL~\cite{SEAL} & --- & --- & 89.4 & 2661 \\
DEER~\cite{DEER} & 49.2 & 9839 & 89.8 & 2143 \\
DEER-Pro~\cite{DEER} & 49.2 & 10046 & 90.2 & 2391 \\
\rowcolor{rowblue} DOT-4K (Ours) & \underline{54.8} & \textbf{2958} & \underline{93.4} & \textbf{1008} \\
\rowcolor{rowblue} DOT-8K (Ours) & \textbf{62.6} & \underline{4903} & \textbf{94.3} & \underline{1293} \\
\bottomrule
\end{tabular}%
}
\caption{Comparison with training-free methods on \texttt{DeepSeek-R1-Distill-Qwen-7B}.}
\label{tab:tableh}
\end{table*}

%% file: figures/tex/figc.tex
\begin{figure*}[t]
    \centering
    \includegraphics[width=\textwidth]{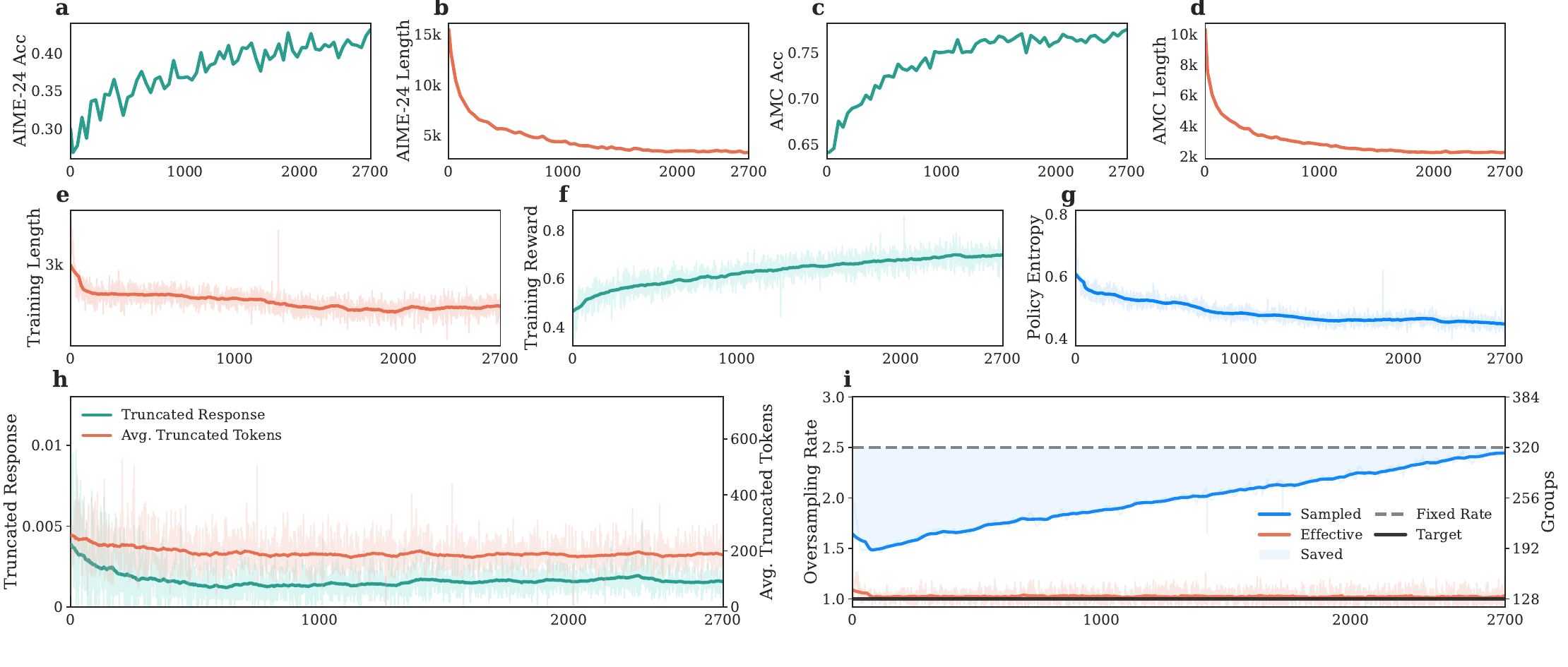}
    \caption{Metric curves monitoring the DOT-4K training process on \texttt{DeepSeek-R1-Distill-Qwen-1.5B}. \\(a, c) Evolution of pass@1 accuracy on AIME-24 and AMC. (b, d) Decrease in average response length for corresponding benchmarks. (e) Global training length reduction. (f) Progression of average training reward. (g) Controlled policy entropy decline. (h) Frequency of truncated responses and average truncated tokens per response. (i) Stability of oversampling rate and effective batch size via Predictive Dynamic Sampling}
    \label{fig:figc1}
\end{figure*}

\begin{figure*}[t]
    \centering
    \includegraphics[width=\textwidth]{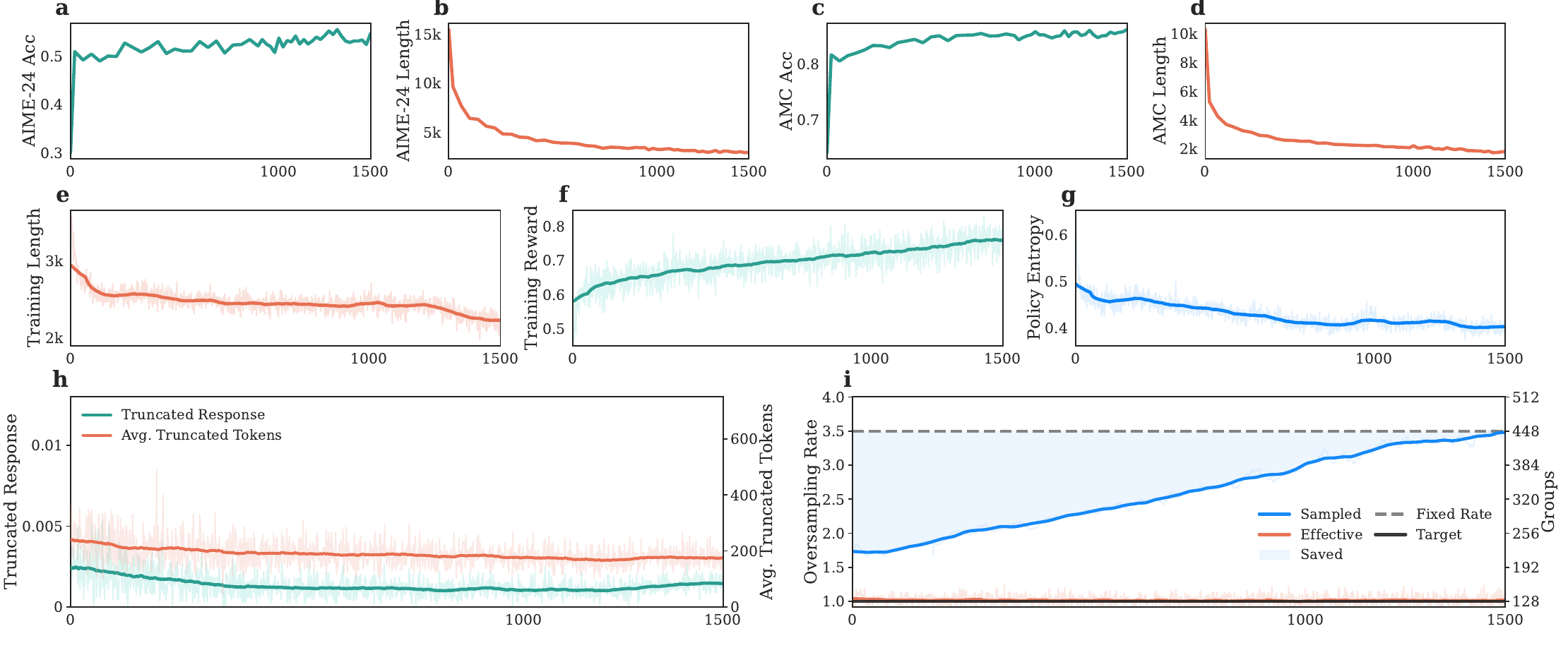}
    \caption{Metric curves monitoring the DOT-4K training process on \texttt{DeepSeek-R1-Distill-Qwen-7B}. \\(a, c) Evolution of pass@1 accuracy on AIME-24 and AMC. (b, d) Decrease in average response length for corresponding benchmarks. (e) Global training length reduction. (f) Progression of average training reward. (g) Controlled policy entropy decline. (h) Frequency of truncated responses and average truncated tokens per response. (i) Stability of oversampling rate and effective batch size via Predictive Dynamic Sampling}
    \label{fig:figc2}
\end{figure*}

\begin{figure*}[t]
    \centering
    \includegraphics[width=\textwidth]{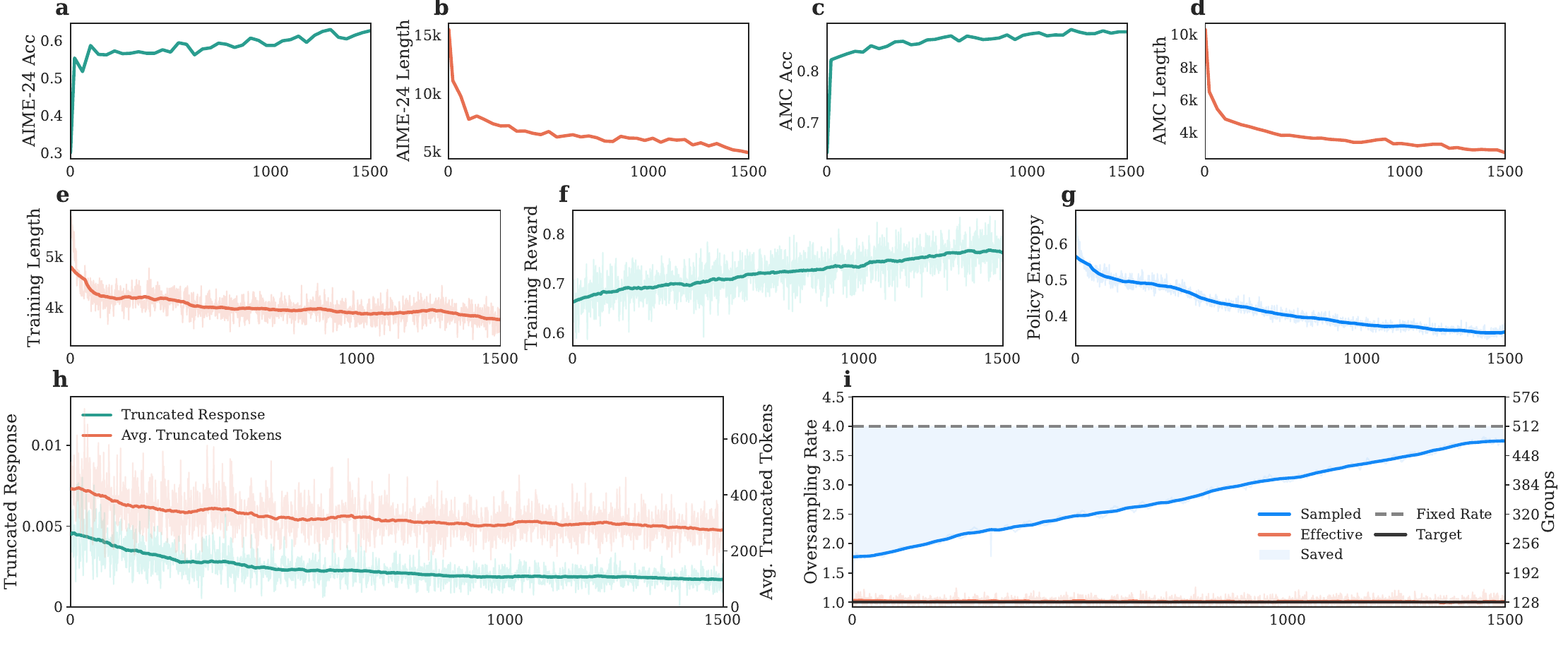}
    \caption{Metric curves monitoring the DOT-8K training process on \texttt{DeepSeek-R1-Distill-Qwen-7B}. \\(a, c) Evolution of pass@1 accuracy on AIME-24 and AMC. (b, d) Decrease in average response length for corresponding benchmarks. (e) Global training length reduction. (f) Progression of average training reward. (g) Controlled policy entropy decline. (h) Frequency of truncated responses and average truncated tokens per response. (i) Stability of oversampling rate and effective batch size via Predictive Dynamic Sampling}
    \label{fig:figc3}
\end{figure*}

\begin{figure*}[t]
    \centering
    \includegraphics[width=\textwidth]{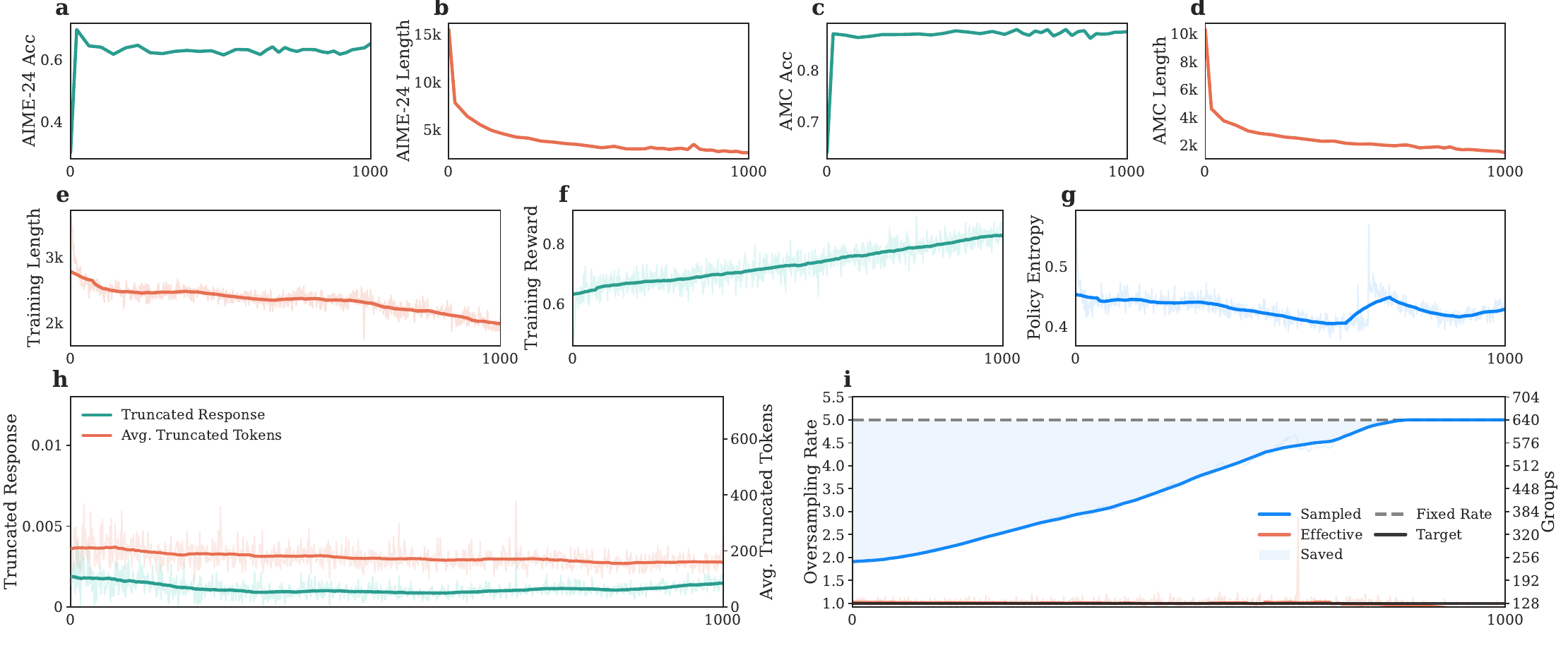}
    \caption{Metric curves monitoring the DOT-4K training process on \texttt{DeepSeek-R1-Distill-Qwen-32B}. \\(a, c) Evolution of pass@1 accuracy on AIME-24 and AMC. (b, d) Decrease in average response length for corresponding benchmarks. (e) Global training length reduction. (f) Progression of average training reward. (g) Controlled policy entropy decline. (h) Frequency of truncated responses and average truncated tokens per response. (i) Stability of oversampling rate and effective batch size via Predictive Dynamic Sampling}
    \label{fig:figc4}
\end{figure*}

\begin{figure*}[t]
    \centering
    \includegraphics[width=\textwidth]{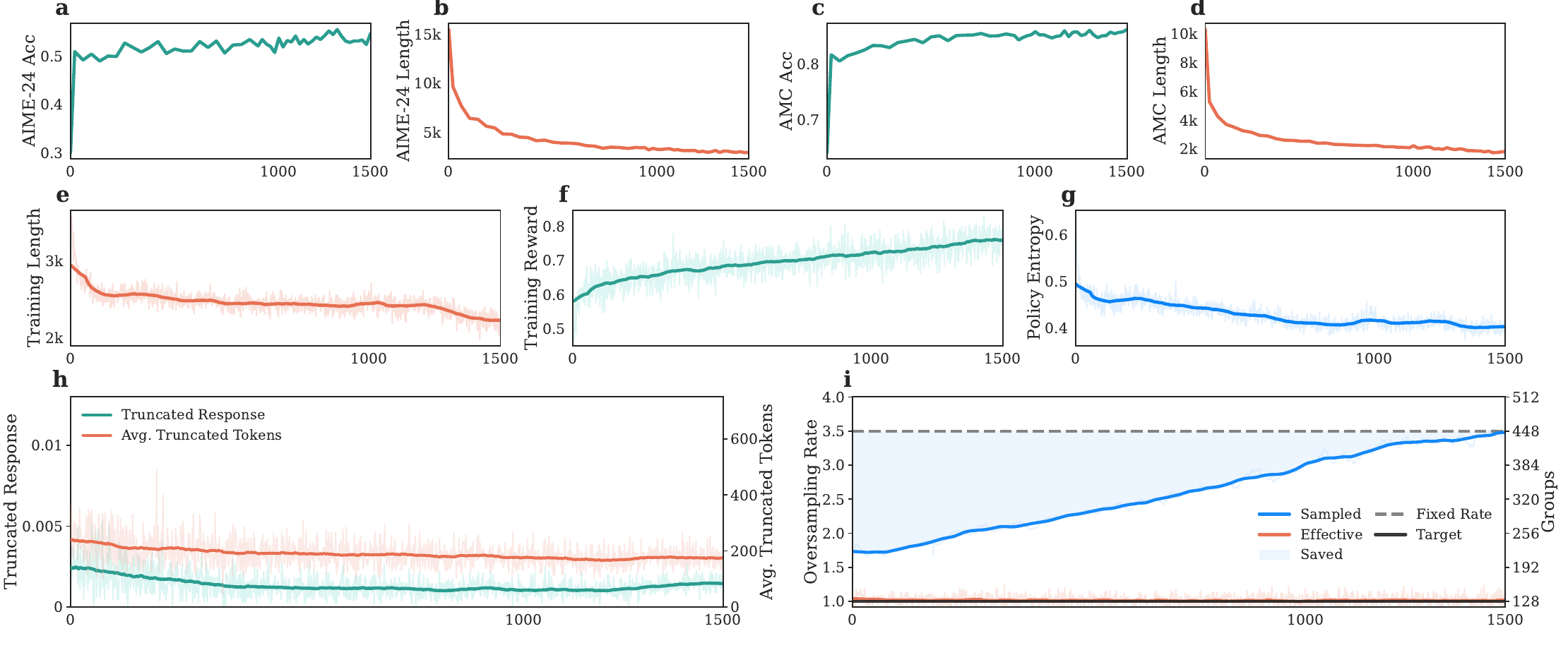}
    \caption{Metric curves monitoring the DOT-8K training process on \texttt{DeepSeek-R1-Distill-Qwen-32B}. \\(a, c) Evolution of pass@1 accuracy on AIME-24 and AMC. (b, d) Decrease in average response length for corresponding benchmarks. (e) Global training length reduction. (f) Progression of average training reward. (g) Controlled policy entropy decline. (h) Frequency of truncated responses and average truncated tokens per response. (i) Stability of oversampling rate and effective batch size via Predictive Dynamic Sampling}
    \label{fig:figc5}
\end{figure*}

%% file: tables/tableb.tex
\begin{table*}[t]
\centering
\small
\begin{tabular}{l c}
\toprule
\textbf{Parameter} & \textbf{Value}\\
\midrule
\texttt{algorithm.adv\_estimator} & grpo\\
\texttt{actor\_rollout\_ref.actor.loss\_agg\_mode} & seq-mean-token-mean\\
\texttt{actor\_rollout\_ref.actor.policy\_loss.loss\_mode} & kl\_cov\\
\texttt{actor\_rollout\_ref.actor.policy\_loss.kl\_cov\_ratio} & 0.002\\
\texttt{actor\_rollout\_ref.actor.policy\_loss.ppo\_kl\_coef} & 1.0\\
\midrule
\texttt{data.train\_batch\_size} & 128\\
\texttt{actor\_rollout\_ref.actor.ppo\_mini\_batch\_size} & 32\\
\texttt{actor\_rollout\_ref.actor.ppo\_epochs} & 1\\
\midrule
\texttt{data.max\_prompt\_length} & 4096\\
\texttt{data.max\_response\_length} & 4096/8192\\
\midrule
\texttt{actor\_rollout\_ref.actor.optim.lr} & $1.0\times10^{-6}$\\
\texttt{actor\_rollout\_ref.rollout.temperature} & 1.0\\
\texttt{actor\_rollout\_ref.rollout.n} & 32\\
\midrule
\texttt{actor\_rollout\_ref.actor.clip\_ratio\_low} & 0.2\\
\texttt{actor\_rollout\_ref.actor.clip\_ratio\_high} & 0.2\\
\bottomrule
\end{tabular}
\caption{Training configuration specified in verl.}
\label{tab:tableb}
\end{table*}

%% file: tables/tablei.tex
\begin{table*}[t]
\centering
\renewcommand{\arraystretch}{0.95}
\resizebox{\textwidth}{!}{%
\begin{tabular}{lccccc}
\toprule
Method & Total Training Time & Steps & Avg Time / Step & Avg Policy Update / Step & Avg Rollout / Step \\
\midrule
GRPO & 11d 1h 6m 19s & 2900 & 5m 29s & 2m 4s & 2m 57s \\
\rowcolor{rowblue} DOT-4K (Ours) & \textbf{5d 8h 34m} & 2850 & \textbf{2m 42s} & \textbf{1m 2s} & \textbf{1m 4s} \\
\rowcolor{rowblue} DOT-8K (Ours) & \underline{7d 10h 34m} & 2940 & \underline{3m 39s} & \underline{1m 35s} & \underline{1m 45s} \\
\bottomrule
\end{tabular}%
}
\caption{Training latency comparison on \texttt{DeepSeek-R1-Distill-Qwen-1.5B}.}
\label{tab:tablei}
\end{table*}

%% file: tables/tablec.tex
\begin{table*}[t]
    \centering
    \small
    \begin{tabular}{llllll}
        \toprule
        \multicolumn{6}{c}{\textbf{Reasoning Words}} \\
        \midrule
        aha & all & alternatively & although & analyze & appears \\
        assume & but & check & clearly & compare & confirm \\
        consider & contradiction & correct & define & ensure & even \\
        finally & hence & hmm & hold on & however & if \\
        implies & initially & instead & instead of & invalid & let \\
        maybe & might be & moreover & need to find & next & oh \\
        oops & otherwise & perhaps & reflect & reflection & so \\
        some & split & suggests & then & therefore & this means that \\
        thus & valid & verify & wait & well & \\
        \bottomrule
    \end{tabular}
    \caption{The full list of reasoning words used in our empirical analysis.}
    \label{tab:tablec}
\end{table*}

%% file: custom.bib
@misc{tts,
      title={Scaling LLM Test-Time Compute Optimally can be More Effective than Scaling Model Parameters}, 
      author={Charlie Snell and Jaehoon Lee and Kelvin Xu and Aviral Kumar},
      year={2024},
      eprint={2408.03314},
      archivePrefix={arXiv},
      primaryClass={cs.LG},
      url={https://arxiv.org/abs/2408.03314}, 
}

@misc{o1,
      title={OpenAI o1 System Card}, 
      author={OpenAI and : and Aaron Jaech and Adam Kalai and Adam Lerer and Adam Richardson and Ahmed El-Kishky and Aiden Low and Alec Helyar and Aleksander Madry and Alex Beutel and Alex Carney and Alex Iftimie and Alex Karpenko and Alex Tachard Passos and Alexander Neitz and Alexander Prokofiev and Alexander Wei and Allison Tam and Ally Bennett and Ananya Kumar and Andre Saraiva and Andrea Vallone and Andrew Duberstein and Andrew Kondrich and Andrey Mishchenko and Andy Applebaum and Angela Jiang and Ashvin Nair and Barret Zoph and Behrooz Ghorbani and Ben Rossen and Benjamin Sokolowsky and Boaz Barak and Bob McGrew and Borys Minaiev and Botao Hao and Bowen Baker and Brandon Houghton and Brandon McKinzie and Brydon Eastman and Camillo Lugaresi and Cary Bassin and Cary Hudson and Chak Ming Li and Charles de Bourcy and Chelsea Voss and Chen Shen and Chong Zhang and Chris Koch and Chris Orsinger and Christopher Hesse and Claudia Fischer and Clive Chan and Dan Roberts and Daniel Kappler and Daniel Levy and Daniel Selsam and David Dohan and David Farhi and David Mely and David Robinson and Dimitris Tsipras and Doug Li and Dragos Oprica and Eben Freeman and Eddie Zhang and Edmund Wong and Elizabeth Proehl and Enoch Cheung and Eric Mitchell and Eric Wallace and Erik Ritter and Evan Mays and Fan Wang and Felipe Petroski Such and Filippo Raso and Florencia Leoni and Foivos Tsimpourlas and Francis Song and Fred von Lohmann and Freddie Sulit and Geoff Salmon and Giambattista Parascandolo and Gildas Chabot and Grace Zhao and Greg Brockman and Guillaume Leclerc and Hadi Salman and Haiming Bao and Hao Sheng and Hart Andrin and Hessam Bagherinezhad and Hongyu Ren and Hunter Lightman and Hyung Won Chung and Ian Kivlichan and Ian O'Connell and Ian Osband and Ignasi Clavera Gilaberte and Ilge Akkaya and Ilya Kostrikov and Ilya Sutskever and Irina Kofman and Jakub Pachocki and James Lennon and Jason Wei and Jean Harb and Jerry Twore and Jiacheng Feng and Jiahui Yu and Jiayi Weng and Jie Tang and Jieqi Yu and Joaquin Quiñonero Candela and Joe Palermo and Joel Parish and Johannes Heidecke and John Hallman and John Rizzo and Jonathan Gordon and Jonathan Uesato and Jonathan Ward and Joost Huizinga and Julie Wang and Kai Chen and Kai Xiao and Karan Singhal and Karina Nguyen and Karl Cobbe and Katy Shi and Kayla Wood and Kendra Rimbach and Keren Gu-Lemberg and Kevin Liu and Kevin Lu and Kevin Stone and Kevin Yu and Lama Ahmad and Lauren Yang and Leo Liu and Leon Maksin and Leyton Ho and Liam Fedus and Lilian Weng and Linden Li and Lindsay McCallum and Lindsey Held and Lorenz Kuhn and Lukas Kondraciuk and Lukasz Kaiser and Luke Metz and Madelaine Boyd and Maja Trebacz and Manas Joglekar and Mark Chen and Marko Tintor and Mason Meyer and Matt Jones and Matt Kaufer and Max Schwarzer and Meghan Shah and Mehmet Yatbaz and Melody Y. Guan and Mengyuan Xu and Mengyuan Yan and Mia Glaese and Mianna Chen and Michael Lampe and Michael Malek and Michele Wang and Michelle Fradin and Mike McClay and Mikhail Pavlov and Miles Wang and Mingxuan Wang and Mira Murati and Mo Bavarian and Mostafa Rohaninejad and Nat McAleese and Neil Chowdhury and Neil Chowdhury and Nick Ryder and Nikolas Tezak and Noam Brown and Ofir Nachum and Oleg Boiko and Oleg Murk and Olivia Watkins and Patrick Chao and Paul Ashbourne and Pavel Izmailov and Peter Zhokhov and Rachel Dias and Rahul Arora and Randall Lin and Rapha Gontijo Lopes and Raz Gaon and Reah Miyara and Reimar Leike and Renny Hwang and Rhythm Garg and Robin Brown and Roshan James and Rui Shu and Ryan Cheu and Ryan Greene and Saachi Jain and Sam Altman and Sam Toizer and Sam Toyer and Samuel Miserendino and Sandhini Agarwal and Santiago Hernandez and Sasha Baker and Scott McKinney and Scottie Yan and Shengjia Zhao and Shengli Hu and Shibani Santurkar and Shraman Ray Chaudhuri and Shuyuan Zhang and Siyuan Fu and Spencer Papay and Steph Lin and Suchir Balaji and Suvansh Sanjeev and Szymon Sidor and Tal Broda and Aidan Clark and Tao Wang and Taylor Gordon and Ted Sanders and Tejal Patwardhan and Thibault Sottiaux and Thomas Degry and Thomas Dimson and Tianhao Zheng and Timur Garipov and Tom Stasi and Trapit Bansal and Trevor Creech and Troy Peterson and Tyna Eloundou and Valerie Qi and Vineet Kosaraju and Vinnie Monaco and Vitchyr Pong and Vlad Fomenko and Weiyi Zheng and Wenda Zhou and Wes McCabe and Wojciech Zaremba and Yann Dubois and Yinghai Lu and Yining Chen and Young Cha and Yu Bai and Yuchen He and Yuchen Zhang and Yunyun Wang and Zheng Shao and Zhuohan Li},
      year={2024},
      eprint={2412.16720},
      archivePrefix={arXiv},
      primaryClass={cs.AI},
      url={https://arxiv.org/abs/2412.16720}, 
}

@misc{r1,
      title={DeepSeek-R1: Incentivizing Reasoning Capability in LLMs via Reinforcement Learning}, 
      author={DeepSeek-AI and Daya Guo and Dejian Yang and Haowei Zhang and Junxiao Song and Ruoyu Zhang and Runxin Xu and Qihao Zhu and Shirong Ma and Peiyi Wang and Xiao Bi and Xiaokang Zhang and Xingkai Yu and Yu Wu and Z. F. Wu and Zhibin Gou and Zhihong Shao and Zhuoshu Li and Ziyi Gao and Aixin Liu and Bing Xue and Bingxuan Wang and Bochao Wu and Bei Feng and Chengda Lu and Chenggang Zhao and Chengqi Deng and Chenyu Zhang and Chong Ruan and Damai Dai and Deli Chen and Dongjie Ji and Erhang Li and Fangyun Lin and Fucong Dai and Fuli Luo and Guangbo Hao and Guanting Chen and Guowei Li and H. Zhang and Han Bao and Hanwei Xu and Haocheng Wang and Honghui Ding and Huajian Xin and Huazuo Gao and Hui Qu and Hui Li and Jianzhong Guo and Jiashi Li and Jiawei Wang and Jingchang Chen and Jingyang Yuan and Junjie Qiu and Junlong Li and J. L. Cai and Jiaqi Ni and Jian Liang and Jin Chen and Kai Dong and Kai Hu and Kaige Gao and Kang Guan and Kexin Huang and Kuai Yu and Lean Wang and Lecong Zhang and Liang Zhao and Litong Wang and Liyue Zhang and Lei Xu and Leyi Xia and Mingchuan Zhang and Minghua Zhang and Minghui Tang and Meng Li and Miaojun Wang and Mingming Li and Ning Tian and Panpan Huang and Peng Zhang and Qiancheng Wang and Qinyu Chen and Qiushi Du and Ruiqi Ge and Ruisong Zhang and Ruizhe Pan and Runji Wang and R. J. Chen and R. L. Jin and Ruyi Chen and Shanghao Lu and Shangyan Zhou and Shanhuang Chen and Shengfeng Ye and Shiyu Wang and Shuiping Yu and Shunfeng Zhou and Shuting Pan and S. S. Li and Shuang Zhou and Shaoqing Wu and Shengfeng Ye and Tao Yun and Tian Pei and Tianyu Sun and T. Wang and Wangding Zeng and Wanjia Zhao and Wen Liu and Wenfeng Liang and Wenjun Gao and Wenqin Yu and Wentao Zhang and W. L. Xiao and Wei An and Xiaodong Liu and Xiaohan Wang and Xiaokang Chen and Xiaotao Nie and Xin Cheng and Xin Liu and Xin Xie and Xingchao Liu and Xinyu Yang and Xinyuan Li and Xuecheng Su and Xuheng Lin and X. Q. Li and Xiangyue Jin and Xiaojin Shen and Xiaosha Chen and Xiaowen Sun and Xiaoxiang Wang and Xinnan Song and Xinyi Zhou and Xianzu Wang and Xinxia Shan and Y. K. Li and Y. Q. Wang and Y. X. Wei and Yang Zhang and Yanhong Xu and Yao Li and Yao Zhao and Yaofeng Sun and Yaohui Wang and Yi Yu and Yichao Zhang and Yifan Shi and Yiliang Xiong and Ying He and Yishi Piao and Yisong Wang and Yixuan Tan and Yiyang Ma and Yiyuan Liu and Yongqiang Guo and Yuan Ou and Yuduan Wang and Yue Gong and Yuheng Zou and Yujia He and Yunfan Xiong and Yuxiang Luo and Yuxiang You and Yuxuan Liu and Yuyang Zhou and Y. X. Zhu and Yanhong Xu and Yanping Huang and Yaohui Li and Yi Zheng and Yuchen Zhu and Yunxian Ma and Ying Tang and Yukun Zha and Yuting Yan and Z. Z. Ren and Zehui Ren and Zhangli Sha and Zhe Fu and Zhean Xu and Zhenda Xie and Zhengyan Zhang and Zhewen Hao and Zhicheng Ma and Zhigang Yan and Zhiyu Wu and Zihui Gu and Zijia Zhu and Zijun Liu and Zilin Li and Ziwei Xie and Ziyang Song and Zizheng Pan and Zhen Huang and Zhipeng Xu and Zhongyu Zhang and Zhen Zhang},
      year={2025},
      eprint={2501.12948},
      archivePrefix={arXiv},
      primaryClass={cs.CL},
      url={https://arxiv.org/abs/2501.12948}, 
}

@misc{qwen,
      title={Qwen3 Technical Report}, 
      author={An Yang and Anfeng Li and Baosong Yang and Beichen Zhang and Binyuan Hui and Bo Zheng and Bowen Yu and Chang Gao and Chengen Huang and Chenxu Lv and Chujie Zheng and Dayiheng Liu and Fan Zhou and Fei Huang and Feng Hu and Hao Ge and Haoran Wei and Huan Lin and Jialong Tang and Jian Yang and Jianhong Tu and Jianwei Zhang and Jianxin Yang and Jiaxi Yang and Jing Zhou and Jingren Zhou and Junyang Lin and Kai Dang and Keqin Bao and Kexin Yang and Le Yu and Lianghao Deng and Mei Li and Mingfeng Xue and Mingze Li and Pei Zhang and Peng Wang and Qin Zhu and Rui Men and Ruize Gao and Shixuan Liu and Shuang Luo and Tianhao Li and Tianyi Tang and Wenbiao Yin and Xingzhang Ren and Xinyu Wang and Xinyu Zhang and Xuancheng Ren and Yang Fan and Yang Su and Yichang Zhang and Yinger Zhang and Yu Wan and Yuqiong Liu and Zekun Wang and Zeyu Cui and Zhenru Zhang and Zhipeng Zhou and Zihan Qiu},
      year={2025},
      eprint={2505.09388},
      archivePrefix={arXiv},
      primaryClass={cs.CL},
      url={https://arxiv.org/abs/2505.09388}, 
}

@misc{interplay,
      title={On the Interplay of Pre-Training, Mid-Training, and RL on Reasoning Language Models}, 
      author={Charlie Zhang and Graham Neubig and Xiang Yue},
      year={2025},
      eprint={2512.07783},
      archivePrefix={arXiv},
      primaryClass={cs.CL},
      url={https://arxiv.org/abs/2512.07783}, 
}

@inproceedings{cot,
 author = {Wei, Jason and Wang, Xuezhi and Schuurmans, Dale and Bosma, Maarten and ichter, brian and Xia, Fei and Chi, Ed and Le, Quoc V and Zhou, Denny},
 booktitle = {Advances in Neural Information Processing Systems},
 editor = {S. Koyejo and S. Mohamed and A. Agarwal and D. Belgrave and K. Cho and A. Oh},
 pages = {24824--24837},
 publisher = {Curran Associates, Inc.},
 title = {Chain-of-Thought Prompting Elicits Reasoning in Large Language Models},
 url = {https://proceedings.neurips.cc/paper_files/paper/2022/file/9d5609613524ecf4f15af0f7b31abca4-Paper-Conference.pdf},
 volume = {35},
 year = {2022}
}

@misc{overthinking,
      title={Do NOT Think That Much for 2+3=? On the Overthinking of o1-Like LLMs}, 
      author={Xingyu Chen and Jiahao Xu and Tian Liang and Zhiwei He and Jianhui Pang and Dian Yu and Linfeng Song and Qiuzhi Liu and Mengfei Zhou and Zhuosheng Zhang and Rui Wang and Zhaopeng Tu and Haitao Mi and Dong Yu},
      year={2025},
      eprint={2412.21187},
      archivePrefix={arXiv},
      primaryClass={cs.CL},
      url={https://arxiv.org/abs/2412.21187}, 
}

@misc{o1-pruner,
      title={O1-Pruner: Length-Harmonizing Fine-Tuning for O1-Like Reasoning Pruning}, 
      author={Haotian Luo and Li Shen and Haiying He and Yibo Wang and Shiwei Liu and Wei Li and Naiqiang Tan and Xiaochun Cao and Dacheng Tao},
      year={2025},
      eprint={2501.12570},
      archivePrefix={arXiv},
      primaryClass={cs.CL},
      url={https://arxiv.org/abs/2501.12570}, 
}

@misc{kimi-1.5,
      title={Kimi k1.5: Scaling Reinforcement Learning with LLMs}, 
      author={Kimi Team and Angang Du and Bofei Gao and Bowei Xing and Changjiu Jiang and Cheng Chen and Cheng Li and Chenjun Xiao and Chenzhuang Du and Chonghua Liao and Chuning Tang and Congcong Wang and Dehao Zhang and Enming Yuan and Enzhe Lu and Fengxiang Tang and Flood Sung and Guangda Wei and Guokun Lai and Haiqing Guo and Han Zhu and Hao Ding and Hao Hu and Hao Yang and Hao Zhang and Haotian Yao and Haotian Zhao and Haoyu Lu and Haoze Li and Haozhen Yu and Hongcheng Gao and Huabin Zheng and Huan Yuan and Jia Chen and Jianhang Guo and Jianlin Su and Jianzhou Wang and Jie Zhao and Jin Zhang and Jingyuan Liu and Junjie Yan and Junyan Wu and Lidong Shi and Ling Ye and Longhui Yu and Mengnan Dong and Neo Zhang and Ningchen Ma and Qiwei Pan and Qucheng Gong and Shaowei Liu and Shengling Ma and Shupeng Wei and Sihan Cao and Siying Huang and Tao Jiang and Weihao Gao and Weimin Xiong and Weiran He and Weixiao Huang and Weixin Xu and Wenhao Wu and Wenyang He and Xianghui Wei and Xianqing Jia and Xingzhe Wu and Xinran Xu and Xinxing Zu and Xinyu Zhou and Xuehai Pan and Y. Charles and Yang Li and Yangyang Hu and Yangyang Liu and Yanru Chen and Yejie Wang and Yibo Liu and Yidao Qin and Yifeng Liu and Ying Yang and Yiping Bao and Yulun Du and Yuxin Wu and Yuzhi Wang and Zaida Zhou and Zhaoji Wang and Zhaowei Li and Zhen Zhu and Zheng Zhang and Zhexu Wang and Zhilin Yang and Zhiqi Huang and Zihao Huang and Ziyao Xu and Zonghan Yang and Zongyu Lin},
      year={2025},
      eprint={2501.12599},
      archivePrefix={arXiv},
      primaryClass={cs.AI},
      url={https://arxiv.org/abs/2501.12599}, 
}

@misc{DAST,
      title={DAST: Difficulty-Adaptive Slow-Thinking for Large Reasoning Models}, 
      author={Yi Shen and Jian Zhang and Jieyun Huang and Shuming Shi and Wenjing Zhang and Jiangze Yan and Ning Wang and Kai Wang and Zhaoxiang Liu and Shiguo Lian},
      year={2025},
      eprint={2503.04472},
      archivePrefix={arXiv},
      primaryClass={cs.LG},
      url={https://arxiv.org/abs/2503.04472}, 
}

@misc{LCR,
      title={Optimizing Length Compression in Large Reasoning Models}, 
      author={Zhengxiang Cheng and Dongping Chen and Mingyang Fu and Tianyi Zhou},
      year={2025},
      eprint={2506.14755},
      archivePrefix={arXiv},
      primaryClass={cs.AI},
      url={https://arxiv.org/abs/2506.14755}, 
}

@misc{L1,
      title={L1: Controlling How Long A Reasoning Model Thinks With Reinforcement Learning}, 
      author={Pranjal Aggarwal and Sean Welleck},
      year={2025},
      eprint={2503.04697},
      archivePrefix={arXiv},
      primaryClass={cs.CL},
      url={https://arxiv.org/abs/2503.04697}, 
}

@misc{ThinkPrune,
      title={ThinkPrune: Pruning Long Chain-of-Thought of LLMs via Reinforcement Learning}, 
      author={Bairu Hou and Yang Zhang and Jiabao Ji and Yujian Liu and Kaizhi Qian and Jacob Andreas and Shiyu Chang},
      year={2025},
      eprint={2504.01296},
      archivePrefix={arXiv},
      primaryClass={cs.CL},
      url={https://arxiv.org/abs/2504.01296}, 
}

@misc{AdaptThink,
      title={AdaptThink: Reasoning Models Can Learn When to Think}, 
      author={Jiajie Zhang and Nianyi Lin and Lei Hou and Ling Feng and Juanzi Li},
      year={2025},
      eprint={2505.13417},
      archivePrefix={arXiv},
      primaryClass={cs.CL},
      url={https://arxiv.org/abs/2505.13417}, 
}

@misc{LASER,
      title={Learn to Reason Efficiently with Adaptive Length-based Reward Shaping}, 
      author={Wei Liu and Ruochen Zhou and Yiyun Deng and Yuzhen Huang and Junteng Liu and Yuntian Deng and Yizhe Zhang and Junxian He},
      year={2025},
      eprint={2505.15612},
      archivePrefix={arXiv},
      primaryClass={cs.CL},
      url={https://arxiv.org/abs/2505.15612}, 
}

@misc{SIRI,
      title={SIRI: Scaling Iterative Reinforcement Learning with Interleaved Compression}, 
      author={Haoming Wen and Yushi Bai and Juanzi Li and Jie Tang},
      year={2025},
      eprint={2509.25176},
      archivePrefix={arXiv},
      primaryClass={cs.LG},
      url={https://arxiv.org/abs/2509.25176}, 
}

@misc{DLER,
      title={DLER: Doing Length pEnalty Right - Incentivizing More Intelligence per Token via Reinforcement Learning}, 
      author={Shih-Yang Liu and Xin Dong and Ximing Lu and Shizhe Diao and Mingjie Liu and Min-Hung Chen and Hongxu Yin and Yu-Chiang Frank Wang and Kwang-Ting Cheng and Yejin Choi and Jan Kautz and Pavlo Molchanov},
      year={2025},
      eprint={2510.15110},
      archivePrefix={arXiv},
      primaryClass={cs.LG},
      url={https://arxiv.org/abs/2510.15110}, 
}

@misc{KL-Cov,
      title={The Entropy Mechanism of Reinforcement Learning for Reasoning Language Models}, 
      author={Ganqu Cui and Yuchen Zhang and Jiacheng Chen and Lifan Yuan and Zhi Wang and Yuxin Zuo and Haozhan Li and Yuchen Fan and Huayu Chen and Weize Chen and Zhiyuan Liu and Hao Peng and Lei Bai and Wanli Ouyang and Yu Cheng and Bowen Zhou and Ning Ding},
      year={2025},
      eprint={2505.22617},
      archivePrefix={arXiv},
      primaryClass={cs.LG},
      url={https://arxiv.org/abs/2505.22617}, 
}

@misc{grpo,
      title={DeepSeekMath: Pushing the Limits of Mathematical Reasoning in Open Language Models}, 
      author={Zhihong Shao and Peiyi Wang and Qihao Zhu and Runxin Xu and Junxiao Song and Xiao Bi and Haowei Zhang and Mingchuan Zhang and Y. K. Li and Y. Wu and Daya Guo},
      year={2024},
      eprint={2402.03300},
      archivePrefix={arXiv},
      primaryClass={cs.CL},
      url={https://arxiv.org/abs/2402.03300}, 
}

@misc{dapo,
      title={DAPO: An Open-Source LLM Reinforcement Learning System at Scale}, 
      author={Qiying Yu and Zheng Zhang and Ruofei Zhu and Yufeng Yuan and Xiaochen Zuo and Yu Yue and Weinan Dai and Tiantian Fan and Gaohong Liu and Lingjun Liu and Xin Liu and Haibin Lin and Zhiqi Lin and Bole Ma and Guangming Sheng and Yuxuan Tong and Chi Zhang and Mofan Zhang and Wang Zhang and Hang Zhu and Jinhua Zhu and Jiaze Chen and Jiangjie Chen and Chengyi Wang and Hongli Yu and Yuxuan Song and Xiangpeng Wei and Hao Zhou and Jingjing Liu and Wei-Ying Ma and Ya-Qin Zhang and Lin Yan and Mu Qiao and Yonghui Wu and Mingxuan Wang},
      year={2025},
      eprint={2503.14476},
      archivePrefix={arXiv},
      primaryClass={cs.LG},
      url={https://arxiv.org/abs/2503.14476}, 
}

@misc{ORZ,
      title={Open-Reasoner-Zero: An Open Source Approach to Scaling Up Reinforcement Learning on the Base Model}, 
      author={Jingcheng Hu and Yinmin Zhang and Qi Han and Daxin Jiang and Xiangyu Zhang and Heung-Yeung Shum},
      year={2025},
      eprint={2503.24290},
      archivePrefix={arXiv},
      primaryClass={cs.LG},
      url={https://arxiv.org/abs/2503.24290}, 
}

@misc{deepscaler,
  title={DeepScaleR: Surpassing O1-Preview with a 1.5B Model by Scaling RL},
  author={Michael Luo and Sijun Tan and Justin Wong and Xiaoxiang Shi and William Y. Tang and Manan Roongta and Colin Cai and Jeffrey Luo and Li Erran Li and Raluca Ada Popa and Ion Stoica},
  year={2025},
  url={https://pretty-radio-b75.notion.site/DeepScaleR-Surpassing-O1-Preview-with-a-1-5B-Model-by-Scaling-RL-19681902c1468005bed8ca303013a4e2},
  note={Notion Blog},
  year={2025},
}

@misc{qwen-math,
      title={Qwen2.5-Math Technical Report: Toward Mathematical Expert Model via Self-Improvement}, 
      author={An Yang and Beichen Zhang and Binyuan Hui and Bofei Gao and Bowen Yu and Chengpeng Li and Dayiheng Liu and Jianhong Tu and Jingren Zhou and Junyang Lin and Keming Lu and Mingfeng Xue and Runji Lin and Tianyu Liu and Xingzhang Ren and Zhenru Zhang},
      year={2024},
      eprint={2409.12122},
      archivePrefix={arXiv},
      primaryClass={cs.CL},
      url={https://arxiv.org/abs/2409.12122}, 
}

@inproceedings{
math500,
title={Measuring Mathematical Problem Solving With the {MATH} Dataset},
author={Dan Hendrycks and Collin Burns and Saurav Kadavath and Akul Arora and Steven Basart and Eric Tang and Dawn Song and Jacob Steinhardt},
booktitle={Thirty-fifth Conference on Neural Information Processing Systems Datasets and Benchmarks Track (Round 2)},
year={2021},
url={https://openreview.net/forum?id=7Bywt2mQsCe}
}

@inproceedings{verl,
author = {Sheng, Guangming and Zhang, Chi and Ye, Zilingfeng and Wu, Xibin and Zhang, Wang and Zhang, Ru and Peng, Yanghua and Lin, Haibin and Wu, Chuan},
title = {HybridFlow: A Flexible and Efficient RLHF Framework},
year = {2025},
isbn = {9798400711961},
publisher = {Association for Computing Machinery},
address = {New York, NY, USA},
url = {https://doi.org/10.1145/3689031.3696075},
doi = {10.1145/3689031.3696075},
booktitle = {Proceedings of the Twentieth European Conference on Computer Systems},
pages = {1279–1297},
numpages = {19},
location = {Rotterdam, Netherlands},
series = {EuroSys '25}
}

@misc{FSDP,
      title={PyTorch FSDP: Experiences on Scaling Fully Sharded Data Parallel}, 
      author={Yanli Zhao and Andrew Gu and Rohan Varma and Liang Luo and Chien-Chin Huang and Min Xu and Less Wright and Hamid Shojanazeri and Myle Ott and Sam Shleifer and Alban Desmaison and Can Balioglu and Pritam Damania and Bernard Nguyen and Geeta Chauhan and Yuchen Hao and Ajit Mathews and Shen Li},
      year={2023},
      eprint={2304.11277},
      archivePrefix={arXiv},
      primaryClass={cs.DC},
      url={https://arxiv.org/abs/2304.11277}, 
}

@misc{SGLang,
      title={SGLang: Efficient Execution of Structured Language Model Programs}, 
      author={Lianmin Zheng and Liangsheng Yin and Zhiqiang Xie and Chuyue Sun and Jeff Huang and Cody Hao Yu and Shiyi Cao and Christos Kozyrakis and Ion Stoica and Joseph E. Gonzalez and Clark Barrett and Ying Sheng},
      year={2024},
      eprint={2312.07104},
      archivePrefix={arXiv},
      primaryClass={cs.AI},
      url={https://arxiv.org/abs/2312.07104}, 
}

@inproceedings{TokenSkip,
    title = "{T}oken{S}kip: Controllable Chain-of-Thought Compression in {LLM}s",
    author = "Xia, Heming  and
      Leong, Chak Tou  and
      Wang, Wenjie  and
      Li, Yongqi  and
      Li, Wenjie",
    editor = "Christodoulopoulos, Christos  and
      Chakraborty, Tanmoy  and
      Rose, Carolyn  and
      Peng, Violet",
    booktitle = "Proceedings of the 2025 Conference on Empirical Methods in Natural Language Processing",
    month = nov,
    year = "2025",
    address = "Suzhou, China",
    publisher = "Association for Computational Linguistics",
    url = "https://aclanthology.org/2025.emnlp-main.165/",
    doi = "10.18653/v1/2025.emnlp-main.165",
    pages = "3351--3363",
    ISBN = "979-8-89176-332-6"
}

@misc{CODI,
      title={CODI: Compressing Chain-of-Thought into Continuous Space via Self-Distillation}, 
      author={Zhenyi Shen and Hanqi Yan and Linhai Zhang and Zhanghao Hu and Yali Du and Yulan He},
      year={2025},
      eprint={2502.21074},
      archivePrefix={arXiv},
      primaryClass={cs.CL},
      url={https://arxiv.org/abs/2502.21074}, 
}

@misc{kimi-k1.5,
      title={Kimi k1.5: Scaling Reinforcement Learning with LLMs}, 
      author={Kimi Team and Angang Du and Bofei Gao and Bowei Xing and Changjiu Jiang and Cheng Chen and Cheng Li and Chenjun Xiao and Chenzhuang Du and Chonghua Liao and Chuning Tang and Congcong Wang and Dehao Zhang and Enming Yuan and Enzhe Lu and Fengxiang Tang and Flood Sung and Guangda Wei and Guokun Lai and Haiqing Guo and Han Zhu and Hao Ding and Hao Hu and Hao Yang and Hao Zhang and Haotian Yao and Haotian Zhao and Haoyu Lu and Haoze Li and Haozhen Yu and Hongcheng Gao and Huabin Zheng and Huan Yuan and Jia Chen and Jianhang Guo and Jianlin Su and Jianzhou Wang and Jie Zhao and Jin Zhang and Jingyuan Liu and Junjie Yan and Junyan Wu and Lidong Shi and Ling Ye and Longhui Yu and Mengnan Dong and Neo Zhang and Ningchen Ma and Qiwei Pan and Qucheng Gong and Shaowei Liu and Shengling Ma and Shupeng Wei and Sihan Cao and Siying Huang and Tao Jiang and Weihao Gao and Weimin Xiong and Weiran He and Weixiao Huang and Weixin Xu and Wenhao Wu and Wenyang He and Xianghui Wei and Xianqing Jia and Xingzhe Wu and Xinran Xu and Xinxing Zu and Xinyu Zhou and Xuehai Pan and Y. Charles and Yang Li and Yangyang Hu and Yangyang Liu and Yanru Chen and Yejie Wang and Yibo Liu and Yidao Qin and Yifeng Liu and Ying Yang and Yiping Bao and Yulun Du and Yuxin Wu and Yuzhi Wang and Zaida Zhou and Zhaoji Wang and Zhaowei Li and Zhen Zhu and Zheng Zhang and Zhexu Wang and Zhilin Yang and Zhiqi Huang and Zihao Huang and Ziyao Xu and Zonghan Yang and Zongyu Lin},
      year={2025},
      eprint={2501.12599},
      archivePrefix={arXiv},
      primaryClass={cs.AI},
      url={https://arxiv.org/abs/2501.12599}, 
}

@inproceedings{CoT-Valve,
    title = "{C}o{T}-Valve: Length-Compressible Chain-of-Thought Tuning",
    author = "Ma, Xinyin  and
      Wan, Guangnian  and
      Yu, Runpeng  and
      Fang, Gongfan  and
      Wang, Xinchao",
    editor = "Che, Wanxiang  and
      Nabende, Joyce  and
      Shutova, Ekaterina  and
      Pilehvar, Mohammad Taher",
    booktitle = "Proceedings of the 63rd Annual Meeting of the Association for Computational Linguistics (Volume 1: Long Papers)",
    month = jul,
    year = "2025",
    address = "Vienna, Austria",
    publisher = "Association for Computational Linguistics",
    url = "https://aclanthology.org/2025.acl-long.300/",
    doi = "10.18653/v1/2025.acl-long.300",
    pages = "6025--6035",
    ISBN = "979-8-89176-251-0"
}

@inproceedings{ThinkSwitcher,
    title = "{T}hink{S}witcher: When to Think Hard, When to Think Fast",
    author = "Liang, Guosheng  and
      Zhong, Longguang  and
      Yang, Ziyi  and
      Quan, Xiaojun",
    editor = "Christodoulopoulos, Christos  and
      Chakraborty, Tanmoy  and
      Rose, Carolyn  and
      Peng, Violet",
    booktitle = "Findings of the Association for Computational Linguistics: EMNLP 2025",
    month = nov,
    year = "2025",
    address = "Suzhou, China",
    publisher = "Association for Computational Linguistics",
    url = "https://aclanthology.org/2025.findings-emnlp.278/",
    doi = "10.18653/v1/2025.findings-emnlp.278",
    pages = "5185--5201",
    ISBN = "979-8-89176-335-7"
}

@inproceedings{Thinkless,
title={Thinkless: {LLM} Learns When to Think},
author={Gongfan Fang and Xinyin Ma and Xinchao Wang},
booktitle={The Thirty-ninth Annual Conference on Neural Information Processing Systems},
year={2025},
url={https://openreview.net/forum?id=ariVQf0KZx}
}

@inproceedings{HRMS,
title={Think Only When You Need with Large Hybrid-Reasoning Models},
author={Lingjie Jiang and Xun Wu and Shaohan Huang and Qingxiu Dong and Zewen Chi and Li Dong and Xingxing Zhang and Tengchao Lv and Lei Cui and Furu Wei},
booktitle={The Thirty-ninth Annual Conference on Neural Information Processing Systems},
year={2025},
url={https://openreview.net/forum?id=fDjDVE4qdj}
}

@misc{lp-reg,
      title={Low-probability Tokens Sustain Exploration in Reinforcement Learning with Verifiable Reward}, 
      author={Guanhua Huang and Tingqiang Xu and Mingze Wang and Qi Yi and Xue Gong and Siheng Li and Ruibin Xiong and Kejiao Li and Yuhao Jiang and Bo Zhou},
      year={2025},
      eprint={2510.03222},
      archivePrefix={arXiv},
      primaryClass={cs.LG},
      url={https://arxiv.org/abs/2510.03222}, 
}

@misc{deepseek-v3.2,
      title={DeepSeek-V3.2: Pushing the Frontier of Open Large Language Models}, 
      author={DeepSeek-AI and Aixin Liu and Aoxue Mei and Bangcai Lin and Bing Xue and Bingxuan Wang and Bingzheng Xu and Bochao Wu and Bowei Zhang and Chaofan Lin and Chen Dong and Chengda Lu and Chenggang Zhao and Chengqi Deng and Chenhao Xu and Chong Ruan and Damai Dai and Daya Guo and Dejian Yang and Deli Chen and Erhang Li and Fangqi Zhou and Fangyun Lin and Fucong Dai and Guangbo Hao and Guanting Chen and Guowei Li and H. Zhang and Hanwei Xu and Hao Li and Haofen Liang and Haoran Wei and Haowei Zhang and Haowen Luo and Haozhe Ji and Honghui Ding and Hongxuan Tang and Huanqi Cao and Huazuo Gao and Hui Qu and Hui Zeng and Jialiang Huang and Jiashi Li and Jiaxin Xu and Jiewen Hu and Jingchang Chen and Jingting Xiang and Jingyang Yuan and Jingyuan Cheng and Jinhua Zhu and Jun Ran and Junguang Jiang and Junjie Qiu and Junlong Li and Junxiao Song and Kai Dong and Kaige Gao and Kang Guan and Kexin Huang and Kexing Zhou and Kezhao Huang and Kuai Yu and Lean Wang and Lecong Zhang and Lei Wang and Liang Zhao and Liangsheng Yin and Lihua Guo and Lingxiao Luo and Linwang Ma and Litong Wang and Liyue Zhang and M. S. Di and M. Y Xu and Mingchuan Zhang and Minghua Zhang and Minghui Tang and Mingxu Zhou and Panpan Huang and Peixin Cong and Peiyi Wang and Qiancheng Wang and Qihao Zhu and Qingyang Li and Qinyu Chen and Qiushi Du and Ruiling Xu and Ruiqi Ge and Ruisong Zhang and Ruizhe Pan and Runji Wang and Runqiu Yin and Runxin Xu and Ruomeng Shen and Ruoyu Zhang and S. H. Liu and Shanghao Lu and Shangyan Zhou and Shanhuang Chen and Shaofei Cai and Shaoyuan Chen and Shengding Hu and Shengyu Liu and Shiqiang Hu and Shirong Ma and Shiyu Wang and Shuiping Yu and Shunfeng Zhou and Shuting Pan and Songyang Zhou and Tao Ni and Tao Yun and Tian Pei and Tian Ye and Tianyuan Yue and Wangding Zeng and Wen Liu and Wenfeng Liang and Wenjie Pang and Wenjing Luo and Wenjun Gao and Wentao Zhang and Xi Gao and Xiangwen Wang and Xiao Bi and Xiaodong Liu and Xiaohan Wang and Xiaokang Chen and Xiaokang Zhang and Xiaotao Nie and Xin Cheng and Xin Liu and Xin Xie and Xingchao Liu and Xingkai Yu and Xingyou Li and Xinyu Yang and Xinyuan Li and Xu Chen and Xuecheng Su and Xuehai Pan and Xuheng Lin and Xuwei Fu and Y. Q. Wang and Yang Zhang and Yanhong Xu and Yanru Ma and Yao Li and Yao Li and Yao Zhao and Yaofeng Sun and Yaohui Wang and Yi Qian and Yi Yu and Yichao Zhang and Yifan Ding and Yifan Shi and Yiliang Xiong and Ying He and Ying Zhou and Yinmin Zhong and Yishi Piao and Yisong Wang and Yixiao Chen and Yixuan Tan and Yixuan Wei and Yiyang Ma and Yiyuan Liu and Yonglun Yang and Yongqiang Guo and Yongtong Wu and Yu Wu and Yuan Cheng and Yuan Ou and Yuanfan Xu and Yuduan Wang and Yue Gong and Yuhan Wu and Yuheng Zou and Yukun Li and Yunfan Xiong and Yuxiang Luo and Yuxiang You and Yuxuan Liu and Yuyang Zhou and Z. F. Wu and Z. Z. Ren and Zehua Zhao and Zehui Ren and Zhangli Sha and Zhe Fu and Zhean Xu and Zhenda Xie and Zhengyan Zhang and Zhewen Hao and Zhibin Gou and Zhicheng Ma and Zhigang Yan and Zhihong Shao and Zhixian Huang and Zhiyu Wu and Zhuoshu Li and Zhuping Zhang and Zian Xu and Zihao Wang and Zihui Gu and Zijia Zhu and Zilin Li and Zipeng Zhang and Ziwei Xie and Ziyi Gao and Zizheng Pan and Zongqing Yao and Bei Feng and Hui Li and J. L. Cai and Jiaqi Ni and Lei Xu and Meng Li and Ning Tian and R. J. Chen and R. L. Jin and S. S. Li and Shuang Zhou and Tianyu Sun and X. Q. Li and Xiangyue Jin and Xiaojin Shen and Xiaosha Chen and Xinnan Song and Xinyi Zhou and Y. X. Zhu and Yanping Huang and Yaohui Li and Yi Zheng and Yuchen Zhu and Yunxian Ma and Zhen Huang and Zhipeng Xu and Zhongyu Zhang and Dongjie Ji and Jian Liang and Jianzhong Guo and Jin Chen and Leyi Xia and Miaojun Wang and Mingming Li and Peng Zhang and Ruyi Chen and Shangmian Sun and Shaoqing Wu and Shengfeng Ye and T. Wang and W. L. Xiao and Wei An and Xianzu Wang and Xiaowen Sun and Xiaoxiang Wang and Ying Tang and Yukun Zha and Zekai Zhang and Zhe Ju and Zhen Zhang and Zihua Qu},
      year={2025},
      eprint={2512.02556},
      archivePrefix={arXiv},
      primaryClass={cs.CL},
      url={https://arxiv.org/abs/2512.02556}, 
}

@article{human-eval,
  title={Evaluating Large Language Models Trained on Code},
  author={Mark Chen and Jerry Tworek and Heewoo Jun and Qiming Yuan and Henrique Ponde de Oliveira Pinto and Jared Kaplan and Harri Edwards and Yuri Burda and Nicholas Joseph and Greg Brockman and Alex Ray and Raul Puri and Gretchen Krueger and Michael Petrov and Heidy Khlaaf and Girish Sastry and Pamela Mishkin and Brooke Chan and Scott Gray and Nick Ryder and Mikhail Pavlov and Alethea Power and Lukasz Kaiser and Mohammad Bavarian and Clemens Winter and Philippe Tillet and Felipe Petroski Such and Dave Cummings and Matthias Plappert and Fotios Chantzis and Elizabeth Barnes and Ariel Herbert-Voss and William Hebgen Guss and Alex Nichol and Alex Paino and Nikolas Tezak and Jie Tang and Igor Babuschkin and Suchir Balaji and Shantanu Jain and William Saunders and Christopher Hesse and Andrew N. Carr and Jan Leike and Josh Achiam and Vedant Misra and Evan Morikawa and Alec Radford and Matthew Knight and Miles Brundage and Mira Murati and Katie Mayer and Peter Welinder and Bob McGrew and Dario Amodei and Sam McCandlish and Ilya Sutskever and Wojciech Zaremba},
  year={2021},
  eprint={2107.03374},
  archivePrefix={arXiv},
  primaryClass={cs.LG}
}

@misc{LCB,
      title={LiveCodeBench: Holistic and Contamination Free Evaluation of Large Language Models for Code}, 
      author={Naman Jain and King Han and Alex Gu and Wen-Ding Li and Fanjia Yan and Tianjun Zhang and Sida Wang and Armando Solar-Lezama and Koushik Sen and Ion Stoica},
      year={2024},
      eprint={2403.07974},
      archivePrefix={arXiv},
      primaryClass={cs.SE},
      url={https://arxiv.org/abs/2403.07974}, 
}

@misc{gspo,
      title={Group Sequence Policy Optimization}, 
      author={Chujie Zheng and Shixuan Liu and Mingze Li and Xiong-Hui Chen and Bowen Yu and Chang Gao and Kai Dang and Yuqiong Liu and Rui Men and An Yang and Jingren Zhou and Junyang Lin},
      year={2025},
      eprint={2507.18071},
      archivePrefix={arXiv},
      primaryClass={cs.LG},
      url={https://arxiv.org/abs/2507.18071}, 
}

@misc{gmpo,
      title={Geometric-Mean Policy Optimization}, 
      author={Yuzhong Zhao and Yue Liu and Junpeng Liu and Jingye Chen and Xun Wu and Yaru Hao and Tengchao Lv and Shaohan Huang and Lei Cui and Qixiang Ye and Fang Wan and Furu Wei},
      year={2025},
      eprint={2507.20673},
      archivePrefix={arXiv},
      primaryClass={cs.CL},
      url={https://arxiv.org/abs/2507.20673}, 
}

@misc{ttrl,
      title={TTRL: Test-Time Reinforcement Learning}, 
      author={Yuxin Zuo and Kaiyan Zhang and Li Sheng and Shang Qu and Ganqu Cui and Xuekai Zhu and Haozhan Li and Yuchen Zhang and Xinwei Long and Ermo Hua and Biqing Qi and Youbang Sun and Zhiyuan Ma and Lifan Yuan and Ning Ding and Bowen Zhou},
      year={2025},
      eprint={2504.16084},
      archivePrefix={arXiv},
      primaryClass={cs.CL},
      url={https://arxiv.org/abs/2504.16084}, 
}

@inproceedings{TCC,
    title = "s1: Simple test-time scaling",
    author = "Muennighoff, Niklas  and
      Yang, Zitong  and
      Shi, Weijia  and
      Li, Xiang Lisa  and
      Fei-Fei, Li  and
      Hajishirzi, Hannaneh  and
      Zettlemoyer, Luke  and
      Liang, Percy  and
      Cand{\`e}s, Emmanuel  and
      Hashimoto, Tatsunori",
    editor = "Christodoulopoulos, Christos  and
      Chakraborty, Tanmoy  and
      Rose, Carolyn  and
      Peng, Violet",
    booktitle = "Proceedings of the 2025 Conference on Empirical Methods in Natural Language Processing",
    month = nov,
    year = "2025",
    address = "Suzhou, China",
    publisher = "Association for Computational Linguistics",
    url = "https://aclanthology.org/2025.emnlp-main.1025/",
    doi = "10.18653/v1/2025.emnlp-main.1025",
    pages = "20275--20321",
    ISBN = "979-8-89176-332-6"
}

@misc{CoD,
      title={Chain of Draft: Thinking Faster by Writing Less}, 
      author={Silei Xu and Wenhao Xie and Lingxiao Zhao and Pengcheng He},
      year={2025},
      eprint={2502.18600},
      archivePrefix={arXiv},
      primaryClass={cs.CL},
      url={https://arxiv.org/abs/2502.18600}, 
}

@misc{NoThinking,
      title={Reasoning Models Can Be Effective Without Thinking}, 
      author={Wenjie Ma and Jingxuan He and Charlie Snell and Tyler Griggs and Sewon Min and Matei Zaharia},
      year={2025},
      eprint={2504.09858},
      archivePrefix={arXiv},
      primaryClass={cs.AI},
      url={https://arxiv.org/abs/2504.09858}, 
}

@inproceedings{
Dynasor-CoT,
title={Reasoning Without Self-Doubt: More Efficient Chain-of-Thought Through Certainty Probing},
author={Yichao Fu and Junda Chen and Yonghao Zhuang and Zheyu Fu and Ion Stoica and Hao Zhang},
booktitle={ICLR 2025 Workshop on Foundation Models in the Wild},
year={2025},
url={https://openreview.net/forum?id=wpK4IMJfdX}
}

@inproceedings{
SEAL,
title={{SEAL}: Steerable Reasoning Calibration of Large Language Models for Free},
author={Runjin Chen and Zhenyu Zhang and Junyuan Hong and Souvik Kundu and Zhangyang Wang},
booktitle={Second Conference on Language Modeling},
year={2025},
url={https://openreview.net/forum?id=klPszYDIRT}
}

@misc{DEER,
      title={Dynamic Early Exit in Reasoning Models}, 
      author={Chenxu Yang and Qingyi Si and Yongjie Duan and Zheliang Zhu and Chenyu Zhu and Qiaowei Li and Minghui Chen and Zheng Lin and Weiping Wang},
      year={2025},
      eprint={2504.15895},
      archivePrefix={arXiv},
      primaryClass={cs.CL},
      url={https://arxiv.org/abs/2504.15895}, 
}

@InProceedings{Arena-Hard,
  title = 	 {From Crowdsourced Data to High-quality Benchmarks: Arena-Hard and Benchbuilder Pipeline},
  author =       {Li, Tianle and Chiang, Wei-Lin and Frick, Evan and Dunlap, Lisa and Wu, Tianhao and Zhu, Banghua and Gonzalez, Joseph E. and Stoica, Ion},
  booktitle = 	 {Proceedings of the 42nd International Conference on Machine Learning},
  pages = 	 {34209--34231},
  year = 	 {2025},
  editor = 	 {Singh, Aarti and Fazel, Maryam and Hsu, Daniel and Lacoste-Julien, Simon and Berkenkamp, Felix and Maharaj, Tegan and Wagstaff, Kiri and Zhu, Jerry},
  volume = 	 {267},
  series = 	 {Proceedings of Machine Learning Research},
  month = 	 {13--19 Jul},
  publisher =    {PMLR},
  url = 	 {https://proceedings.mlr.press/v267/li25h.html}
}

@misc{AlpacaEval,
      title={Length-Controlled AlpacaEval: A Simple Way to Debias Automatic Evaluators}, 
      author={Yann Dubois and Balázs Galambosi and Percy Liang and Tatsunori B. Hashimoto},
      year={2025},
      eprint={2404.04475},
      archivePrefix={arXiv},
      primaryClass={cs.LG},
      url={https://arxiv.org/abs/2404.04475}, 
}

@inproceedings{
MMLU,
title={Measuring Massive Multitask Language Understanding},
author={Dan Hendrycks and Collin Burns and Steven Basart and Andy Zou and Mantas Mazeika and Dawn Song and Jacob Steinhardt},
booktitle={International Conference on Learning Representations},
year={2021},
url={https://openreview.net/forum?id=d7KBjmI3GmQ}
}

@inproceedings{
GPQA,
title={{GPQA}: A Graduate-Level Google-Proof Q\&A Benchmark},
author={David Rein and Betty Li Hou and Asa Cooper Stickland and Jackson Petty and Richard Yuanzhe Pang and Julien Dirani and Julian Michael and Samuel R. Bowman},
booktitle={First Conference on Language Modeling},
year={2024},
url={https://openreview.net/forum?id=Ti67584b98}
}

@inproceedings{
BFCL,
title={The Berkeley Function Calling Leaderboard ({BFCL}): From Tool Use to Agentic Evaluation of Large Language Models},
author={Shishir G Patil and Huanzhi Mao and Fanjia Yan and Charlie Cheng-Jie Ji and Vishnu Suresh and Ion Stoica and Joseph E. Gonzalez},
booktitle={Forty-second International Conference on Machine Learning},
year={2025},
url={https://openreview.net/forum?id=2GmDdhBdDk}
}

@inproceedings{tokenselect,
    title = "{T}oken{S}elect: Efficient Long-Context Inference and Length Extrapolation for {LLM}s via Dynamic Token-Level {KV} Cache Selection",
    author = "Wu, Wei  and
      Pan, Zhuoshi  and
      Fu, Kun  and
      Wang, Chao  and
      Chen, Liyi  and
      Bai, Yunchu  and
      Wang, Tianfu  and
      Wang, Zheng  and
      Xiong, Hui",
    editor = "Christodoulopoulos, Christos  and
      Chakraborty, Tanmoy  and
      Rose, Carolyn  and
      Peng, Violet",
    booktitle = "Proceedings of the 2025 Conference on Empirical Methods in Natural Language Processing",
    month = nov,
    year = "2025",
    address = "Suzhou, China",
    publisher = "Association for Computational Linguistics",
    url = "https://aclanthology.org/2025.emnlp-main.1079/",
    doi = "10.18653/v1/2025.emnlp-main.1079",
    pages = "21264--21281",
    ISBN = "979-8-89176-332-6"
}

@inproceedings{LEMMA,
    title = "{LEMMA}: Learning from Errors for {M}athe{M}atical Advancement in {LLM}s",
    author = "Pan, Zhuoshi  and
      Li, Yu  and
      Lin, Honglin  and
      Pei, Qizhi  and
      Tang, Zinan  and
      Wu, Wei  and
      Ming, Chenlin  and
      Zhao, H. Vicky  and
      He, Conghui  and
      Wu, Lijun",
    editor = "Che, Wanxiang  and
      Nabende, Joyce  and
      Shutova, Ekaterina  and
      Pilehvar, Mohammad Taher",
    booktitle = "Findings of the Association for Computational Linguistics: ACL 2025",
    month = jul,
    year = "2025",
    address = "Vienna, Austria",
    publisher = "Association for Computational Linguistics",
    url = "https://aclanthology.org/2025.findings-acl.605/",
    doi = "10.18653/v1/2025.findings-acl.605",
    pages = "11615--11639",
    ISBN = "979-8-89176-256-5"
}

@inproceedings{MMEA,
 author = {Chen, Liyi and Sun, Ying and Zhang, Shengzhe and Ye, Yuyang and Wu, Wei and Xiong, Hui},
 booktitle = {Advances in Neural Information Processing Systems},
 doi = {10.52202/079017-3794},
 editor = {A. Globerson and L. Mackey and D. Belgrave and A. Fan and U. Paquet and J. Tomczak and C. Zhang},
 pages = {119386--119410},
 publisher = {Curran Associates, Inc.},
 title = {Tackling Uncertain Correspondences for Multi-Modal Entity Alignment},
 url = {https://proceedings.neurips.cc/paper_files/paper/2024/file/d7ed243b13831bdd468f35039936bcef-Paper-Conference.pdf},
 volume = {37},
 year = {2024}
}

@inproceedings{SEPIT,
author = {Wu, Wei and Wang, Chao and Chen, Liyi and Yin, Mingze and Zhu, Yiheng and Fu, Kun and Ye, Jieping and Xiong, Hui and Wang, Zheng},
title = {Structure-Enhanced Protein Instruction Tuning: Towards General-Purpose Protein Understanding with LLMs},
year = {2025},
isbn = {9798400714542},
publisher = {Association for Computing Machinery},
address = {New York, NY, USA},
url = {https://doi.org/10.1145/3711896.3737138},
doi = {10.1145/3711896.3737138},
booktitle = {Proceedings of the 31st ACM SIGKDD Conference on Knowledge Discovery and Data Mining V.2},
pages = {3216–3227},
numpages = {12},
location = {Toronto ON, Canada},
series = {KDD '25}
}
